%% file: main.tex
\documentclass[11pt]{article}
\usepackage[margin=1in]{geometry}
\usepackage{times}

\usepackage[round,authoryear]{natbib}

\usepackage[utf8]{inputenc} % allow utf-8 input
\usepackage[T1]{fontenc}    % use 8-bit T1 fonts
\usepackage{hyperref}       % hyperlinks
\usepackage{url}            % simple URL typesetting
\usepackage{booktabs}       % professional-quality tables
\usepackage{amsfonts}       % blackboard math symbols
\usepackage{nicefrac}       % compact symbols for 1/2, etc.
\usepackage{microtype}      % microtypography
\usepackage{xcolor}         % colors

\usepackage{wrapfig}  
\usepackage{amsmath}
\usepackage{amssymb}
\usepackage{mathtools}
\usepackage{amsthm}
\usepackage{bbm}
\usepackage{multirow}
\usepackage[table]{xcolor}

\usepackage{algorithm}
\usepackage{algpseudocode}

\usepackage[capitalize,noabbrev]{cleveref}
\input{mysymb}

\usepackage[most]{tcolorbox}
\usepackage{xcolor}
\usepackage{enumitem}
\usepackage{siunitx}

\newtcolorbox{yellowhighlight}{
  colback=yellow!15,
  colframe=yellow!60!black,
  boxrule=0.4pt,
  arc=2pt,
  left=4pt,
  right=20pt,
  top=-1pt,
  bottom=4pt,
  boxsep=0pt
}

\title{Constrained Diffusion Models with Primal-Dual Inference}

\date{}
\author{%
  Samar Hadou
  \quad 
  Yi\u{g}it Berkay Uslu \quad 
  Alejandro Ribeiro\footnote{Emails: \texttt{selaraby,ybuslu,aribeiro@engineering.upenn.edu}} \\
  \\
  Department of Electrical and Systems Engineering\\
  University of Pennsylvania\\
}

\begin{document}

\maketitle

\begin{abstract}
This paper develops constrained diffusion models with primal-dual inference (PDI) to sample from optimal distributions of entropy-regularized optimization problems with \emph{average} constraints. We formalize constrained sampling in the Lagrangian dual domain, where the optimal distribution takes the form of a Gibbs distribution indexed by the optimal dual variable.
Rather than estimating this dual multiplier before sampling and freezing it throughout generation, PDI jointly infers the optimal primal distribution and its parametrizing dual variable. Each reverse diffusion step denoises using the score field associated with the current multiplier and then updates the multiplier through dual ascent using the estimated constraint violation of the denoised samples. To enable this conditional score field, we train a single dual-conditioned score network over the family of Gibbs distributions induced by the dual variables encountered during inference. We prove that the dual iterates converge to a neighborhood of the dual optimum, and that
the generated distribution approaches that of an oracle sampler that knows the optimal multiplier, up to a bias floor. We evaluate PDI on constrained sampling from a mixture of Gaussians, wireless resource allocation, and portfolio management. Our code is available at: \href{https://github.com/SMRhadou/PDI-Diffusion}{https://github.com/SMRhadou/PDI-Diffusion}.
\end{abstract}

\input{sections/01intro}

\input{sections/02related_work}

\input{sections/03constrianed_energy_diffusion}

\input{sections/04Convergence}

\input{sections/04Numerical}

\input{sections/06Conclusions}

\section*{Acknowledgment}
Generative AI tools were used to assist with proofreading and improving the grammar and clarity of the manuscript. They were also used to support experimental implementation and to assist in drafting portions of the experimental appendices. All technical content, experimental results, and final text were reviewed and edited by the authors.

%%%%%%%%%%%%%%%%%%%%%%%%%%%%%%%%%%%%%%%%%%%%%%%%%%%%%%%%%%%%

% \clearpage
\bibliographystyle{plainnat}
\bibliography{Bib}

\clearpage
\appendix
\input{sections/appendix_extended_related_work}

\input{sections/appendix_b_proposed_algorithms}

\input{sections/appendix_clean_proof_final}

\input{sections/appendix_dual_training}

\input{sections/appendix_synthetic_data}

\input{sections/appendix_wra}

\input{sections/appendix_portfolio}

\end{document}

%% file: mysymb.tex
\def \bbxt {\widetilde\bbx_{t+1}}
\def \muDagger {\mu_{\bblambda}^\dagger}
\def \muDaggert {\mu_{\bblambda_t}^\dagger}

\def \bbxt {\widetilde\bbx_{t}}
\def \xpred {\mbE \big[\, \bby_0 \, | \, \bby_{T-t} = \bbx_{t}, \bblambda_t \,\big]}
\def \xpredtheta {\widehat{\bby}_0(\bbx_{t}, \bblambda_t)}
\def \xpredplusone {\mbE \big[\, \bby_0 \, | \, \bby_{T-t} = \bbx_{t}, \bblambda_t \,\big]}
\def \xpredtilde {\mbE \big[\, \bby_0 \, | \, \bby_{T-t} = \widetilde\bbx_{t}, \bblambda_t \,\big]}

\def \KL {D_{\text{KL}}}

\def \kernel {K_t^{{\bblambda}^*}}

\newcommand{\diff}{\mathop{}\!\mathrm{d}}

\newtheorem{assumption}{\hspace{0pt}\bf Assumption}
\newtheorem{lemma}{\hspace{0pt}\bf Lemma}
\newtheorem{proposition}{\hspace{0pt}\bf Proposition}
\newtheorem{theorem}{\hspace{0pt}\bf Theorem}
\newtheorem{corollary}{\hspace{0pt}\bf Corollary}
\newcounter{excercise}
\newcounter{excercisepart}

\def \SNR     {\text{\normalfont SNR}   }

\def \Cov     {\text{\normalfont Cov}  }

\def \reals    {{\mathbb R}}

\newcommand{\argmin}{\operatornamewithlimits{argmin}}

\def\mbE{{\ensuremath{\mathbb E}}}

\def\ccalB{{\ensuremath{\mathcal B}}}

\def\ccalD{{\ensuremath{\mathcal D}}}

\def\ccalG{{\ensuremath{\mathcal G}}}
\def\ccalH{{\ensuremath{\mathcal H}}}

\def\ccalL{{\ensuremath{\mathcal L}}}

\def\ccalN{{\ensuremath{\mathcal N}}}
\def\ccalO{{\ensuremath{\mathcal O}}}
\def\ccalP{{\ensuremath{\mathcal P}}}

\def\ccalV{{\ensuremath{\mathcal V}}}
\def\ccalS{{\ensuremath{\mathcal S}}}
\def\ccalT{{\ensuremath{\mathcal T}}}
\def\ccalX{{\ensuremath{\mathcal X}}}

\def\ccal0{{\ensuremath{\mathcal 0}}}
\def\bbA{{\ensuremath{\mathbf A}}}

\def\bbH{{\ensuremath{\mathbf H}}}
\def\bbI{{\ensuremath{\mathbf I}}}

\def\bbX{{\ensuremath{\mathbf X}}}
\def\bbY{{\ensuremath{\mathbf Y}}}

\def\bbb{{\ensuremath{\mathbf b}}}

\def\bbf{{\ensuremath{\mathbf f}}}

\def\bbh{{\ensuremath{\mathbf h}}}

\def\bbr{{\ensuremath{\mathbf r}}}
\def\bbw{{\ensuremath{\mathbf w}}}
\def\bbu{{\ensuremath{\mathbf u}}}
\def\bbv{{\ensuremath{\mathbf v}}}
\def\bbs{{\ensuremath{\mathbf s}}}

\def\bbx{{\ensuremath{\mathbf x}}}
\def\bby{{\ensuremath{\mathbf y}}}
\def\bbz{{\ensuremath{\mathbf z}}}
\def\bb0{{\ensuremath{\mathbf 0}}}
\def\rmP{{\ensuremath\text{P}}}

\def\rmd{{\ensuremath\text{d}}}

\def\rmp{{\ensuremath\text{p}}}

\def\bbdelta{\boldsymbol{\delta}}
\def\bbepsilon{\boldsymbol{\epsilon}}

\def\bbtheta{\boldsymbol{\theta}}

\def\bblambda{\boldsymbol{\lambda}}

\def\bbmu{\boldsymbol{\mu}}

\def\bbphi{\boldsymbol{\phi}}

\def\bbLambda{\boldsymbol{\Lambda}}

\def\bbSigma{\boldsymbol{\Sigma}}

\DeclarePairedDelimiterX{\infdivx}[2]{(}{)}{%
  #1\;\delimsize\|\;#2%
}

\makeatletter
\DeclareRobustCommand{\cev}[1]{%
  {\mathpalette\do@cev{#1}}%
}
\newcommand{\do@cev}[2]{%
  \vbox{\offinterlineskip
    \sbox\z@{$\m@th#1 x$}%
    \ialign{##\cr
      \hidewidth\reflectbox{$\m@th#1\vec{}\mkern4mu$}\hidewidth\cr
      \noalign{\kern-\ht\z@}
      $\m@th#1#2$\cr
    }%
  }%
}

\makeatother

%% file: sections/01intro.tex
\section{Introduction}

Diffusion models have become a dominant framework for sampling from complex, high-dimensional distributions by learning to reverse a noising process \citep{sohl2015deep, ho2020denoising, song2021scorebased}. While their most visible successes lie in unconstrained perceptual tasks such as image \citep{rombach2022highresolution}, audio \citep{kong2021diffwave}, and video synthesis \citep{ho2022video}, the same modeling capabilities are valuable for constrained optimization problems whose optimal solutions are distributions rather than single points. This is because when the constraints need only hold on average, randomized solutions can do strictly better. In wireless resource allocation, for instance, interference can prevent any single power allocation from meeting every user's rate requirement, whereas a time-sharing policy alternating between allocations meet them on average while maximizing the network utility \citep{uslu2025generative, darabi2024diffusion}. Similarly, portfolio management seeks diverse allocations that maximize expected returns under expected risk constraints \citep{tiwari2026generativediffusionmodelriskneutral, he2026factorbased}. Standard diffusion samplers, however, lack effective mechanisms to enforce such average constraints. Guidance and projection methods correct each sample individually \citep{chung2023diffusion, christopher2024constrained}, which imposes a stricter per-sample requirement and can exclude solutions that are feasible only on average. This paper bridges that gap by developing \emph{primal--dual inference} (PDI), a diffusion sampler that enforces average constraints during sampling.

Entropy-regularized problems over distributions with average constraints admit a clean form in the Lagrangian dual domain \citep{chamon2024PDL, khalafi2024constrained}. For a fixed dual variable, the distributional minimizer of the Lagrangian is a Gibbs distribution whose energy is linear in the dual variable. This connects constrained sampling to the vast literature on diffusion-based samplers for unnormalized Gibbs targets \citep{zhang2022path, vargas2023denoising, berner2024optimal, richter2024improved, akhoundsadegh2024idem}. In principle, one could first solve for the optimal dual variable and then train a diffusion model to sample from the corresponding Gibbs distribution. In our setting, however, the sampler must serve a family of problem instances, each with its own optimal multiplier, and each multiplier is defined through expectations under the very Gibbs distribution the sampler is meant to learn. Dual-training (DT) approaches \citep{khalafi2024constrained, khalafi2025composition} resolve this circularity by alternating dual ascent with score-model training. Retraining the score model to convergence after every dual update is expensive, while truncating it leaves a score model that blends past targets rather than matching the Gibbs distribution at the final multiplier. Either way, once training finalizes, the sampler is frozen and cannot respond to constraint violations or adapt to shifted constraint levels during generation. Primal--dual Langevin Monte Carlo \citep{chamon2024PDL} instead updates the multiplier at sampling time, but it relies on a Markov chain that must mix, which is slow for multi-modal targets, and it does not learn a generative model that transfers across instances.

PDI proposes a different view by bringing the sampling-time dual updates of primal--dual Langevin into the reverse diffusion itself. We make the multiplier part of the sampling state. At each denoising step, PDI updates the samples with the score field of the current multiplier, evaluates the constraints at Tweedie estimates of the denoised samples \citep{efron2011tweedie}, and takes a dual ascent step.
% on their average over an ensemble that shares the multiplier. 
Because this step uses the ensemble average with a shared multiplier, the constraints are enforced on average rather than on each sample. Different from Langevin dynamics, PDI does not rely on mixing. The reverse process transports noise to the target in a fixed number of steps, following a path-dependent sequence of Gibbs score fields while the multiplier approaches its optimum. These scores are supplied by a single score network conditioned on the multiplier and the problem instance, so that one model represents the entire family of Gibbs targets. Since no samples from these targets are available directly, the network regresses onto Monte Carlo score estimates computed from the energy alone \citep{akhoundsadegh2024idem}.

Our theoretical analysis makes the benefits of this coupling precise. The PDI multipliers converge to a neighborhood of its optimum within a single reverse pass, while the generated distribution approaches that of an oracle sampler that knows the optimal multiplier, up to a floor set by the Tweedie, reverse-sampling and score-approximation errors. Thus not knowing the optimal multiplier in advance costs little beyond this bias, and the generated samples are approximately feasible in expectation. Empirically, we show that PDI matches the feasibility of projection- and Langevin-based baselines with a better objective and better mode coverage in a constrained Gaussian mixture setup. In wireless resource allocation, PDI attains better feasibility and tail rates than DT and degrades more gracefully under shifted constraint levels. DT becomes competitive only after an additional stage that freezes its final multiplier and continues score training for that single Gibbs target, which requires more than ten times as many outer training iterations as PDI. This suggests that DT's gap lies in fitting the score at the right multiplier, which the conditional network provides without retraining. Freezing the multiplier within PDI at its own final or time-averaged value increases the total constraint violation, indicating that the inference-time coupling itself also contributes.

We make the following contributions.
\begin{enumerate}[label = \textbf{(C\arabic*)}]
    \vspace{-5pt}
    \item We cast constrained sampling with average constraints as a saddle-point problem in the Lagrangian dual domain, where the optimal distribution is a Gibbs distribution indexed by an optimal dual variable.
    \item We propose PDI, a reverse-diffusion sampler that treats the dual multiplier as part of its state and solves the dual problem within a single diffusion pass by interleaving each denoising diffusion step with a dual ascent step.
    \item We train a single score network that represents the entire family of Gibbs targets parametrized by arbitrary dual multipliers, without requiring the optimal multiplier. The network adapts to new problem instances and shifted constraint levels at inference time without retraining.
    \item We prove that the dual iterates converge to a neighborhood of the optimal multiplier despite using biased gradient estimates, and that the output distribution of PDI is within $\mathcal{O}(T^{-1/2})$ in KL divergence of an oracle sampler that knows the optimal multiplier, up to a bias floor. 
    \item We validate PDI empirically on a constrained Gaussian mixture, wireless power control, and portfolio construction. PDI achieves a better objective than per-sample baselines at matched feasibility, matches strengthened dual training with over ten times fewer outer training iterations, and degrades more gracefully under shifted constraint levels.
    \vspace{-5pt}
\end{enumerate}

\begin{figure}
    \centering
    \includegraphics[width=\linewidth]{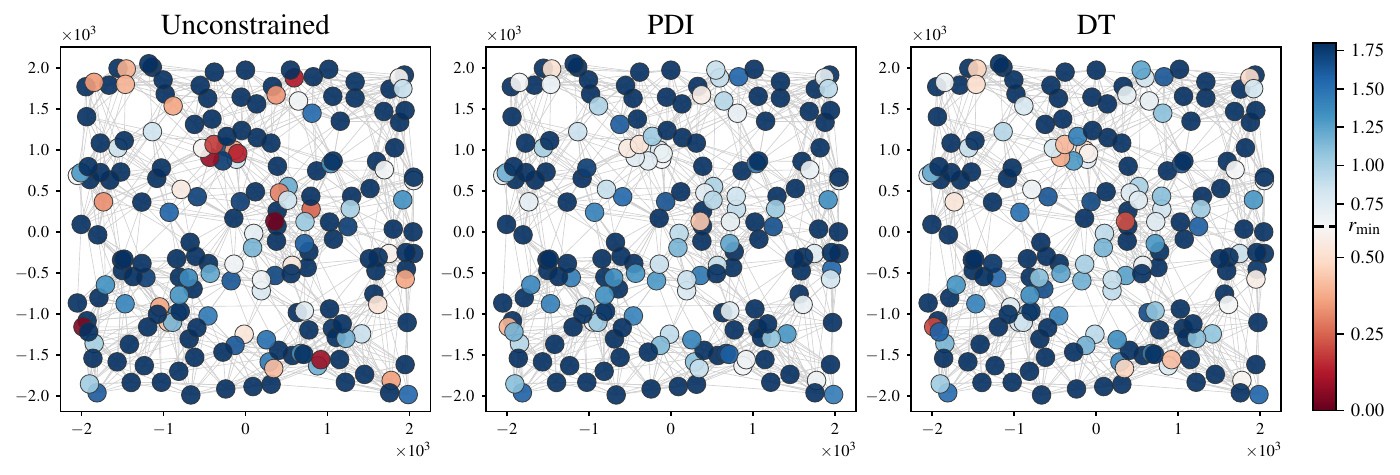}
    \vspace{-20pt}
    \caption{\small \textbf{Wireless power allocation under ergodic minimum-rate $r_{\min}$ constraints.} 
    % Nodes represent users and colors indicate achieved rates. Due to interference, users must time-share communication channels to satisfy minimum-rate requirements on average. 
    Nodes represent users and colors indicate how well the rate requirement of each user is met, with red marking those that fall short. Here, average constraints are meaningful because mutual interference prevents any single solution from satisfying every requirement at once. Optimal solutions must therefore alternate samples (i.e., power allocations and/or channel access) over time, so that some compensate for others. 
    % Node color indicates how well the requirement of each user is met, with red marking those that fall short.
    The unconstrained sampler leaves many users below the target rate (red nodes), while DT partially improves feasibility through training-time dual updates. PDI further improves the lower-rate users by updating the dual variables during denoising.
    }
    \vspace{-10pt}
    \label{fig:intro}
\end{figure}

Figure~\ref{fig:intro} illustrates the effect of inference-time dual updates on wireless power allocation, where users share a channel and each requires a minimum average rate. An unconstrained sampler leaves many users underserved. DT improves feasibility, but its multipliers are set during training, so it cannot correct violations that arise during generation. PDI updates the multiplier as it denoises, steering samples toward feasibility on average and lifting the worst-served users closer to their targets.

%% file: sections/02related_work.tex
\vspace{-5pt}
\section{Related Work}
\vspace{-5pt}

% \vspace{-10pt}
\paragraph{Diffusion model sampling with constraints.}
Guidance techniques steer reverse diffusions toward reward or constraint alignment by modifying the score in Tweedie's denoising formula \citep{efron2011tweedie}, as in classifier guidance \citep{dhariwal2021diffusion}, classifier-free guidance \citep{ho2022classifierfree}, and CLIP-based variants \citep{nichol2022glide}. A separate line of work enforces domain-feasibility constraints directly on the diffusion process through boundary reflections \citep{lou2023reflected}, mirror maps \citep{liu2023mirror}, Riemannian formulations \citep{de2022riemannian}, log-barrier constructions \citep{fishman2023metropolis}, feasibility projections \citep{christopher2024constrained}, and constrained reverse steps \citep{huang2024trust}. Most importantly, these methods address pointwise constraints rather than distributional (average) ones.

\vspace{-5pt}
\paragraph{Constrained diffusion modeling with Lagrangian dual formulations.}
\vspace{-5pt}

Closely related to our work are \citep{khalafi2024constrained, khalafi2025composition, chamon2024PDL, gutjahr2026constrained}, which adopt primal–dual formulations for average constraints. Specifically, \citep{khalafi2024constrained, khalafi2025composition} pair outer-loop dual ascent with inner-loop score matching but retrain the score network for each dual iterate, while primal–dual Langevin Monte Carlo \citep{chamon2024PDL} jointly evolves samples and multipliers at sampling time but forgoes a learned generative model. Our method shares the Lagrangian saddle-point structure of these works but differs in two key aspects. First, the dual variables evolve jointly with the denoising trajectory rather than in a separate training loop or ergodic Langevin chain. Second, a single score network, conditioned on the dual variable, is trained across a distribution of multipliers rather than refitted per iterate. An 
extended related work is provided in Appendix~\ref{app:rlw}.

%% file: sections/03constrianed_energy_diffusion.tex
\section{Constrained Diffusion Models}
Let $\ccalX \subset \reals^p$ be a compact domain of decision variables $\bbx$ and $\ccalP(\ccalX)$ the space of all probability distributions supported on $\ccalX$. 
A constraint function $\bbf: \ccalX \to \reals^d$ encodes the set of requirements that need only hold \emph{on average}. 
Given also an objective cost $f_0: \ccalX \to \reals$, and a temperature $\beta > 0$, the target distribution $\mu^\star$ solves the entropy-regularized problem,
\begin{align}\label{eq:constrained_problem}
    P^\star ~=~ \min_{\mu\in\ccalP(\ccalX)} \quad & 
    % \Phi(\mu) ~\coloneqq~ 
    \mbE_\mu\big[f_0(\bbx)\big] - \beta\,\ccalH(\mu), \nonumber\\
    \text{subject to} \quad & \mbE_\mu\big[\bbf(\bbx)\big] ~\preceq~ \mathbf 0,
\end{align}
\noindent where $\ccalH(\mu)=-\int_\ccalX\mu\log\mu\,\rmd\bbx$ is the differential entropy if $\mu$ admits a density, and $\ccalH(\mu)=-\infty$ otherwise. Throughout, $f_0$ and $\bbf$ are bounded and measurable. Problem~\eqref{eq:constrained_problem} is equivalently KL-regularized as $\KL(\mu\|\nu)=\int\log(\rmd\mu/\rmd\nu)\,\rmd\mu$  toward a uniform distribution $\nu=\mathrm{Uniform}(\ccalX)$.

Constraining averages rather than individual samples is valuable in settings such as multi-player cooperative games and wireless resource optimization problems, where a single deterministic policy may not satisfy all competing requirements simultaneously. For instance, when interfering users cannot all be served at once, a $\mu^\star$ can alternate between decisions that compensate for another via time-sharing to improve the trade-off between objective utility and feasibility.
% In such settings,  
In our applications, $f_0$ and $\bbf$ depend on a problem instance $\ccalG$, e.g., the channel gains of a wireless network. The solution $\mu^\star$ is therefore instance-specific, but we suppress this dependence for simplicity of presentation.

The problem in \eqref{eq:constrained_problem} is ($\beta$-strongly) convex in $\mu$ with a zero duality gap and can be handled in the dual domain. Given Lagrangian multipliers $\bblambda \succeq \mathbf{0}$, we form the Lagrangian function,
\vspace{-5pt}
\begin{align}\label{eq:Lagrangian}
    \ccalL \big( \mu, \bblambda \big) ~=~
                                    \mbE_{\mu} \Big[ \, f_0(\bbx) + \bblambda^\top \bbf(\bbx) \, \Big] - \beta \, \ccalH(\mu),
\end{align}
% where $\bblambda \succeq \mathbf{0}$ collects the Lagrangian multipliers. 
The dual function is $g(\bblambda) = \min_{\mu \in \ccalP(\ccalX)} \ccalL(\mu, \bblambda)$, and the dual problem is given by
\vspace{-5pt}
\begin{align}\label{eq:dual_problem}
    D^\star ~=~ \max_{\bblambda \succeq \bb0} ~ g(\bblambda) ~=~ \max_{\bblambda \succeq \mathbf{0}} \, \min_{\mu \in \ccalP(\ccalX)} 
                                    \mbE_{\mu} \Big[ \, f_0(\bbx) + \bblambda^\top \bbf(\bbx) \, \Big] - \beta \, \ccalH(\mu).
\end{align}
For a fixed $\bblambda$, the inner minimizer of the Lagrangian is the Gibbs distribution~\citep[p.~34]{dupuis1997weak}, 
\begin{align} \label{eq:gibbs-minimizer}
    \muDagger(\bbx) =  \frac{ \exp \big( - E(\bbx, \bblambda) \big) }{Z(\lambda)}, \qquad
    E(\bbx, \bblambda) ~=~ \frac{1}{\beta} \Big( f_0(\bbx) + {\bblambda}^\top \bbf(\bbx) \Big),
\end{align}
\noindent where $E(\bbx, \bblambda)$ is the energy, and $Z(\bblambda) = 
% \mbE_{\nu} \big[ e^{-E(\bbx, \bblambda)} \big] =
\int_{\ccalX} \exp( -E (\bbx, \bblambda) ) \diff \bbx$ is the partition function for normalization $\int_{\ccalX} \muDagger \diff \bbx = 1$. The Lagrangian minimizer $\muDagger$ is thus an exponential tilt of the implicit uniform prior distribution $\nu$, and the dual function is given by $g(\bblambda) = -\beta \log Z(\bblambda)$.

% The two problems are equivalent by the following proposition.
The next proposition shows that the primal and dual problems are equivalent under Slater, and the optimal distribution $\mu^\star$ is also a Gibbs distribution parametrized by an optimal dual multiplier $\bblambda^\star$.

\begin{proposition}[Strong duality \citep{chamon2024PDL}] \label{lemma:strong-duality}
Under Slater's condition (stated as Assumption \ref{as:feasibility} in the Appendix~\ref{app:assumptions}), and for $\beta>0$, we have $P^\star = D^\star$, and the dual problem \eqref{eq:dual_problem} is attained at some finite $\bblambda^\star \in \bbLambda$. The optimal pair $(\mu^\star, \bblambda^\star)$ is a saddle-point of the Lagrangian, i.e.,
$
    \ccalL(\mu^\star, \bblambda) ~\leq~ 
                \ccalL(\mu^\star, \bblambda^\star) ~\leq~
                                            \ccalL(\mu, \bblambda^\star),
$ for any $\mu \in \ccalP(\ccalX)$ and $\bblambda \in \reals^d_+$, and $\mu^\star = \mu^\dagger_{\bblambda^\star}$.
\end{proposition}
% The proof is relegated to Appendix \ref{proof:strong-duality}. 

The dual function is smooth and concave, with
\begin{align}\label{eq:dual_gradient}
    \nabla g(\bblambda) ~=~ \mbE_{\muDagger}\big[\bbf(\bbx)\big],
    \qquad
    \nabla^2 g(\bblambda) ~=~ -\tfrac{1}{\beta}\,\Cov_{\muDagger}\big(\bbf(\bbx)\big).
\end{align}
Dual ascent on~\eqref{eq:dual_problem} only requires the expected constraint values under the Gibbs distribution indexed by the current multiplier. The energy is known in closed form for every $\bblambda$; however, this does not reveal the optimal distribution, since $\bblambda^\star$ is defined implicitly through the distribution it indexes.
Nor does it provide samples; $\muDagger$ is known only up to its normalizing constant, and importance-reweighting or rejecting proposals from $\nu$ become inefficient in high dimensions~\citep{chatterjee2018sample}.

Solving~\eqref{eq:dual_problem} iteratively thus requires samples from $\muDagger$ at every multiplier visited by dual ascent, and existing approaches differ in where this dual search takes place. Primal--dual Langevin Monte Carlo~\citep{chamon2024PDL} performs it at sampling time where a Langevin chain sidesteps the normalizing constant by following the energy gradient, and its samples supply the constraint estimates for each dual update. Prior constrained diffusion models~\citep{khalafi2024constrained, gutjahr2026constrained} perform it at training time, alternating dual updates with score-model training, which leaves a frozen sampler once training is complete. Diffusion-based samplers for unnormalized densities~\citep{zhang2022path, vargas2023denoising, berner2024optimal, richter2024improved} avoid the mixing time of Langevin chains~\citep{menz2014poincare, dalalyan2017theoretical} by transporting Gaussian noise through a sequence of smoothed marginals. 

We take a different route and propose \emph{primal--dual inference (PDI)}, which moves the dual search to sampling time within a diffusion process by making the multiplier a state of the reverse process.

\subsection{Primal--Dual Inference}

We want to generate samples from the Gibbs family $\{\muDagger\,|\,\bblambda\in\bbLambda\}$, which contains $\mu^\star=\mu^\dagger_{\bblambda^\star}$ by Proposition~\ref{lemma:strong-duality}.
We build the sampler from the reverse process of a score-based diffusion model~\citep{song2021scorebased}. To construct it, consider the variance-preserving (VP) forward process that would noise samples of $\muDagger$,
\begin{align}\label{eq:forward_process_family}
    \bby_{\tau}(\bblambda) ~=~ \sqrt{a_\tau} \, \bby_{\tau-1}(\bblambda) + \sqrt{b_\tau} \, \bbepsilon_\tau, \qquad \bby_0(\bblambda) \sim \muDagger,
\end{align}
with $a_\tau=1-b_\tau$ and $\bbepsilon_\tau \sim \ccalN(\mathbf{0}, \bbI) \perp \bby_0$. Its marginals $q_{\tau}^{\bblambda}\coloneqq q_\tau(\cdot|\bblambda)$ are the Gaussian smoothings $q_\tau(\bbx|\bblambda)=\int\ccalN(\bbx;\alpha_\tau\bby,\sigma_\tau^2\bbI)\,\muDagger(\rmd\bby)$, where $\alpha_\tau^2=\prod_{s\leq\tau}a_s$ and $\sigma_\tau^2=1-\alpha_\tau^2$, so that $q_T^{\bblambda}\approx\ccalN(\mathbf 0,\bbI)$ hold for every $\bblambda$. We write $\SNR_\tau\coloneqq\alpha_\tau^2/\sigma_\tau^2$ for the signal-to-noise ratio (SNR) which decays monotonically in $\tau$. Since samples of $\muDagger$ are unavailable, we never simulate this process and thus cannot train on its noised samples as standard diffusion models do. The process only serves to construct the reverse process, and we explain shortly how its scores are learned without such samples.

We index the reverse (sampling) process by inference time $t = T - \tau$. It starts from $\bbx_0 \sim \ccalN(\mathbf{0}, \bbI)$ at $t=0$ and runs to $t=T$, producing a sample $\bbx_T$. For each fixed $\bblambda$, the reverse process is
\begin{align}\label{eq:backward_process_family}
    \bbx_{t+1} ~=~ \frac{1}{\sqrt{a_{T-t}}} \Big( \bbx_{t} + b_{T-t}\,\nabla \log q_{T-t}(\bbx_t|\bblambda) \Big) + \sqrt{b_{T-t}}\, \bbepsilon_t,
\end{align}
\noindent where the $\bbepsilon_t \sim \ccalN(\mathbf{0}, \bbI)$ are independent of each other and of $\bbx_0$. This is a discretization of the time reversal of the forward diffusion~\citep{anderson1982reverse, song2021scorebased}, and its final iterate $\bbx_T$ is approximately distributed as $\muDagger$ for large enough $T$ and suitable noise schedules~\citep{chen2023sampling}. In particular, it would approximately sample $\mu^\star$ if $\bblambda^\star$ were known.

PDI updates the multiplier along the reverse process instead. At each step, a dual-conditioned score model $s_{\bbtheta}(\bbx_t,t,\bblambda_t)$ drives both a primal and a dual update:
\begin{align}
    \bbx_{t+1} & ~=~ \frac{1}{\sqrt{a_{T-t}}} \Big( \bbx_{t} + b_{T-t} \, s_{\bbtheta}(\bbx_{t},t,\bblambda_t) \Big) + \sqrt{b_{T-t}}\, \bbepsilon_t, \tag{PDI-P}\label{eq:sample_update} \\
    \bblambda_{t+1} & ~=~ \Big[ \, \bblambda_{t} + \eta_t \, \mbE_{p_t} \big[ \bbf\big( \widehat\bby_0(\bbx_t, \bblambda_t) \big) \big] \Big]_+. \tag{PDI-D} \label{eq:dual_update}
\end{align}
\noindent Here, $p_t$ is the law of $\bbx_t$ under PDI, $\eta_t>0$ is the dual step size, and 
$[\cdot]_{+}$ is the projection onto the nonnegative orthant.
The clean-sample estimate $\widehat\bby_0(\bbx,\bblambda)=\big(\bbx+\sigma_{T-t}^2\,s_{\bbtheta}(\bbx,t,\bblambda)\big)/\alpha_{T-t}$ is Tweedie's estimate of the posterior mean $\mbE[\,\bby_0\,|\,\bby_{T-t}=\bbx,\bblambda]$, and the expectation in \eqref{eq:dual_update} is estimated over an ensemble of samples that share one multiplier. Since $\widehat\bby_0(\bbx_t,\bblambda_t)$ reuses the score evaluation $s_{\bbtheta}(\bbx_t,t,\bblambda_t)$ of~\eqref{eq:sample_update}, each PDI step requires a single score evaluation, as in fixed-multiplier sampling, plus the evaluation of the constraints $\bbf$. In practice, we run several chains in parallel, each with its own multiplier, and this preserves the average-constraint interpretation while adding robustness to perturbations in the final multiplier (see Algorithm~\ref{alg:inference}).

A well-trained score model $s_{\bbtheta}$ represents the entire Gibbs family by approximating the $\bblambda$-conditional Gibbs scores, $s_{\bbtheta}(\bbx,t,\bblambda)\approx\nabla\log q_{T-t}(\bbx|\bblambda)$. 
% The network also takes different problem instances $\ccalG$ as input, which we suppress from the notation. 
It is also conditioned on the problem instance $\ccalG$, which we suppress from the notation. Conditioning diffusion models on side information is standard practice~\citep{dhariwal2021diffusion, ho2022classifierfree, rombach2022highresolution}. Conditioning on $\bblambda$ is particularly benign, as $\bblambda$ enters the energy $E(\cdot, \bblambda)$ linearly, and the score field varies smoothly with $\bblambda$ at every noise level (Lemma~\ref{lemma:score_lipschitz}).
Unlike standard diffusion models, however, the network cannot be trained on noisy data samples because the samples from the Gibbs targets are unavailable. Instead, its regression targets are estimated from the energy alone by reweighting Gaussian candidates for the clean sample using the Gibbs density~\citep{akhoundsadegh2024idem, huang2024reverse}. These targets are evaluated at the noisy samples and the relevant multipliers visited by PDI rollouts; Appendix~\ref{app:training} details the training pipeline. In our experiments, we refer to PDI with the trained network as PDI-Net, whereas the training-free variant, which uses these Monte Carlo estimates directly in place of $s_{\bbtheta}$, is called PDI-MC.

%% file: sections/04Convergence.tex
\section{Convergence Analysis}\label{sec:analysis}

Two aspects of PDI set it apart from standard dual ascent and diffusion sampling. Firstly, the dual step in \eqref{eq:dual_update} is not an exact, unbiased dual ascent step. By~\eqref{eq:dual_gradient}, the dual gradient is the expected constraint value under $\muDaggert$. In contrast, \eqref{eq:dual_update} evaluates $\bbf$ at Tweedie estimates, which introduces bias from the nonlinearity of $\bbf$, from sampling $p_t$ rather than $q_{T-t}^{\bblambda_t}$, and from score errors. 
Theorem~\ref{theorem:dual-convergence} shows that this bias does not prevent convergence, but it determines how close the dual iterates get to $\bblambda^\star$. In particular, the dual mismatch bias of the early steps, where the Tweedie estimates are coarsest, decays geometrically. 

Secondly, because the multiplier changes along the trajectory, the marginal law $p_t$ generated by PDI is generally not equal to $q_{T-t}(\cdot|\bblambda_t)$. The latter only defines the local fixed-multiplier score field used at that step. In other words, PDI is a path-dependent reverse process rather than the exact reversal of any single forward process in \eqref{eq:forward_process_family}. Theorem~\ref{theorem:primal-convergece} nevertheless 
compares $p_T$ with the output of an oracle sampler, which knows $\bblambda^\star$ and keeps it fixed throughout, and bounds their terminal KL-divergence by the score error and the dual mismatch accumulated along the trajectory.

The PDI dual update is biased relative to exact dual ascent because it is evaluated along the evolving reverse process using Tweedie estimates and an approximate conditional score. This bias combines four effects analyzed separately in Appendix~\ref{app:bias}: the Tweedie posterior-mean error, the marginal mismatch induced by the evolving multiplier, the reverse-sampling error of the exact-score sampler, and the accumulated score approximation error. We denote their aggregate effect at step \(s\) by \(\bbb_s\). In Theorem~\ref{theorem:dual-convergence}, we establish convergence of the last dual iterate to a neighborhood of the optimal multiplier in the presence of this bias.
\begin{theorem}[Dual Convergence]\label{theorem:dual-convergence}
% Let $g(\bblambda) := \ccalL(\mu^\dagger_{\bblambda}, \bblambda)$ be the dual function with strong concavity index $\vartheta$.
For a sequence of $\{\bblambda_t\}$ generated by PDI, it holds under Assumptions 
\ref{as:f_lipschitz}--\ref{as:VP} that, for $\eta_t = \eta < \frac{2}{\vartheta}, \forall t$, and $r := 1- \frac{\eta \vartheta}{2}$, the last-iterate multiplier satisfies
    \vspace{-5pt}
    \begin{align}
        \|\bblambda_{T} - \bblambda^\star\|^2  
                   &  ~\leq~
                   r^T \|\bblambda_{0} - \bblambda^\star\|^2 
                   + \frac{2\eta}{\vartheta} R^2
                   +  \frac{2\eta}{\vartheta} \sum_{s<T} r^{T-1-s} 
                    \|\bbb_s\|^2.
    \end{align}
    \end{theorem}  
    \vspace{-10pt}

The proof is provided in Appendix \ref{prrof:dual-convergence}. The bound reveals a favorable interaction between the diffusion trajectory and the dual recursion. Errors incurred early in denoising, when the Tweedie estimates are coarsest, are strongly discounted by the geometric factor \(r^{T-1-s}\). Conversely, the later updates receive larger weights precisely as the Tweedie posterior-mean and score-parameterization errors decrease toward the data end. 
% The remaining trajectory-history error is controlled by the dual step size and the stability of the reverse process. 
Consequently, large early errors are progressively forgotten, and the convergence region is governed primarily by the dual step size and the residual bias of the later updates. We additionally derive an \(O(T^{-1/2})\) time-average dual-gap bound for \(\eta=\kappa/\sqrt{T}\) in Appendix~\ref{app:average_dual_gap}, which is used subsequently to establish the primal convergence result.

\begin{theorem}[Primal Convergence]\label{theorem:primal-convergece}
    Let $\widetilde{p}_{T}^{\,\bblambda}$ be the output law of \eqref{eq:backward_process_family} with exact scores, fixed multiplier $\bblambda$, and the initialization of PDI. Under Assumptions~\ref{as:f_lipschitz}--~\ref{as:lambda-lipschitz}, and for $\eta = \kappa / \sqrt{T}$, the KL divergence between the terminal distribution of PDI and the oracle's output distribution satisfies
    \begin{align}
            % D_{\mathrm{KL}}\Big(p_T\|q_0(\cdot|\bblambda^\star)\Big)
            D_{\mathrm{KL}}\Big(p_T\|{\widetilde{p}_{T}}^{\,\bblambda^\star}\Big)
            \le{}&
            % D_{\mathrm{KL}}\Big(p_0\|q_T(\cdot|\bblambda^\star)\Big)
             % +
            O\!\left(\frac{1}{\sqrt{T}}\right)
            +
            O\!\left(
            \frac{1}{T}\sum_{t=0}^{T-1}\chi_{t}^2 + \delta_{t}^2 \right)
            + O\!\left( \varepsilon^2
            \right),
    \end{align}
    where $\chi_{t}^2$ and $\delta_{t}^2$ are the Tweedie and reverse-sampling errors (Lemmas~\ref{lemma:bias_in_posterior} and~\ref{lemma:bias_disc}), and $\varepsilon$ is the score approximation error.
    \end{theorem}

The proof is in Appendix \ref{proof:primal-convergence}. The theorem shows that primal accuracy depends on the \emph{trajectory-wide} quality of the dual updates, rather than only on the final multiplier. The $O(T^{-1/2})$ term reflects the average dual suboptimality accumulated along inference, while the remaining terms quantify how imperfect Tweedie estimates and score approximation limit the attainable accuracy. In particular, errors that are large only over a small portion of the reverse trajectory are averaged out rather than determining the terminal distribution by themselves. 
The oracle's own error with respect to $\mu^\star$ stems only from the discretization and initialization, and Corollary~\ref{cor:primal-feasibility} turns both into an approximate feasibility in expectation statement.

%% file: sections/04Numerical.tex
\vspace{-5pt}
\section{Numerical Results}
We provide numerical evidence in three stochastic optimization problems: constrained MoG, wireless resource allocation, and portfolio management. More  details are provided in Appendices \ref{app:MoG}, \ref{app:wra}, and \ref{app:pf}. 

\vspace{-5pt}
\subsection{Mixture of Gaussians}
We first evaluate our approach on sampling from a weighted mixture of Gaussians in $\reals^{d}$,
$
    f_0(\bbx) = -\log \sum_{k=1}^K w_k \, \ccalN\big( \bbx; \bbmu_k, \bbSigma_k \big),
$
truncated to a polytope $\{\bbx: \bbA^\top\bbx\preceq\bbb\}$.  The optimization problem is
\vspace{-10pt}
\begin{align}\label{eq:synthetic}
  P^\star_{\text{MoG}} ~=~ \min_{\mu(\bbx)}\;
    \mathbb{E}_{\bbx\sim \mu} \bigl[f_0(\bbx)\bigr] - \beta \ccalH(\mu)
  \quad\text{s.t.}\quad
    \mathbb{E}_{\bbx\sim \mu} \bigl[\bbA^\top\bbx - \bbb\bigr]
      \;\preceq\; \mathbf{0},
\end{align}
i.e., the sampler must produce a distribution that concentrates on
high-density regions of the mixture while satisfying $M$ linear
inequality constraints in expectation. 

\textbf{Baselines.}  We compare our approach to two baselines:  i) a projected diffusion model (PDM) \citep{christopher2024constrained}, and ii) primal-dual Langevin (PDL) dynamics \citep{chamon2024PDL}. For PDM, we reuse our trained model with $\bblambda = \mathbf{0}$ as the base model that samples from the unconstrained MoG and performs a Cimmino parallel projection \citep{SLOBODA1991435} onto the polytope at each step.

\begin{figure}
    \centering
    \includegraphics[width=\linewidth]{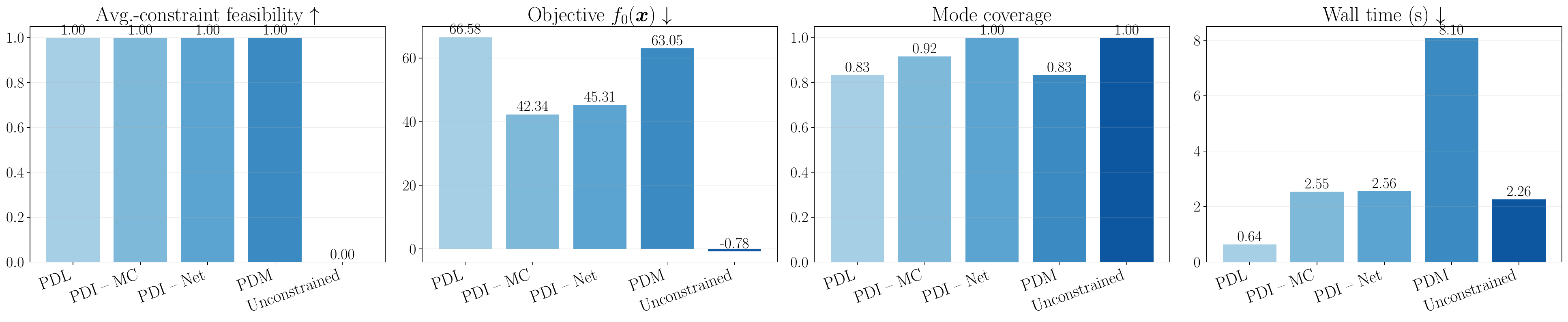}
    \vspace{-15pt}
    \caption{\small \textbf{Mixture of Gaussians} ($K=12$ modes with $M =10$ constraints in a $30$-dimensional space). While PDI, PDL and PDM maintain full feasibility, PDI exhibits the best objective and mode diversity.}
    \vspace{-8pt}
    \label{fig:MoG-performance}
\end{figure}

\textbf{Results.}
As shown in Figure \ref{fig:MoG-performance}, all constrained methods achieve the same constraint feasibility. However, our method (PDI-Net and PDI-MC) obtains a better objective value than the other constrained baselines. This is because the two baselines enforce the constraints on every generated sample, whereas our formulation only requires the constraints to hold in expectation. Pointwise enforcement is therefore more conservative and can discard useful samples for optimality that violate the constraints individually but could be offset by other samples in the distribution. 
We also include an ablation on the dual variable through the unconstrained baseline, which runs our model with $\bblambda$ fixed to zero at all time steps. As the figure shows, the unconstrained model ignores the constraints entirely and achieves the lowest objective. This performance confirms that the constrained behavior is not merely a consequence of the training algorithm.

\vspace{-3pt}
\subsection{Wireless Resource Allocation}
\label{sec:exp_wra}

For a given network state~$\mathbf{H}$ with $N=200$ users, the goal is to maximize the ergodic sum-rate while satisfying a minimum ergodic rate constraint for each user. That is, 
\vspace{-3pt}
\begin{equation}
\label{eq:power_control}
P^\star_{\mathrm{pc}} ~=~ 
                \max_{\mu(\bbx)} \; \mathbf{1}^\top \mathbf{r}\!\left(\mu, \mathbf{H}\right) 
                + \beta \ccalH(\mu), 
                \quad \text{s.t.} \quad \mathbf{r}\!\left(\mu, \mathbf{H}\right) \geq \mathbf{1} \cdot r_{\min},
\end{equation}
where $\mu$ denotes the distribution of power allocations, and $r_{\min}$ is the minimum ergodic rate requirement. The optimal solution is rarely a degenerate policy. Since users compete for the communication channels and cause interference, activating all users simultaneously is typically suboptimal and may prevent the rate constraints from being satisfied. Instead, the optimal distribution often corresponds to a multimodal switching policy, in which different subsets of users transmit at different times while meeting their rate requirements in expectation.

\textbf{Baselines.} In addition to PDL and PDM, we include two other diffusion-based baselines: i) diffusion posterior sampling (DPS) \citep{chung2023diffusion}, and ii) supervised training (ST) with expert data \citep{uslu2026graphsignaldiffusionmodels}. The expert data are generated by a primal-dual method run for the deterministic version of \eqref{eq:power_control} and its resulting trajectories (after a transient period) are collected to train the diffusion model.

\begin{table}[t]
\centering
\scriptsize
\caption{\small \textbf{Baseline comparisons.} PDI achieves a near-feasible performance and provides the best balance between mean rates and constraint violations in the wireless power allocation problem ($r_{\min}=0.6$).}
\label{tab:wra_performance}
\resizebox{\textwidth}{!}{%
\begin{tabular}{lcccccc}
\toprule
  Method  & Mean rates $(\uparrow)$ & 5th\%ile rate $(\uparrow)$ & 1st\%ile rate $(\uparrow)$ & Mean viol. $(\downarrow)$ & Sum viol. $(\downarrow)$ & Feas. \% $(\uparrow)$   \\
    \midrule
    Unconstrained & $2.957 {\scriptstyle\pm 0.123}$ & $0.239 {\scriptstyle\pm 0.052}$ & $0.081 {\scriptstyle\pm 0.037}$ & $0.0421 {\scriptstyle\pm 0.0078}$ & $8.43 {\scriptstyle\pm 1.57}$ & $85.6 {\scriptstyle\pm 2.4}$   \\
    \midrule
    PDI--Net &
    $\textbf{2.803} {\scriptstyle\pm 0.132}$ & $\textbf{0.613} {\scriptstyle\pm 0.042}$ & $0.424 {\scriptstyle\pm 0.082}$ & $\textbf{0.0062} {\scriptstyle\pm 0.0027}$ & $\textbf{1.25} {\scriptstyle\pm 0.54}$ & $95.5 {\scriptstyle\pm 1.5}$   \\
    PDI--Net (warm-up) & 
    $2.578 {\scriptstyle\pm 0.132}$ & $\textbf{0.670} {\scriptstyle\pm 0.055}$ & $0.429 {\scriptstyle\pm 0.102}$ & $\textbf{0.0056} {\scriptstyle\pm 0.0026}$ & $\textbf{1.12} {\scriptstyle\pm 0.51}$ & $96.7 {\scriptstyle\pm 1.4}$  \\
    PDI--MC & $\textbf{2.839} {\scriptstyle\pm 0.130}$ & $\textbf{0.614} {\scriptstyle\pm 0.052}$ & $0.355 {\scriptstyle\pm 0.120}$ & ${0.0085} {\scriptstyle\pm 0.0037}$ & $1.71 {\scriptstyle\pm 0.75}$ & $95.5 {\scriptstyle\pm 1.5}$ \\
    \midrule
    PDL  & 
    $2.635 {\scriptstyle\pm 0.141}$ & $\textbf{0.641} {\scriptstyle\pm 0.057}$ & $0.405 {\scriptstyle\pm 0.106}$ & ${0.0068} {\scriptstyle\pm 0.0031}$ & $1.36 {\scriptstyle\pm 0.63}$ & $96.2 {\scriptstyle\pm 1.5}$  \\
    PDM  & 
    $2.591 {\scriptstyle\pm 0.157}$ & $0.510 {\scriptstyle\pm 0.082}$ & $0.165 {\scriptstyle\pm 0.107}$ & $0.0180 {\scriptstyle\pm 0.0065}$ & $3.61 {\scriptstyle\pm 1.29}$ & $91.9 {\scriptstyle\pm 2.7}$   \\
    DPS  &
    $2.731 {\scriptstyle\pm 0.138}$ & $0.461 {\scriptstyle\pm 0.071}$ & $0.184 {\scriptstyle\pm 0.082}$ & $0.0198 {\scriptstyle\pm 0.0060}$ & $3.96 {\scriptstyle\pm 1.19}$ & $90.8 {\scriptstyle\pm 2.5}$    \\
    ST  & $2.482 {\scriptstyle\pm 0.134}$ & $\textbf{0.825} {\scriptstyle\pm 0.060}$ & $\textbf{0.666} {\scriptstyle\pm 0.071}$ & $\textbf{0.0007} {\scriptstyle\pm 0.0008}$ & $\textbf{0.14} {\scriptstyle\pm 0.16}$ & $99.3 {\scriptstyle\pm 0.6}$ \\
\bottomrule
\end{tabular}
}
\vspace{-5pt}
\end{table}

\textbf{Results.} Table~\ref{tab:wra_performance} shows that PDI achieves the best overall balance between mean rates (objective) and feasibility represented by the $5$th-percentile rate and mean and sum violations. The $1$st-percentile users, however, are harder to correct by the dual variables and suffer under all methods, except for ST. Similarly to the MoG experiment, PDL and PDM achieve lower mean rates (worse objectives) with no substantial gains in feasibility because they sacrifice optimality for forcing pointwise constraints.
% PDM and DPS struggle to provide comparable performance as they provide less diverse sample. 
Moreover, ST generates feasible solutions at the expense of optimality. Because ST is trained only to imitate expert trajectories, it is agnostic to the dual structure of the constrained distribution and cannot explicitly balance utility against feasibility during sampling. Lastly, instead of using the untrained score model to generate rollouts during the early epochs of training, we initialize these rollouts with the ST model. We observe that the trained PDI-Net with ST warm-up inherits the same bias toward feasibility, achieving a higher, feasible 1st-percentile rate. However, this comes at the expense of the objective value, which decreases from $2.80$ to $2.58$.

\begin{figure}[t]
    \centering
    \includegraphics[width=0.32\linewidth]{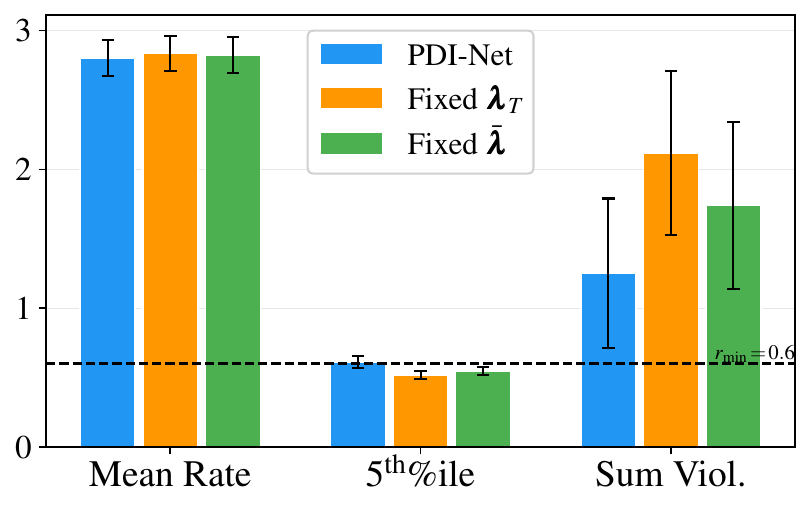}
    \includegraphics[width=0.32\linewidth]{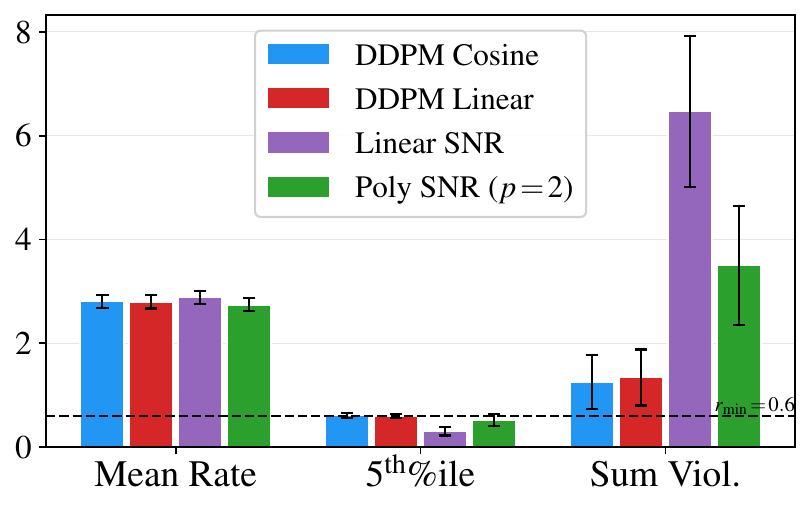}
    \includegraphics[width=0.32\linewidth]{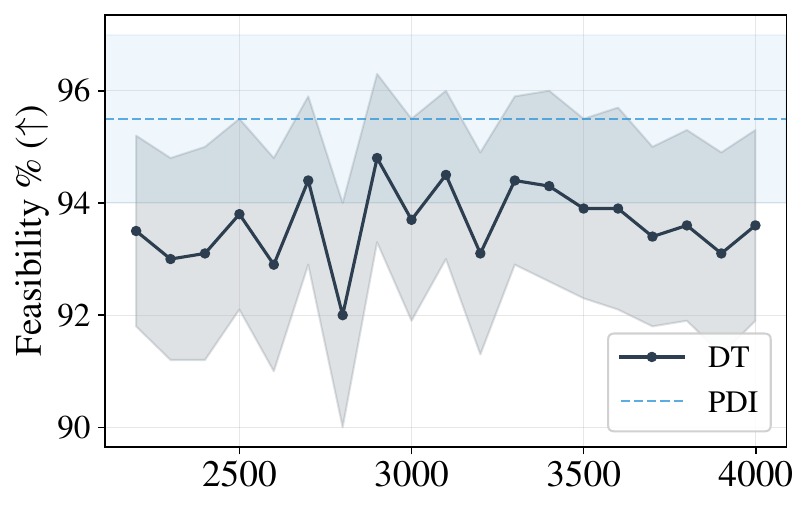}
    \caption{\small \textbf{Ablation.} Comparisons between our base PDI-Net model, plotted in blue, and (left) a PDI-Net while setting $\bblambda_t$ to a fixed value along the denoising trajectory, (middle) a PDI-Net with different noise schedules, and (right) DT checkpoints corresponding to different training epochs and different estimates of the dual variable.}
    \vspace{-15pt}
    \label{fig:ablation_wra}
\end{figure}

\textbf{Ablation.} Figure~\ref{fig:ablation_wra} (left) evaluates the effect of keeping the dual variable fixed during inference. We run the score model with two fixed values: the final dual iterate $\bblambda_T$ and the time-average dual variable $\bar\bblambda$, both computed from our base experiment. The results suggest that the distribution generated by the full PDI dynamics is not equivalent to the Gibbs distribution associated with either fixed value. Instead, it behaves like a mixture induced by the changing dual variable, and this mixture exhibits better feasibility. Among the two values, the Gibbs distribution with $\bar\bblambda$ provides a better approximation to the PDI-generated distribution than the one associated with the final iterate.
To evaluate the role of the noise schedule, we report the performance metrics of four schedules in Figure~\ref{fig:ablation_wra} (middle): the DDPM cosine used in our base experiment, the DDPM linear schedule and the linear and polynomial SNR schedules. The latter two schedules fail to generate feasible samples.

 \begin{wrapfigure}{r}{0.44\textwidth}
    \centering
    \vspace{-15pt}
    \includegraphics[width=\linewidth]{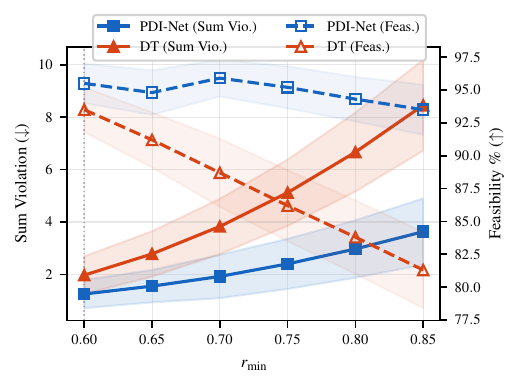}
    \vspace{-25pt}
    \caption{\small \textbf{Out-of-distribution constraints.} The models are trained under $r_{\min}=0.6$ and tested under different values of constraint levels. PDI degrades more gracefully than DT, showing more robustness to distribution shifts.}
    \vspace{-5pt}
    \label{fig:ood}
\end{wrapfigure}
\textbf{Dual training.} To ablate the role of inference-time dual updates, we train a score model $\bbs_{\bbphi}(\bbx_t, t)$ that does not take the dual variable as input. During training, we still update the dual variables using a primal-dual algorithm, so the score model is trained against targets induced by the evolving dual variables. At inference, there are no more dual updates and the trained model therefore acts as a fixed sampling policy that is intended to sample from the optimal distributions. We refer to this approach as dual training (DT) and the training algorithm is provided in Appendix~\ref{app:dt}.

Figures \ref{fig:ablation_wra} (right) and \ref{fig:dt_metrics} compare DT checkpoints along the training trajectory with PDI. Although some DT checkpoints reach dual variables close to those produced by PDI, their generated samples exhibit lower tail rates and lower feasibility. This indicates matching the dual variables alone is 
not sufficient. Since DT does not condition the score model on the current multiplier, a checkpoint trained
over a changing dual trajectory need not represent the score field associated with its current dual value. Instead, it reflects the accumulated effect of the preceding training trajectory. We further verify this in Appendix~\ref{app:wra_results} by freezing DT multipliers and continuing score training, which substantially improves performance but requires a much larger training budget. PDI avoids this limitation by keeping the dual variable as an explicit state during inference. Thus, when the dual update increases the penalty on violated constraints, the penalty is reflected directly in the denoising dynamics.

\textbf{Out-of-distribution constraints.} The dual state conditioning in our PDI score model also benefits OOD performance. In Figure~\ref{fig:ood}, we plot the sum violation and feasibility percentage under different constraint levels $r_{\min}$, with $r_{\min}=0.6$ corresponding to the in-distribution setting. We observe that PDI is more robust to the shift in $r_{\min}$. This is because the score function $\nabla \log q_{T-t}(\bbx_t\mid \bblambda_t)$ does not depend on $r_{\min}$ directly. Instead, changing $r_{\min}$ affects the constraint residual and, therefore, induces a different trajectory of dual variables. Since PDI is trained with both exploitation and exploration over $\bblambda$, the score model learns a family of score fields over a range of in-distribution dual values. Thus, as long as the dual trajectory induced by the new $r_{\min}$ remains close to this learned range, PDI can adapt to the new OOD constraints. 

\subsection{Constrained Portfolio Sampling}
\label{sec:exp_portfolio}
\vspace{-5pt}

We consider constrained portfolio optimization over $N=500$ assets organized into $10$ sectors, with returns generated from a latent factor model. Each portfolio is represented by weights on the probability simplex, and the objective is to maximize expected return over $1,000$ return scenarios. We impose a large and heterogeneous set of constraints covering per-asset variance contributions, sector exposures,  and portfolio diversification. In total, the problem contains $521$ inequality constraints.  The average constraints allow the sampler to trade risk across different portfolio samples, admitting occasional high-risk, high-return allocations while ensuring that the overall exposure of each asset remains within its prescribed budget in expectation.

\begin{table}[t]
\centering
\footnotesize
\caption{\small \textbf{Constrained portfolio sampling}, $N\!=\!500$ assets, 521 constraints, 40 instances $\times$ 1024 samples. Reported are 95th\%ile and max violations over constraints within each family; $\pm$ std across instances. \textbf{Bold} = best, \underline{underlined} = second best among constrained samplers. Time is seconds per instance (RTX\,3090).}
\label{tab:portfolio_p95}
\setlength{\tabcolsep}{4pt}
\resizebox{\textwidth}{!}{%
\begin{tabular}{@{}l cc cc cc c c c@{}}
\toprule
& \multicolumn{2}{c}{Variance Viol. $\downarrow$} & \multicolumn{2}{c}{Sector upper Viol. $\downarrow$} & \multicolumn{2}{c}{Sector lower Viol. $\downarrow$} & $N_{\mathrm{eff}}$ viol. $\downarrow$ & Return $\uparrow$ & Time (s) \\
\cmidrule(lr){2-3}\cmidrule(lr){4-5}\cmidrule(lr){6-7}
Method & 95th\%ile & max & 95th\%ile & max & 95th\%ile & max & & & \\
\midrule
Unconstrained & 0.0013\,\tiny{$\pm$0.0006} & 0.0233\,\tiny{$\pm$0.0247} & 0.0012\,\tiny{$\pm$0.0038} & 0.0021\,\tiny{$\pm$0.0069} & 0.0002\,\tiny{$\pm$0.0008} & 0.0004\,\tiny{$\pm$0.0015} & 0.0000\,\tiny{$\pm$0.0000} & 0.1993\,\tiny{$\pm$0.0228} & 37.9 \\
\midrule
PDI-MC (ours) & \textbf{0.0000\,\tiny{$\pm$0.0000}} & \textbf{0.0000\,\tiny{$\pm$0.0000}} & 0.0046\,\tiny{$\pm$0.0067} & 0.0074\,\tiny{$\pm$0.0105} & \textbf{0.0069\,\tiny{$\pm$0.0069}} & \textbf{0.0098\,\tiny{$\pm$0.0095}} & 1.8662\,\tiny{$\pm$5.7070} & \textbf{0.1773\,\tiny{$\pm$0.0188}} & 29.2 \\
PDI-Net (ours) & \textbf{0.0000\,\tiny{$\pm$0.0000}} & \textbf{0.0000\,\tiny{$\pm$0.0000}} & 0.0041\,\tiny{$\pm$0.0060} & 0.0060\,\tiny{$\pm$0.0082} & 0.0281\,\tiny{$\pm$0.0095} & 0.0350\,\tiny{$\pm$0.0097} & \textbf{0.0000\,\tiny{$\pm$0.0000}} & 0.1537\,\tiny{$\pm$0.0180} & 37.7 \\
\midrule
PDL & \textbf{0.0000\,\tiny{$\pm$0.0000}} & \underline{0.0008\,\tiny{$\pm$0.0008}} & \underline{0.0006\,\tiny{$\pm$0.0014}} & \underline{0.0011\,\tiny{$\pm$0.0024}} & 0.0415\,\tiny{$\pm$0.0028} & 0.0451\,\tiny{$\pm$0.0029} & \underline{0.1666\,\tiny{$\pm$0.8680}} & 0.1334\,\tiny{$\pm$0.0162} & 0.9 \\
PDM & \textbf{0.0000\,\tiny{$\pm$0.0000}} & \textbf{0.0000\,\tiny{$\pm$0.0000}} & \textbf{0.0004\,\tiny{$\pm$0.0011}} & \textbf{0.0006\,\tiny{$\pm$0.0017}} & 0.0480\,\tiny{$\pm$0.0035} & 0.0510\,\tiny{$\pm$0.0037} & \textbf{0.0000\,\tiny{$\pm$0.0000}} & 0.1259\,\tiny{$\pm$0.0160} & 88.3 \\
DPS & \underline{0.0016\,\tiny{$\pm$0.0004}} & 0.0268\,\tiny{$\pm$0.0337} & \underline{0.0006\,\tiny{$\pm$0.0019}} & \underline{0.0011\,\tiny{$\pm$0.0034}} & \underline{0.0176\,\tiny{$\pm$0.0060}} & \underline{0.0209\,\tiny{$\pm$0.0065}} & \textbf{0.0000\,\tiny{$\pm$0.0000}} & \underline{0.1625\,\tiny{$\pm$0.0172}} & 38.2 \\
Equal weight & 0.0001\,\tiny{$\pm$0.0001} & 0.0007\,\tiny{$\pm$0.0007} & 0.0000\,\tiny{$\pm$0.0000} & 0.0000\,\tiny{$\pm$0.0000} & 0.0000\,\tiny{$\pm$0.0000} & 0.0000\,\tiny{$\pm$0.0000} & 440.0000\,\tiny{$\pm$0.0000} & 0.2207\,\tiny{$\pm$0.0235} & $<$0.1 \\
\bottomrule
\end{tabular}%
}
% \vspace{-15pt}
\end{table}

\paragraph{Results.}
Table~\ref{tab:portfolio_p95} shows that PDI achieves a strong return--constraint tradeoff across the different constraint families. Both PDI variants satisfy the variance constraints essentially exactly, while maintaining small violations of the sector constraints. PDI-MC achieves the highest return among the constrained diffusion-based methods, whereas PDI-Net provides a more balanced solution across the different constraint families. In particular, PDI-Net eliminates the diversification violation observed with PDI-MC while retaining a higher return than PDL and PDM. The baselines exhibit different forms of conservatism. PDL and PDM control the variance and upper-sector constraints, but incur larger violations of the lower-sector constraints and obtain lower returns. DPS preserves a competitive return, but does so at the cost of noticeably larger variance violations. These results show that the constraints cannot be treated independently. Enforcing one family more aggressively can shift the violation to another family or reduce the achievable return, while PDI learns a distribution that balances these competing requirements jointly. Additional ablations are provided in Appendix~\ref{app:pf}.

%% file: sections/06Conclusions.tex
\section{Conclusions}

We introduced a primal--dual inference (PDI) framework that enforces distributional constraints by coupling a reverse diffusion process with dual ascent dynamics. We trained a single dual-variable-conditioned score network that serves the entire family of Gibbs targets encountered along the dual trajectory. 
At inference, PDI generates samples through a path-dependent sequence of Gibbs score fields indexed by the evolving multipliers, avoiding the need to first estimate and freeze an optimal dual variable.
% At inference, we produce samples from the target primal distribution as the dual ascent dynamics iterate through near-optimal multipliers, without requiring exact convergence to an optimal dual variable. 
% We analyzed PDI through last-iterate dual convergence, a time-averaged dual-gap bound, and a primal convergence result that controls the terminal distribution through the dual trajectory. 
We showed that the last dual iterate converges to a neighborhood of the optimal dual multiplier and that PDI tracks an oracle sampler that knows it. 
We validated our framework and methodology on constrained Gaussian sampling, wireless power control, and portfolio management. Extending PDI to adaptive dual step sizes, chance-constrained problems, and larger-scale setups are promising directions for future work. Our work has some limitations worth noting. 
% Our primal guarantee relative to an oracle sampler assumes the score network is Lipschitz in the multiplier. 
% Additionally, 
PDI requires problem-specific tuning of the initial dual multiplier distribution used during training and the entropy regularization strength $\beta$, for which principled selection rules remain an open question.

%% file: sections/appendix_extended_related_work.tex
\section{Extended Related Work} \label{app:rlw}

\paragraph{Diffusion models and score-based generative modeling.}

Score-based diffusion models learn a stochastic reversal of a progressive noising process by matching the gradients of the log-density $\nabla_x \log p(x)$, a.k.a. the score function, for noisy observations \citep{hyvarinen2005estimation}. Noise-conditional score networks \citep{song2019generative} train a single network across noise levels and sample via annealed Langevin dynamics, while denoising diffusion probabilistic models \citep{ho2020denoising} frame generation as iterative Gaussian denoising guided by a simplified ELBO. These formulations are unified in \citep{song2021scorebased} as instances of a common SDE, whose time reversal yields a generative process; and admits an equivalent deterministic probability-flow ODE that enables exact likelihood computation and has motivated accelerated sampling schemes such as denoising diffusion implicit models (DDIMs) \citep{song2021denoising, nichol2021improved, lu2022dpmsolver,karras2022elucidating}.

\paragraph{Energy-based models and sampling from Gibbs targets.}

Energy-based models define distributions of the form $p(x) \propto \exp(-E(x))$, whose normalizing constant is generally intractable. Score-based methods sidestep this because the score $\nabla_x \log p(x) = -\nabla_x E(x)$ is independent of this constant. This motivated the use of diffusion as a sampler from unnormalized Gibbs targets by learning a controlled SDE, whose terminal law matches the target, e.g., \citep{zhang2022path, vargas2023denoising, berner2024optimal, richter2024improved, chen2024sequential, domingo2024adjoint, sendera2024improved}. In our setting, the Lagrangian minimizers are also Gibbs distributions whose energies are linear in the dual variables. Crucially, the energies are not fixed as they evolve during inference through dual ascent. Furthermore, the score network is trained over a distribution of dual variables, or equivalently, a distribution of target energies, rather than for a single one.

\paragraph{Guided and constrained diffusion sampling.}

Guidance techniques steer reverse diffusion processes toward improved fidelity or reward/constraint alignment more generally. They do so by modifying the score entering Tweedie's denoising formula \citep{efron2011tweedie}, which expresses the MMSE-estimate of a clean signal (posterior mean) given a noisy signal observation in terms of the score function. Notably, classifier guidance \citep{dhariwal2021diffusion} perturbs the predicted score with the gradient of a noise-aware classifier, trading diversity for fidelity. Classifier-free guidance \citep{ho2022classifierfree} removes the external classifier, instead interpolating between conditional and unconditional modes of a shared model during inference. Subsequent variants, such as CLIP-based guidance \citep{nichol2022glide}, underpin most state-of-the-art diffusion systems. 

Beyond guidance, an extensive body of work enforces domain-feasibility constraints on the diffusion process directly, e.g., boundary reflections \citep{lou2023reflected}, mirror diffusion \citep{liu2023mirror}, Riemannian score-based models \citep{de2022riemannian}, and log-barrier constructions \citep{fishman2023metropolis}. A separate line of works target constraint-aware inference. In \citep{christopher2024constrained}, projected diffusions that apply feasibility projections after each denoising step are proposed, while a trust sampling approach \citep{huang2024trust} reformulates each reverse step as a separate constrained optimization problem. 

Closest to this paper are \citep{khalafi2024constrained, khalafi2025composition, chamon2024PDL}, which adopt a constrained learning formulation for average (expectation) constraints. Specifically, \citep{khalafi2024constrained} shows that the optimal KL-constrained distribution is an entropy-tilted mixture of reference distributions, enforced through a primal–dual training scheme that pairs an outer-loop dual ascent with inner-loop Lagrangian score matching, with \citep{khalafi2025composition} extending this to reward-based alignment and product/mixture compositions. Primal–dual Langevin Monte Carlo (PD-LMC) \citep{chamon2024PDL} operates instead at sampling time and enforces equality and average constraints via gradient descent–ascent in Wasserstein space, without a learned generative model.

Our method shares the Lagrangian saddle-point structure of these works but differs in three aspects. (i) The dual search takes place within the reverse diffusion itself. Each denoising step is followed by a dual ascent step that evaluates the constraints at Tweedie estimates computed from the same score evaluation. In contrast, dual training performs this search in a separate outer loop that leaves a frozen sampler \citep{khalafi2024constrained, khalafi2025composition}, and PD-LMC performs it along a Langevin chain whose accuracy depends on mixing \citep{chamon2024PDL}. (ii) A single score network conditioned on the multiplier and the problem instance represents the entire Gibbs family. It is trained on Monte Carlo score targets computed from the energy alone, at the multipliers visited by PDI rollouts. An unconditional score model trained against the evolving dual iterates instead reflects the history of those iterates rather than the Gibbs target at its current multiplier. (iii) The target is the Lagrangian minimizer at the optimal multiplier, i.e., the Gibbs distribution $\mu^*=\mu^\dagger_{\bblambda^*}$ that tilts the reference (uniform) distribution as in the characterization of \citep{khalafi2024constrained} and in the energy-based samplers discussed above. PDI reaches this target without estimating $\bblambda^*$ beforehand. Theorem~\ref{theorem:dual-convergence} shows that the dual iterates converge to a neighborhood of $\bblambda^*$ despite the bias of the Tweedie-based dual updates. Theorem~\ref{theorem:primal-convergece} shows that the output distribution tracks that of an oracle sampler that knows $\bblambda^*$, up to a floor set by the Tweedie, reverse-sampling, and score-approximation errors. Empirically, the path-dependent dynamics also improve feasibility over freezing the multiplier at its final or time-averaged value (Section~\ref{sec:exp_wra}).

\paragraph{Diffusion-enabled Stochastic Optimization.}
Pretrained diffusion models constrained on the data manifold can be steered toward downstream objectives through gradient-based guidance or fine-tuning \citep{guo2024graidentguidance}, making them viable tools for learning-enabled optimization. Several works have exploited this connection between guidance with optimization gradients and regularized optimization. \citep{kong2025diffusionmodelsconstrainedsamplers} combined guided diffusion models and Langevin dynamics in a two-stage scheme that learns constrained samplers for optimization with unknown constraints. \citep{black2024training} reinterpreted the denoising chain as a multi-step Markov Decision Process (MDP) and applied policy-gradient methods to fine-tune diffusion models on non-differentiable rewards. Beyond continuous settings, diffusion models have also advanced combinatorial optimization \citep{sun2023difusco}, and distributionally robust optimization~\citep{wen2025distributionally, kong2025robustgreensecurity, guo2026conditional}.

\paragraph{Diffusion models in wireless optimization and portfolio management.}
Diffusion models recast stochastic optimization as conditional sampling from learned solution distributions. This perspective is especially relevant to wireless communication systems, whose non-convex, constrained design problems typically admit optimal solutions that are probability distributions over optimization variables. E.g., \citep{arvinte2023mimo, zilberstein2024jointchannelestimation, zhizhou2025scorebasedrismimo} leveraged diffusion priors for score-based MIMO channel estimation, with subsequent works extending their use to CSI compression \citep{lee2026userspecificchannel, ankireddy2025residual, kim2025generativecompressionmimocsi}, beamforming design \citep{guo2025diffusionmultipleantenna, hui2026channelaware}, and resource allocation \citep{zhang2026drl, darabi2024diffusion, uslu2025generative, uslu2026graphsignaldiffusionmodels}. Specifically for power control, \citep{darabi2024diffusion} trains DDPMs to generate optimal allocations conditioned on the channel state, while \citep{uslu2025generative, uslu2026graphsignaldiffusionmodels} proposes graph-signal diffusion models for matching primal-dual expert policies with near-optimal ergodic rates and cross-topology transferability.

Diffusion models have found great use in quantitative finance applications, where non-convex constraints and high-dimensional, heavy-tailed return distributions present analogous challenges to those encountered in wireless system design. E.g., \citep{takahashi2025generation} proposed wavelet-DDPMs for scenario and stylized facts generation. \citep{choudhary2025diffusionaugmentedrlrobust, he2026factorbased} tackled robust portfolio management with conditional diffusion models, with \citep{jin2025forecastingimpliedvolatilitysurface, tiwari2026generativediffusionmodelriskneutral} enforcing arbitrage-free and risk-neutral diffusion dynamics for volatility forecasting.

%% file: sections/appendix_b_proposed_algorithms.tex
\section{Proposed Algorithms}\label{app:alg}

\input{sections/appendix_b1_scorenet_training}

\input{sections/appendix_b2_inference_algo}

\input{sections/appendix_b3_practical}

%% file: sections/appendix_b1_scorenet_training.tex
\subsection{Score Network Training}\label{app:training}

The dual-conditioned score model $s_{\bbtheta}(\bbx_t,t,\bblambda,\ccalG)$ is shared across problem instances $\ccalG$, multipliers $\bblambda$, and diffusion times $t$. It is trained with the regression objective
\begin{align}\label{eq:training-loss}
    \bbtheta^* ~\in~ \argmin_{\bbtheta} \quad \mbE_{(\ccalG,\bblambda,t,\bbx_t)\sim\rho} \Big[ \omega(t) \,
    \big\| s_{\bbtheta}\big(\bbx_t, t, \bblambda,\ccalG\big) - \nabla_{\bbx_t} \log q_{T-t}(\bbx_t | \bblambda,\ccalG) \big\|_2^2 \Big],
\end{align}
where $\omega(t)$ is a time-dependent weighting function and $\rho$ is a training distribution over instances, multipliers, diffusion times, and noisy samples. This objective differs from standard score matching in two ways.

\emph{Targets.} Standard diffusion models regress onto the score of the Gaussian noising kernel around noised data samples, which requires samples of the target distribution. Here, samples of $\muDagger$ are unavailable, and the regression target is the score of the noised Gibbs marginal itself. We estimate it from the energy with the Monte Carlo estimator of Lemma~\ref{lem:MC}. Given $\bbx_t$, the estimator draws Gaussian candidates for the clean sample and reweights them by the Gibbs density $e^{-E(\cdot,\bblambda)}$, and the unknown partition function cancels in the normalization.

\emph{Inputs.} Because each target is computed directly at its input, rather than through noised data samples, any training distribution $\rho$ is admissible, and $\rho$ only determines where the model is accurate. We draw $\bbx_t$ and $\bblambda$ from rollouts of PDI with the current model, which places accuracy on the states and multipliers encountered at inference. We complement them with multipliers drawn from an initial distribution and with small perturbations for coverage. The distribution of multipliers is not known in advance, since the dual trajectory itself induces it, so training alternates between collecting rollouts and fitting the model (Algorithm~\ref{alg:training}; see also the practical considerations below). Using the estimator of Lemma~\ref{lem:MC} directly in place of $s_{\bbtheta}$ in~\eqref{eq:sample_update} yields the training-free variant PDI-MC.

\begin{lemma}[MC estimator of the Gibbs score \cite{akhoundsadegh2024idem}]\label{lem:MC} For any $\bby_\tau$, the score function of the Gibbs distribution can be estimated by a \emph{biased} $K_{\text{MC}}$ Monte Carlo estimator:
    \begin{align}\label{eq:MC_approximation}
        \nabla_{\bby_\tau} \log q_{\tau}(\bby_\tau|\bblambda)  ~\approx~
                            \nabla_{\bby_\tau} \,
                                    \log \frac{1}{K_{\text{MC}}} \sum_{k=1}^{K_{\text{MC}}} \exp\Bigg( -E\Big(\,\frac{\bby_\tau + \sigma_\tau \bbepsilon_k }{\alpha_\tau},\, \bblambda\,\Big) \Bigg),
    \end{align}
    where $\bbepsilon_k\sim\ccalN(\mathbf{0}, \bbI)$.
\end{lemma}
We refer the reader to \cite{akhoundsadegh2024idem} for the proof.

\paragraph{Relation to guidance.}\label{app:pdi-guidance}
Training a dual-conditioned model, rather than guiding a model of the unconstrained optimum $\mu^\dagger_{\mathbf 0}$, is motivated by the following identity. By Proposition~\ref{lemma:strong-duality}, $\rmd\muDagger/\rmd\mu^\dagger_{\mathbf 0}=e^{-\bblambda^\top\bbf/\beta}\,Z(\mathbf 0)/Z(\bblambda)$. Hence,
\begin{align}
    q_\tau(\bbx|\bblambda)
    ~=~ \int\ccalN\big(\bbx;\alpha_\tau\bby,\sigma_\tau^2\bbI\big)\,\muDagger(\rmd\bby)
    ~=~ \frac{Z(\mathbf 0)}{Z(\bblambda)}\;q_\tau(\bbx|\mathbf 0)\;\mbE_{\mu^\dagger_{\mathbf 0}}\Big[e^{-\bblambda^\top\bbf(\bby_0)/\beta}\,\Big|\,\bby_\tau=\bbx\Big],
\end{align}
and taking $\nabla_\bbx\log$ of both sides yields the exact guidance identity
\begin{align}\label{eq:exact_guidance}
    \nabla\log q_\tau(\bbx|\bblambda) ~=~ \nabla\log q_\tau(\bbx|\mathbf 0) + \nabla_{\bbx}\log\mbE_{\mu^\dagger_{\mathbf 0}}\Big[e^{-\bblambda^\top\bbf(\bby_0)/\beta}\,\Big|\,\bby_\tau=\bbx\Big].
\end{align}
The correction is a Doob $h$-transform of the unconstrained reverse process. Guidance methods such as DPS~\citep{chung2023diffusion} replace the conditional expectation of the tilt by the tilt evaluated at the Tweedie mean $\widehat\bby_0^{(\mathbf 0)}(\bbx)$ of $\mu^\dagger_{\mathbf 0}$. This yields the correction $-\beta^{-1}\nabla_\bbx\big[\bblambda^\top\bbf\big(\widehat\bby_0^{(\mathbf 0)}(\bbx)\big)\big]$. The approximation is exact when the posterior is a point mass, but its error grows with the posterior spread, i.e., at low SNR, and with $1/\beta$, since the tilt is then sharply nonlinear. In practice, guidance also replaces $\bblambda^*/\beta$ by a hand-tuned step size and penalizes each sample separately, as in our DPS baseline~\eqref{eq:dps}. This drives every sample toward feasibility rather than enforcing the constraints on average. PDI avoids both approximations: the dual-conditioned score represents~\eqref{eq:exact_guidance} for every $\bblambda$, and dual ascent determines the weight.

\input{sections/appendix_b1_scorenet_training_table}

%% file: sections/appendix_b1_scorenet_training_table.tex
\begin{algorithm}[t]
\caption{Score Network Training}
\label{alg:training}
\begin{algorithmic}[1]
\Require Decay and noise schedule $a_{0:T}, b_{0:T}$, Perturbation params $(\rho_{\text{pert}}, \epsilon_{\bbx}, \epsilon_{\bblambda})$, exploitation fraction $\rho_{\text{exp}}(n)$ for all $n$
\State Initialize replay buffer $\mathcal{B} \leftarrow \emptyset$, score network $\bbs_{\bbtheta}$
\For{$n= 1, \ldots, N_{\mathrm{outer}}$}
    \For{each problem $\ccalG$ in a batch $\ccalB_{\text{train}}$} \Comment{Rollouts}
        \State Sample $\{\bbx_0^{(i)} \sim \mathcal{N}(\mathbf{0}, \mathbf{I})\}_i$, 
        Initialize $\bblambda \leftarrow \bblambda_0$
        \State Run inference (Algorithm~\ref{alg:inference}) with $\bbs_{\bbtheta}$
              to obtain
              $\{(\mathbf{x}_t^{(i)}, t, \bblambda, \ccalG)\}_{t=0}^{T-1}, \ \forall i \in [I]$
        % \For{$t = 0, \ldots, T-1$} 
        %     \State $\hat{\bby}_0^{(i)} \leftarrow \big(\bbx_t^{(i)} - \sigma_{T-t}\, \bbs_\theta(\bbx_t^{(i)}, t, \bblambda, \ccalG)\big) / \sqrt{\alpha_{T-t}}, \qquad i=1, \dots, I$
        %     \State $\bblambda \leftarrow \Big[\bblambda + \eta_{\mathrm{train}} \cdot \sum_{i}\bbf(\hat{\bby}_0^{(i)})/I \Big]_+$ 
        %     \State $ \bbx_{t+1}^{(i)} ~=~ \frac{1}{\sqrt{a_{T-t}}} \Big( \, \bbx_{t}^{(i)} + b_{T-t} \, \bbs_{\theta^*}(\bbx_t^{(i)}, t, \bblambda, \ccalG) \Big)
        %     + \sqrt{ b_{T-t} } \bbz,
        %     \quad \bbz \sim \mathcal{N}(\mathbf{0}, \mathbf{I})$
            \State Push $\{(\bbx_{t}^{(i)}, t, \bblambda, \ccalG)\}_{i}^I$ to $\mathcal{B}$ 
        \EndFor
    % \EndFor
    \For{inner $= 1, \ldots, N_{\mathrm{inner}}$} 
        \State Sample a minibatch $\{(\bbx_t^{(j)}, t^{(j)}, \bblambda^{(j)}, \ccalG^{(j)})\}_{j=1}^B$ from $\mathcal{B}$
        \For{$j = 1, \ldots, B$}
        \If{$\mathrm{Bernoulli}(\rho_{\text{pert}}) = 1$} \Comment{Perturb training pairs}
                \State $\bbx_t^{(j)} \leftarrow \bbx_t^{(j)} + \epsilon_{\bbx} \cdot \boldsymbol{z}_x$, \quad $\boldsymbol{z}_x \sim \mathcal{N}(\mathbf{0}, \mathbf{I})$
                \State $\bblambda^{(j)} \leftarrow [\bblambda^{(j)} + \epsilon_{\bblambda} \cdot \boldsymbol{z}_\lambda]_+$, \quad $\boldsymbol{z}_\lambda \sim \mathcal{N}(\mathbf{0}, \mathbf{I})$
            \EndIf
            \If{$\mathrm{Bernoulli}(1 - \rho_{\text{exp}}(n)) = 1$}
                \State $\bblambda^{(j)} \leftarrow \bbv, \quad \bbv \sim \pi_{\text{prior}}(\cdot)$ \Comment{Sample from a prior}
            \EndIf
        \EndFor
        \State $\bbtheta \leftarrow \bbtheta - \alpha \nabla
        \frac{1}{B}\sum_{j}
          w(t^{(j)})
          \bigl\lVert
            \bbs_{\bbtheta}(\mathbf{x}_t^{(j)}, t^{(j)}, \bblambda^{(j)}, \ccalG^{(j)})
            - \nabla \log q_{T-t^{(j)}} (\bbx_t^{(j)}\mid \bblambda^{(j)}; \ccalG^{(j)})
            \bigr\rVert_2^2$
    \EndFor
\EndFor
\State \Return $\bbtheta$
\end{algorithmic}
\end{algorithm}

%% file: sections/appendix_b2_inference_algo.tex
\subsection{Inference Algorithm}\label{app:inference}
At inference, we run multiple chains in parallel, as described in Algorithm~\ref{alg:inference}. Each chain carries one multiplier and an ensemble of samples, and the expectation in~\eqref{eq:dual_update} is replaced by the ensemble average. A single score evaluation per step drives both the primal and the dual update. In addition to the projection $[\cdot]_+$, the implementation clips each multiplier at $\lambda_{\max}$, i.e., it projects onto the box $\bbLambda=[0,\lambda_{\max}]^d$ of Proposition~\ref{lemma:strong-duality}, with $\lambda_{\max}$ reported with the hyperparameters of each experiment (Appendices~\ref{app:MoG}, \ref{app:wra}, and~\ref{app:pf}).

\paragraph{Sensitivity to the multiplier.}
By Lemma~\ref{lemma:score_lipschitz}, the noised Gibbs score is Lipschitz in the multiplier, $\|\nabla\log q_\tau(\bbx|\bblambda)-\nabla\log q_\tau(\bbx|\bblambda')\|\leq\gamma_\tau\|\bblambda-\bblambda'\|$, with $\gamma_\tau=\frac{R\,\alpha_\tau}{\beta\,\sigma_\tau^2}\sup_{\bbx,\bblambda}\|\Cov(\bby_0\,|\,\bby_\tau=\bbx,\bblambda)\|^{1/2}$. The posterior is supported on $\ccalX$, so its covariance is at most $\mathrm{diam}(\ccalX)^2$ in operator norm, and
\begin{align}
    \gamma_\tau ~\leq~ \frac{R\,\mathrm{diam}(\ccalX)}{\beta}\,\frac{\alpha_\tau}{\sigma_\tau^2}
    ~=~ \frac{R\,\mathrm{diam}(\ccalX)}{\beta}\,\frac{\sqrt{\SNR_\tau}}{\sigma_\tau}.
\end{align}
The bound vanishes as $\SNR_\tau\to0$ and grows with the SNR. In the reverse step with exact scores, a multiplier error $\bbdelta$ at step $t$ perturbs the drift by at most $b_{T-t}\gamma_{T-t}\|\bbdelta\|/\sqrt{a_{T-t}}$. Early steps are therefore insensitive to the multiplier, while the nonlinearity bias of the dual step is largest there. Schedules that keep the SNR low for longer thus give the dual iterates more steps before the samples become sensitive to the multiplier, consistent with the noise-schedule ablations in Figure~\ref{fig:ablation_wra} and Appendix~\ref{app:pf}.

\input{sections/appendix_b2_inference_algo_table}

%% file: sections/appendix_b2_inference_algo_table.tex
\begin{algorithm}[t]
\caption{PDI Dynamics for a given problem instance $\ccalG$}
\label{alg:inference}
\begin{algorithmic}[1]
\Require Trained score network $\bbs_{\theta^*}$, dual step size $\eta_t$, schedules $a_{1:T}, b_{1:T}$, initial multiplier $\bblambda_0$
\State Initialize output set $\ccalS \leftarrow \emptyset$
\For{chain $= 1, \dots, C$}
\State Sample $\bbx_0^{(i)} \sim \mathcal{N}(\mathbf{0}, \mathbf{I}), \quad i=1, \dots, I$ \Comment{ensemble sharing one multiplier}
\State Initialize $\bblambda \leftarrow \bblambda_0$
\For{$t = 0, \ldots, T-1$}
    \State $\bbs^{(i)} \leftarrow \bbs_{\theta^*}(\bbx_t^{(i)}, t, \bblambda, \ccalG)$ \Comment{one score evaluation per step}
    \State $\widehat{\bby}_0^{(i)} \leftarrow \Pi_{\ccalX}\big[(\bbx_t^{(i)} + \sigma_{T-t}^2\, \bbs^{(i)}) / \overline{\alpha}_{T-t}\big]$ \Comment{Tweedie estimate}
    \State $\bbx_{t+1}^{(i)} \leftarrow \frac{1}{\sqrt{a_{T-t}}} \big( \bbx_{t}^{(i)} + b_{T-t} \, \bbs^{(i)} \big) + \sqrt{ b_{T-t} } \, \bbz^{(i)}, \quad \bbz^{(i)} \sim \mathcal{N}(\mathbf{0}, \mathbf{I})$ \Comment{primal step}
    \State $\bblambda \leftarrow \min\big\{\big[\bblambda + \eta_t \sum_{i} \bbf(\widehat{\bby}_0^{(i)})/I\big]_+,\ \lambda_{\max}\big\}$ \Comment{dual step, clipped at $\lambda_{\max}$}
\EndFor
\State Add $\{ \bbx_T^{(i)} \}_i$ to $\ccalS$
\EndFor
\State \Return $\ccalS$
\end{algorithmic}
\end{algorithm}

%% file: sections/appendix_b3_practical.tex
\subsection{Practical Considerations}\label{app:pdi-impl}
In training the score model, we combine exploitation and exploration to generate training pairs $(\bbx,\bblambda)$. Exploitation is performed by rolling out the sampling process using the most recent score model and storing the resulting primal-dual trajectories in a replay buffer. These rollouts expose the model to the states and dual variables it is likely to encounter during inference. Exploration is introduced in two ways. First, we sample dual variables from an initial distribution to cover regions that may not be visited by the current rollouts. Second, we perturb both the primal and dual variables to improve robustness to local deviations from the replayed trajectories. The exploitation fraction $\rho_{\mathrm{exp}}(n)$ controls the balance between replayed dual variables and reference-sampled dual variables throughout training.

Although we refer to the learned network as a score model, our implementation uses the noise-prediction parameterization. For each training pair $(\bbx_t,\bblambda)$, the MC estimator computes the score of the Gibbs distribution indexed by $\bblambda$ directly. Rather than regressing the network directly to this score, we reparameterize the target into the equivalent noise via $\boldsymbol{\epsilon}_t = - \sigma_{T-t}\mathbf{s}_t $, where $\mathbf{s}$ is the MC score estimate. The network is therefore trained to predict this noise target. During sampling, the predicted noise is converted back into score units using the same relation, and the resulting score is used in the PDI reverse update. Thus, throughout the paper, training a score model should be understood as training a noise-prediction network whose target is derived from the same MC score estimate.

In some of our numerical examples, the samples must satisfy physical domain constraints, such as box constraints on transmit powers or simplex constraints on portfolio weights. For convex $\ccalX$, the exact posterior mean lies in $\ccalX$, since it averages a posterior supported on $\ccalX$. With an estimated score, projecting the Tweedie estimate $\widehat\bby_0$ onto $\ccalX$, as in Step~7 of Algorithm~\ref{alg:inference}, can therefore only reduce its distance to the exact mean. We also divide by $\overline{\alpha}_{T-t} = \max\{\alpha_{\min}, \alpha_{T-t}\}$ when computing the Tweedie estimate to avoid numerical instabilities in the low-SNR regime. Beyond this projection, the implementation also enforces $\ccalX$ inside the reverse process. In the Gaussian-mixture and wireless experiments, it clips the clean-sample estimates and the Monte Carlo candidates of Lemma~\ref{lem:MC} to the box $\ccalX$. In the portfolio experiment, it runs the diffusion on logits and maps them to the simplex by a softmax. These devices keep every sample in $\ccalX$ but alter the effective reference measure, most strongly near the boundary of $\ccalX$, and our analysis does not cover them.

Finally, we implement~\eqref{eq:sample_update} through the posterior mean of the forward process given $\bbx_t$ and the clean-sample estimate $\widehat\bby_0$, with the posterior variance in place of $b_{T-t}$~\citep{ho2020denoising}. Without clipping, this posterior-mean form has the same drift as~\eqref{eq:sample_update}.

%% file: sections/appendix_clean_proof_final.tex
\section{Analytical Proofs}

The dual function of \eqref{eq:constrained_problem} is defined as
\begin{align}
    g(\bblambda) ~=~ \ccalL(\muDagger, \bblambda) ~=~ \min_{\mu\in \ccalP(\ccalX)} \, \, \ccalL(\mu, \bblambda),
\end{align}
where $\muDagger$ is the Gibbs distribution induced by the dual variable $\bblambda$. We can then write the dual function in a closed form as
\begin{align}
    g(\bblambda) ~=~ \mbE_{\muDagger} \Big[ f_0(\bbx) + \bblambda^\top \bbf(\bbx) \Big] - \beta \ccalH(\muDagger) = -\beta \log Z(\bblambda).
\end{align}

The true gradient of the dual function is 
\begin{align}
    \nabla g(\bblambda_t) = \mbE_{\muDaggert}[\bbf(\bbx_T)],
\end{align}
which is the violation estimated by clean samples of the distribution $\muDaggert$.  Since these clean samples are not available during the denoising process, PDI replaces this gradient with the biased gradient
\begin{align}
    \widehat\nabla g(\bblambda_t) = \mbE_{\bbx_t \sim p_t} \Bigg[ \bbf \left( \Pi_{\ccalX} \big[ \xpredtheta \big] \right) \Bigg],
\end{align}
where $p_t$ is the marginal distribution at time $t$ and $\xpredtheta$ is the Tweedie posterior-mean estimate at $\bbx_t$ under the score model $\bbs_{\bbtheta}$. It is computed from the same score evaluation as the primal step. Here, $\Pi_{\ccalX}$ denotes the Euclidean projection onto a convex set $\ccalX$, as in Algorithm~\ref{alg:inference}, and guarantees $\| \widehat{\nabla} g(\bblambda_t) \| \leq R$ under Assumption~\ref{as:f_lipschitz}. The exact posterior means lie in $\ccalX$ and are left unchanged by the projection. The projection being nonexpansive, the bias bounds below hold as stated. 

Additionally, the dual updates we write are of the population form, while the implementation averages over an ensemble instead, which we discuss after introducing Lemma~\ref{lem:contraction}. This formula replaces clean samples with the Tweedie posterior means under the forward process started with $\muDaggert$. The bias in this gradient comes from four sources: i) replacing the score field with a parameterized network $\bbs_{\bbtheta}$, ii) propagating the dual mismatch and score-approximation error across diffusion steps (marginal mismatch), iii) the sampling (discretization) error of the exact-score reverse process, and iv) Tweedie's error, i.e., replacing clean samples $\bby_0$ with a posterior mean $\xpredplusone$.

More concretely, we write the difference between the estimated gradient $\widehat\nabla g$ and true gradient $\nabla g$ as
\begin{align}
     \bbb_{t}(\bblambda_t) &\coloneqq \widehat\nabla g(\bblambda_t) -  \nabla g(\bblambda_t)  \label{eq:bias}\\
    & ~=~  \mbE_{p_{t}} \Big[ \bbf \Big( \, \Pi_{\ccalX}\big[\xpredtheta\big] \, \Big) \Big]  -
            \mbE_{\muDaggert} \big[\bbf(\bby_0)\big] 
        \nonumber \\
    &~=~    
        \mbE_{p_{t}} \Big[ \bbf \Big( \, \Pi_{\ccalX}\big[\xpredtheta\big] \, \Big) \Big]
        - \mbE_{p_{t}} \Big[ \bbf \Big( \, \xpred \, \Big) \Big] \label{eq:bias_score}\\
        & \quad + 
            \mbE_{p_{t}} \Big[ \bbf \Big( \, \xpred \, \Big) \Big]
            - \mbE_{\widetilde{p}_{t}} \Big[ \bbf \Big( \, \xpredtilde \, \Big) \Big] \label{eq:bias_mismatch} \\
        & \quad + 
            \mbE_{\widetilde{p}_{t}} \Big[ \bbf \Big( \, \xpredtilde \, \Big) \Big]
            - \mbE_{q_{T-t}^{\bblambda_t}} \Big[ \bbf \Big( \, \mbE[\bby_0\,|\,\bby_{T-t},\bblambda_t] \, \Big) \Big] \label{eq:bias_disc} \\
    & \quad + 
            \mbE_{q_{T-t}^{\bblambda_t}} \Big[ \bbf \Big( \, \mbE[\bby_0\,|\,\bby_{T-t},\bblambda_t] \, \Big) \Big] 
            - \mbE_{\muDaggert} \big[\bbf(\bby_0) \big]. \label{eq:bias_Tweedie}
\end{align}
Here, we write $\widetilde{p}_{t}^{\,\bblambda}$ for the law at step $t$ of~\eqref{eq:backward_process_family} with exact scores, fixed multiplier $\bblambda$, and initialization $\mathcal{N}(\bb0, \bbI)$, and abbreviate $ \widetilde{p}_t = \widetilde{p}_{t}^{\,\bblambda}$. The measure $q_{T-t}^{\bblambda_t}$ is the corresponding forward marginal. All posterior means are taken under the forward process~\eqref{eq:forward_process_family} started at $\muDaggert$. The four terms are the score approximation error, the marginal mismatch caused by the changing multiplier, the sampling error of the exact-score reverse process with a fixed multiplier, and Tweedie's error. We bound them in Lemmas~\ref{lem:param}, \ref{lemma:gap_in_marginals}, \ref{lemma:bias_disc}, and~\ref{lemma:bias_in_posterior} (Section~\ref{app:bias}).

The bias in the gradient results in dual convergence with a small but irreducible floor error. Theorem~\ref{theorem:dual-convergence} characterizes this floor error and shows that it is controlled by the expected Tweedie posterior-mean error, the reverse-sampling error, and the approximation error of the parameterization. Theorem~\ref{theorem:primal-convergece} then compares the terminal law $p_T$ of PDI with that of an oracle sampler that knows $\bblambda^*$. In Sections~\ref{prrof:dual-convergence} and~\ref{proof:primal-convergence}, we provide the analytical proofs of these theorems. The assumptions under which our analysis holds are listed in Section~\ref{app:assumptions}.

Throughout the analysis, we denote by \(K_t^{\bblambda}\) the exact-score reverse kernel associated with a fixed multiplier \(\bblambda\),
$$ K_t^{\bblambda}(\bbx,\cdot) := \mathcal N\!\left( \frac{1}{\sqrt{a_{T-t}}} \Big(\bbx+b_{T-t}\bbs_t(\bbx\mid\bblambda)\Big), b_{T-t}\bbI \right), $$
where \(\bbs_t(\bbx\mid\bblambda)=\nabla_{\bbx}\log q_{T-t}(\bbx\mid\bblambda)\). We denote by \(K_{\bbtheta,t}^{\bblambda}\) the corresponding parameterized reverse kernel obtained by replacing \(\bbs_t\) with \(\bbs_{\bbtheta}\),
$$ K_{\bbtheta,t}^{\bblambda}(\bbx,\cdot) := \mathcal N\!\left( \frac{1}{\sqrt{a_{T-t}}} \Big(\bbx+b_{T-t}\bbs_{\bbtheta}(\bbx,t,\bblambda)\Big), b_{T-t}\bbI \right). $$
For any distribution \(p\), we denote by \(pK_t^{\bblambda}\) and \(pK_{\bbtheta,t}^{\bblambda}\) the distributions obtained after applying the corresponding one-step kernels to \(\bbx_t\sim p\).

\subsection{Assumptions}\label{app:assumptions}

Our theoretical results hold under the following set of assumptions:

\begin{assumption}[Existence of a strictly feasible solution]\label{as:feasibility} 
There exists $\mu' \in \ccalP(\ccalX)$ such that $\mbE_{\mu'} [ f_0(\bbx)] - \beta \,\ccalH(\mu') < C < \infty$, and $\mbE_{\mu'} [ \bbf(\bbx)] \preceq - \xi \mathbf{1} \prec \mathbf{0}$.
\end{assumption}

\begin{assumption}[Constraint regularity]\label{as:f_lipschitz}
    The constraint function $\bbf$ is $L_f$-Lipschitz, i.e., $\|\bbf(\bbx) - \bbf(\bby)\| \leq L_f \, \|\bbx-\bby\|$. Also, the constraint function is bounded on the compact set $\ccalX$, i.e., $\sup_{\bbx \in \ccalX} \|\bbf(\bbx)\| \leq R$.
\end{assumption}

\begin{assumption}[Strong dual concavity]\label{as:dual_strong_concave}
    The dual function $g$ is $\vartheta$-strongly concave on $\bbLambda$, i.e., $\nabla^2_{\bblambda} g(\bblambda)\preceq -\vartheta\bbI$ for all $\bblambda\in\bbLambda$, where $\bbLambda=[0,\lambda_{\max}]^d$ is the box of Proposition~\ref{lemma:strong-duality}, which contains $\bblambda^\star$ and onto which the dual iterates are projected (Appendix~\ref{app:inference}).
\end{assumption}
Since $\nabla^2 g(\bblambda)=-\frac{1}{\beta}\textbf{Cov}_{\muDagger}(\bbf(\bbx))$, the modulus is $\vartheta=\frac{1}{\beta}\min_{\bblambda\in\bbLambda}\sigma_{\min}\big(\textbf{Cov}_{\muDagger}(\bbf(\bbx))\big)$, where $\sigma_{\min}$ denotes the smallest eigenvalue. The Gibbs distributions $\muDagger$ all share the null sets of the uniform distribution on $\ccalX$. The covariance is therefore nonsingular for every $\bblambda$ as soon as no nonzero linear combination of the constraints is almost everywhere constant on $\ccalX$, i.e., as soon as no constraint is redundant. Since the covariance is continuous in $\bblambda$ and $\bbLambda$ is compact, $\vartheta>0$ then follows. Strong concavity must hold along the segments joining the iterates to $\bblambda^\star$, which is why we require it on the convex set $\bbLambda$ that contains all iterates rather than only near $\bblambda^\star$. A uniform modulus generally does not exist on all of $\reals^d_+$, because the constraint covariance can vanish as the multipliers grow.

\begin{assumption}[Rich parameterization]\label{as:rich}
    For any score function $\bbs_t = \nabla \log q_t$, there exists a parameterization $\bbtheta$ such that, uniformly over $\bblambda \in \bbLambda$ and $t$, it holds that
    \begin{align}
         \sup_t \, \sup_{\bblambda \in \bbLambda}\, \mbE_{p_{t},\ccalG} \Big[ \big\|  \bbs_{\bbtheta}(\bbx_{t}, t, \bblambda, \ccalG) - \bbs_t(\bbx_{t} | \bblambda; \ccalG) \big\|_2^2 \Big] \leq \varepsilon^2.
    \end{align}
\end{assumption}
Assumption 4 is an approximation condition on the score network. We require it uniformly over the compact set $\bbLambda$ in which the dual iterates evolve and over time.
We emphasize that this is an expected squared-error condition under the
inference distribution, rather than a pointwise uniform bound over $\bbx$.
The error $\varepsilon$ aggregates three sources: i) the finite expressivity of the parameterization, ii) the variance of the Monte Carlo score estimator (Lemma~\ref{lem:MC}) used to form the regression targets, and iii) the mismatch between the rollout-induced training distribution of $(\bbx_t, \bblambda)$ pairs and the distribution encountered at inference. The expressivity of the parameterization controls the first source, and our training procedure is designed to control the latter two. That is, the exploration/exploitation rollout scheme with a replay buffer reduces the distribution shift, and $K_{\text{MC}}$ candidates reduce the estimator variance. Our numerical experiments show that the difference between PDI with a trained score model (PDI--Net) and PDI with an MC score field (PDI--MC) is negligible. In some cases, PDI--Net improves PDI--MC on feasibility and tail-rate metrics. 

\begin{assumption}\label{as:score-lipschitz}
    Assume the noised optimal score is spatially Lipschitz, i.e. $\nabla^2_{\bbx} \log q_{T-t}(\bbx|\bblambda)$ exists and is bounded in operator norm, $\sup_{\bbx\in \reals^p} \|\nabla^2_{\bbx} \log q_{T-t}(\bbx|\bblambda)\|_{\emph{op}} \leq M_t$, for each t and $\bblambda \in \bbLambda$. 
\end{assumption}

\begin{assumption}[VP noise schedule] \label{as:VP}
    The diffusion process follows a variance-preserving schedule with
     $a_\tau=1-b_\tau, \, 0<b_\tau\le \frac{c_b}{T}, $
    for all \(\tau=1,\ldots,T\), where \(c_b>0\) is independent of \(T\).
    For the exact reverse kernel at inference step $s$, define the worst-case
one-step stability factor
\begin{align}
    \rho_s
    :=
    \sup_{\bblambda\in\Lambda}
    \frac{1}{\sqrt{a_{T-s}}}
    \sup_{\bbx}
    \left\|
        \bbI
        +
        b_{T-s}
        \nabla_{\bbx}^2
        \log q_{T-s}(\bbx\mid\bblambda)
    \right\|_{\rm op},
\end{align}
and let
$
    \Psi_{j,t}
    :=
    \prod_{s=j+1}^{t}\rho_s,
    \,
    0\le j\le t,
$
with $\Psi_{t,t}=1$. For the chosen \(T\)-step reverse process, define
% We assume that the accumulated stability of the
% reverse process is uniformly bounded,
\begin{align}
    C_\Psi(T)
    :=
    \sup_{t\leq T}
    \sum_{j=0}^{t-1}\Psi_{j,t-1}\,(t-j)
    <\infty,
\end{align}
and assume \(C_\Psi(T)\) is bounded.
\end{assumption}
The two parts of Assumption~6 play different roles. The condition
$b_\tau\le c_b/T$ is a refinement condition on the VP discretization. As the
number of reverse steps increases, the perturbation introduced by each
individual transition decreases proportionally to $1/T$. The second
condition controls the accumulation of errors along the reverse trajectory.
Here $\Psi_{j,t}$ is the product of the one-step stability factors between
steps $j$ and $t$, so $C_\Psi<\infty$ requires the total propagated influence
of past perturbations to remain bounded. The weights $t-j$ account for the age of a perturbation, since the multipliers of steps $j$ and $t$ differ by at most $(t-j)\eta R$.

\begin{assumption}[Dual sensitivity of the score model]\label{as:lambda-lipschitz}
    The score model is $L$-Lipschitz in the multiplier, i.e., $\|\bbs_{\bbtheta}(\bbx,t,\bblambda)-\bbs_{\bbtheta}(\bbx,t,\bblambda')\|\leq L\,\|\bblambda-\bblambda'\|$ for all $\bbx\in\reals^p$, all $t$, and all $\bblambda,\bblambda'\in\bbLambda$.
\end{assumption}
Assumption~\ref{as:lambda-lipschitz} concerns the trained network, which PDI evaluates at $\bblambda_t$ rather than $\bblambda^\star$. 
% It is a property of the network that can be estimated on rollouts or enforced by the architecture, and 
Note that it does not follow from Assumption~\ref{as:rich}, which controls the score error in value only.

\subsection{Gradient Bias}\label{app:bias}
In this subsection, we analyze the four sources of bias in the dual gradients, in the following order: the Tweedie error~\eqref{eq:bias_Tweedie}, the reverse-sampling error~\eqref{eq:bias_disc}, marginal mismatch error~\eqref{eq:bias_mismatch}, and the score approximation error~\eqref{eq:bias_score}.

\subsubsection{Tweedie's Error}
\begin{lemma}\label{lemma:bias_in_posterior}
    Under Assumption~\ref{as:f_lipschitz}, it holds that
    \begin{align}
        \Big\| \, \mbE_{q_{T-t}^{\bblambda_t}} \Big[ \bbf \Big( \, \mbE[\bby_0\,|\,\bby_{T-t},\bblambda_t] \, \Big) \Big] - 
                \mbE_{\muDaggert} \big[\bbf(\bby_0) \big] \, \Big\|_2^2  ~\leq~
                                                                        L_f^2 \,  \chi^2_{t},
    \end{align}
    where $\chi^2_{t} := \sup_{\bblambda\in \bbLambda} \mbE \big[\| \mbE[\bby_0 \mid \bby_{T-t}, \bblambda] - \bby_0\|_2^2\big]$ is the uniform expected Tweedie posterior-mean error under the forward process~\eqref{eq:forward_process_family} started at $\muDagger$.
\end{lemma}

\begin{proof} 
Let $\Pi_{t}^{\bblambda_t}$ be the joint law of $(\bby_0,\bby_{T-t})$ under the forward process started at $\muDaggert$. Its marginals are $\muDaggert$ and $q_{T-t}^{\bblambda_t}$, and the conditional law of $\bby_0$ given $\bby_{T-t}=\bbx$ is the posterior $\pi_t(\cdot\,|\,\bbx,\bblambda_t)$. We can then write the bias as 
    \begin{align}
        b_1(\bblambda_t) &~\coloneqq~ 
                                    \mbE_{q_{T-t}^{\bblambda_t}} \Big[ \bbf \Big( \, \mbE[\bby_0\,|\,\bby_{T-t},\bblambda_t] \, \Big) \Big] - \mbE_{\muDaggert} \big[\bbf(\bby_0) \big] \\
                        & ~=~ \mbE_{\Pi_{t}^{\bblambda_t}} \Big[ \bbf \Big( \, \mbE[\bby_0\,|\,\bby_{T-t},\bblambda_t] \, \Big) - \bbf(\bby_0) \Big].
    \end{align}  
Taking the norm and applying Jensen's inequality and Assumption \ref{as:f_lipschitz}, we get
    \begin{align}
        \|b_1(\bblambda_t)\|_2^2 &~\leq~ 
                                    \mbE_{\Pi_{t}^{\bblambda_t}} \Big[ \big\| \bbf \Big( \, \mbE[\bby_0\,|\,\bby_{T-t},\bblambda_t] \, \Big) - \bbf(\bby_0) \big\|_2^2 \Big] \\
                                &~\leq~
                                    L_f^2 \,  \mbE_{\Pi_{t}^{\bblambda_t}} \Big[ \big\| \mbE[\bby_0\,|\,\bby_{T-t},\bblambda_t] -\bby_0 \big\|_2^2 \Big]
                                ~\leq~ L_f^2 \, \chi^2_{t}.
    \end{align}  
For convex $\ccalX$, the posterior mean lies in $\ccalX$, so $\bbf$ is evaluated on its domain.
\end{proof}

\subsubsection{Reverse-Sampling Error}
\begin{lemma}\label{lemma:bias_disc}
    Under Assumption~\ref{as:f_lipschitz}, it holds that
    \begin{align}
        \Big\| \, \mbE_{\widetilde{p}_{t}} \Big[ \bbf \Big( \, \xpredtilde \, \Big) \Big]
            - \mbE_{q_{T-t}^{\bblambda_t}} \Big[ \bbf \Big( \, \mbE[\bby_0\,|\,\bby_{T-t},\bblambda_t] \, \Big) \Big] \, \Big\|_2^2
        ~\leq~ 4R^2\,\delta_t^2,
    \end{align}
    where $\delta_t\coloneqq\sup_{\bblambda\in\bbLambda}\mathrm{TV}\big(\widetilde p_t^{\,\bblambda},q_{T-t}^{\bblambda}\big)$ and $\widetilde p_t^{\,\bblambda}$ is the law at step $t$ of the reverse process~\eqref{eq:backward_process_family} with exact scores and multiplier $\bblambda$.
\end{lemma}
\begin{proof}
Both expectations integrate the function $\bbh(\bbx)=\bbf\big(\mbE[\bby_0\,|\,\bby_{T-t}=\bbx,\bblambda_t]\big)$. For convex $\ccalX$, the posterior mean lies in $\ccalX$, so $\|\bbh\|\leq R$ by Assumption~\ref{as:f_lipschitz}. Hence, the difference of the two expectations has norm at most $2R\,\mathrm{TV}(\widetilde p_t,q_{T-t}^{\bblambda_t})\leq 2R\,\delta_t$.
\end{proof}
The quantity $\delta_t$ is the reverse-sampling error, i.e., the error of an exact-score diffusion sampler at an intermediate step. With exact scores, it arises from both the discrete Gaussian reverse steps and the Gaussian initialization. It vanishes for an exact time reversal, and bounds of the type in~\citep{chen2023sampling, benton2024nearly} control it under schedule conditions. It is the intermediate-step counterpart of the oracle error in Appendix~\ref{proof:primal-convergence}.

\subsubsection{Marginal Mismatch}

\begin{lemma}\label{lemma:gap_in_marginals}
    Under Assumptions~\ref{as:f_lipschitz}-- ~\ref{as:VP}, it holds that
    \begin{align}
        \Big\| \, 
                \mbE_{\widetilde{p}_{t}} \Big[ \bbf \Big( \, \xpredtilde \, \Big) \Big]  - 
                \mbE_{p_{t}} & \Big[ \bbf \Big( \, \xpred \, \Big) \Big] 
            \, \Big\|_2^2 \nonumber \\
           & ~\leq~
                  2L_h^2 C_{\mathrm{hist}}^2\eta^2R^2
        +
       \ccalO\left( \frac{\varepsilon^2}{T^2} \right). 
    \end{align}
    where $L_h := L_f\max_{0\le t<T} (1+\sigma_{T-t}^2M_{t})/ \alpha_{T-t}$.
    % and $\Psi_{s,t}=\prod_{r=s+1}^{t}\rho_r$ is the stability product and $\rho_r,\gamma_j$ are the moduli of Lemma~\ref{lem:contraction} and Lemma~\ref{lemma:score_lipschitz}, generalized to any given $\bblambda$.
\end{lemma}

\begin{proof} 
    We use synchronous coupling to compare the Tweedie posterior mean at step $t$ under two processes: 1) PDI, and 2) a reference reverse process \eqref{eq:backward_process_family} under $\bblambda_t$.
    
    Under the PDI dynamics, one reverse step is given by
    \begin{align}
        \bbx_{s+1} &~=~ 
                \frac{1}{\sqrt {a_{T-s}}} \Big( \bbx_s + b_{T-s}  \, \bbs_{\bbtheta}(\bbx_{s}, s, \bblambda_s) \Big) 
                + \sqrt {b_{T-s} } \bbepsilon_s,
     \end{align}
    where $\bbx_s \sim p_s$ while the corresponding step under \eqref{eq:backward_process_family} with fixed $\bblambda_t$ is
    \begin{align}
        \widetilde\bbx_{s+1} &~=~ 
                \frac{1}{\sqrt {a_{T-s}}} \Big( \widetilde\bbx_s + b_{T-s}  \, \bbs_s(\widetilde\bbx_{s} | \bblambda_t) \Big) 
                + \sqrt {b_{T-s}} \bbepsilon_s.
    \end{align}
    Under the synchronous coupling, the two processes share the same noise $\bbepsilon_s$ and the same initialization, i.e., $\bbx_0 = \widetilde\bbx_0$. Thus, the difference between the two noisy samples $\bbx_{s+1}$ and $\widetilde\bbx_{s+1}$ is 
    \begin{align}
        \bbx_{s+1} - \widetilde\bbx_{s+1} ~=~ 
                \frac{1}{\sqrt {a_{T-s}}} \Big( \bbx_s - \widetilde\bbx_s \Big) & + \frac{b_{T-s}}{\sqrt {a_{T-s}}}  \, 
                \Big(
                \bbs_s(\bbx_{s} | \bblambda_s) 
                - \bbs_s(\widetilde\bbx_{s} | \bblambda_s) 
                \Big)\nonumber \\
                & +
                \frac{b_{T-s}}{\sqrt {a_{T-s}}}  \, 
                \Big(
                \bbs_{\bbtheta}(\bbx_{s}, s, \bblambda_s)
                - \bbs_s(\bbx_{s} | \bblambda_s)
                \Big)\nonumber \\
                & + \frac{b_{T-s}}{\sqrt {a_{T-s}}}  \, 
                \Big(
                \bbs_s(\widetilde\bbx_{s} | \bblambda_s)
                - \bbs_s(\widetilde\bbx_{s} | \bblambda_t)
                \Big), 
     \end{align}
    where we added and subtracted $ \bbs_s(\widetilde\bbx_{s} | \bblambda_s)$ and $\bbs_s(\bbx_{s} | \bblambda_s)$.
    These quantities have been analyzed in Lemmas~\ref{lem:contraction} and~\ref{lemma:score_lipschitz} for the optimal kernel $\kernel$. Here, we use generalized formulas for any kernel associated with~\eqref{eq:backward_process_family}. Similarly to~\eqref{eq:diff}--\eqref{eq:contraction}, we can bound the first term with
    \begin{align}
        \Big\| \frac{1}{\sqrt {a_{T-s}}} \Big( \bbx_s - \widetilde\bbx_s \Big) + \frac{b_{T-s}}{\sqrt {a_{T-s}}}  \, 
                \Big(
                \bbs_s(\bbx_{s} | \bblambda_s) 
                - \bbs_s(\widetilde\bbx_{s} | \bblambda_s) \Big)\Big\|_2 ~\leq~ \rho_s(\bblambda_s) \cdot
                \|\bbx_s - \widetilde\bbx_s\|_2,
    \end{align}
    where $\rho_s(\bblambda_s)= \frac{1}{\sqrt{a_{T-s}}} \, \sup_\bbx \,  \left\| \bbI + b_{T-s} \nabla_\bbx^2 \log  q_{T-s}(\bbx|\bblambda_s) \right\|_{\text{op}}$ is the stability factor of the one-step kernel $K_s^{\bblambda_s}$. For the last term, we follow~\eqref{eq:score_diff_lambda}--\eqref{eq:score_lipshitz_result} to get 
    \begin{align}
        \| \bbs_s(\widetilde\bbx_{s} | \bblambda_s)
                - \bbs_s(\widetilde\bbx_{s} | \bblambda_t) \|_2 
                ~\leq~ \gamma_s \cdot
                \|\bblambda_s - \bblambda_t \|_2,
    \end{align}
    where $\gamma_s = \frac{R \alpha_{T-s}}{\beta \sigma^2_{T-s}} \sqrt{ \sup_{\bblambda \in \bbLambda} \|\textbf{Cov}_{\pi_s} (\bby_0|\widetilde\bbx_s, \bblambda) \|_{\text{op}}}$. 

    Combining the two results, along with Assumption~\ref{as:rich}, we obtain   
    \begin{align}
        \|\bbx_{s+1} - \widetilde\bbx_{s+1}\|_2 ~\leq~ 
                        &     \rho_s(\bblambda_s) \cdot
                            \|\bbx_s - \widetilde\bbx_s\|_2 \nonumber \\
                        &+
                            \frac{b_{T-s}}{\sqrt {a_{T-s}}}  
                            \Big( \gamma_s 
                \|\bblambda_s - \bblambda_t \|_2 
                + 
                \|
                \bbs_{\bbtheta}(\bbx_{s}, s, \bblambda_s)
                - \bbs_s(\bbx_{s} | \bblambda_s)
                \|_2
                \Big).
    \end{align}
    Then, we take expectation with respect to the coupling $\Pi_{s+1}$ to get
    \begin{align}
        \mbE_{\Pi_{s+1}} \Big[\|\bbx_{s+1} - \widetilde\bbx_{s+1}\|_2 \Big] ~\leq~ 
                    &         \rho_s(\bblambda_s) \cdot
                            \mbE_{\Pi_{s}} \Big[\|\bbx_s - \widetilde\bbx_s\|_2\Big] \nonumber\\
                    & +
                            \frac{b_{T-s}}{\sqrt {a_{T-s}}}  
                            \Big(\gamma_s \cdot
                            \|\bblambda_s - \bblambda_t \|_2 
                            + \mbE_{p_s} \|
                            \bbs_{\bbtheta}(\bbx_{s}, s, \bblambda_s)
                            - \bbs_s(\bbx_{s} | \bblambda_s)
                            \|_2
                            \Big) \nonumber\\
                ~\leq~
                    &
                             \rho_s(\bblambda_s) \cdot
                            \mbE_{\Pi_{s}} \Big[\|\bbx_s - \widetilde\bbx_s\|_2\Big]
                            + \frac{b_{T-s}}{\sqrt {a_{T-s}}}  
                            \Big(\gamma_s \cdot
                            \|\bblambda_s - \bblambda_t \|_2 
                            + \varepsilon\Big).
    \end{align}
    The last inequality follows from Assumption~\ref{as:rich} and Jensen's inequality.
    With this inequality in place, let $\bbh(\bbx_{t})~\coloneqq~ \bbf \Big(\xpred\Big)$, and bound the bias with
    \begin{align}
        \|b_2(\bblambda_t)\|_2 &~\coloneqq~ 
                                \Big\| \, \mbE_{\widetilde{p}_{t}} \big[ \bbh(\bbxt) \big] - 
                                \mbE_{p_{t}} \big[ \bbh (\bbx_{t}) \big] \, \Big\|_2 \nonumber\\
                            &~=~ \Big\| \, \mbE_{\Pi_{t}} \big[ \bbh(\bbxt)  - 
                                 \bbh (\bbx_{t}) \big] \, \Big\|_2 \\
                            &~\leq~
                                L_h \mbE_{\Pi_{t}}\| \bbxt - \bbx_{t} \|_2 \\
                            &~\leq~
                                L_h \Big( \rho_{t-1}(\bblambda_{t-1}) \cdot \mbE_{\Pi_{t-1}}\big[\| \bbx_{t-1} - \widetilde\bbx_{t-1} \|_2\big] + \frac{b_{T-t+1}}{\sqrt {a_{T-t+1}}} 
                                \big(\gamma_{t-1} \cdot \|\bblambda_{t-1} -\bblambda_t\|_2+\varepsilon\big) \Big),
    \end{align}
    where $L_h$ is the Lipschitz constant of $\bbh(\xpred)$ and depends on $L_f$ and the score function.

    Recursively, we get the bound:
    \begin{align}
        \|b_2(\bblambda_t)\|_2 &~\leq~
                    L_h \sum_{j=0}^{t-1} \Bigg(  \prod_{s=j+1}^{t-1} \rho_{s}(\bblambda_s) \Bigg) \cdot \frac{b_{T-j}}{\sqrt {a_{T-j}}}  
                    \Big(\gamma_j \|\bblambda_j - \bblambda_t\|_2 + \varepsilon\Big)\nonumber \\
                    & ~=~ 
                    L_h \sum_{j=0}^{t-1} \Psi_{j, t-1} \frac{b_{T-j}}{\sqrt {a_{T-j}}} \Big(\gamma_j \|\bblambda_j - \bblambda_t\|_2+\varepsilon\Big),
    \end{align}
    where $\Psi_{j,t}$ is the stability product of Assumption~\ref{as:VP} and $\rho_s = \sup_{\bblambda \in \bbLambda} \rho_s(\bblambda)$.

    For a variance-preserving noise schedule (Assumption~\ref{as:VP}), we further have
    \begin{align}
    \frac{b_{T-j}}{\sqrt{a_{T-j}}}\gamma_j
    &=
    \frac{R}{\beta}
    \frac{b_{T-j}\alpha_{T-j-1}}{\sigma_{T-j}^2}
    \sqrt{
    \sup_{\bbx,\bblambda\in\Lambda}
    \left\|
    \operatorname{Cov}_{\pi_j}
    (\bby_0\mid\bbx,\bblambda)
    \right\|_{\mathrm{op}}
    } \\
    &\le
    \frac{R}{\beta}
    \sqrt{
    \sup_{\bbx,\bblambda\in\Lambda}
    \left\|
    \operatorname{Cov}_{\pi_j}
    (\bby_0\mid\bbx,\bblambda)
    \right\|_{\mathrm{op}}
    }.
    \end{align}
    We also have
    \begin{align}\label{eq:_hist_2}
        \|\bblambda_j - \bblambda_t\|_2 \leq (t-j) \eta R,
    \end{align}
    Hence,
    \begin{align}
        C_{\text{hist}}  ~:=~\sum_{j=0}^{t-1} \Psi_{j, t-1} \frac{b_{T-j}}{\sqrt {a_{T-j}}} \gamma_j (t-j)
        ~\leq~ 
        \frac{R\,\mathrm{diam}(\ccalX)}{2\beta}\, C_\Psi
    < \infty,
    \end{align}
    since the posterior is supported on $\ccalX$, so that its covariance has operator norm at most $\mathrm{diam}(\ccalX)^2/4$, and by Assumption~\ref{as:VP}. 
    Therefore, we have $\|b_2(\bblambda_t)\|_2^2 \leq 2 L_h^2 C_{\text{hist}}^2 \eta^2 R^2 + 2 L_h^2 H_t^2$, using Young's inequality.
    Under the VP noise schedule we also have $\frac{b_{T-j}}{\sqrt{a_{T-j}}}
\le
\frac{c_b}{T\sqrt{1-c_b/T}}$. Therefore,  
    \begin{align}
        H_t ~:=~ 
        \varepsilon \sum_{j=0}^{t-1} \Psi_{j, t-1} \frac{b_{T-j}}{\sqrt {a_{T-j}}} 
        &\le
        \frac{c_b\varepsilon}{T\sqrt{1-c_b/T}}
        \sum_{j=0}^{t-1}\Psi_{j,t-1} ~\leq~ \frac{c_b\,C_\Psi\,\varepsilon}{T\sqrt{1-c_b/T}} = \ccalO\left(\frac{\varepsilon}{T}\right).
    \end{align}
    
    Moreover, the marginal-mismatch bias satisfies
    \begin{align}
        \|b_2(\bblambda_t)\|_2^2
        \le
        2L_h^2 C_{\mathrm{hist}}^2\eta^2R^2
        +
       \ccalO\left( \frac{\epsilon^2}{T^2} \right),
    \end{align}
    which completes the proof.
    \end{proof}

\subsubsection{Parameterization Error}
\begin{lemma}\label{lem:param}
    Under Assumptions~\ref{as:f_lipschitz}, ~\ref{as:rich} and~\ref{as:VP}, it holds that
    \begin{align}
        \Big\|\mbE_{p_{t}} \Big[ \bbf \big( \, \Pi_{\ccalX}\big[\xpredtheta\big] \, \big)
            - \ \bbf \big( \, \xpred \, \big) \Big]\Big\|_2^2
            ~\leq~
            L_f^2 \frac{\sigma_{T-t}^4}{\alpha_{T-t}^2} \varepsilon^2.
    \end{align}    
\end{lemma}
\begin{proof}
    The Tweedie posterior means are given as
    \begin{align}
        \xpredtheta &~=~ \frac{1}{\alpha_{T-t}} \Big( 
                                    \bbx_{t} + \sigma_{T-t}^2 \bbs_{\bbtheta}(\bbx_{t}, t, \bblambda_t)
                                    \Big),\\
        \xpred      &~=~ \frac{1}{\alpha_{T-t}} \Big( 
                                    \bbx_{t} + \sigma_{T-t}^2 \bbs_{t}(\bbx_{t} | \bblambda_t)
                                    \Big),
    \end{align}
    where the second expression is the exact posterior mean by Tweedie's formula. The implementation divides by $\overline{\alpha}_{T-t} = \max\{\alpha_{\min}, \alpha_{T-t}\}$ instead, which only affects the lowest-SNR steps (Appendix~\ref{app:pdi-impl}).
    Therefore, we have
    \begin{align}
        \Big\|\mbE_{p_{t}} \Big[ \bbf \big( \, & \Pi_{\ccalX}\big[\xpredtheta\big] \, \big)
            - \ \bbf \big( \, \xpred \, \big) \Big]\Big\|_2 \nonumber\\
            &~\leq~
                    L_f \frac{\sigma^2_{T-t}}{\alpha_{T-t}}
                    \mbE_{p_{t}}\Big\|\bbs_{\bbtheta}(\bbx_{t}, t, \bblambda_t)
                    - \bbs_{t}(\bbx_{t} | \bblambda_t) \Big\|_2 \nonumber \\
            &~\leq~
                    L_f \frac{\sigma^2_{T-t}}{\alpha_{T-t}} \varepsilon,
    \end{align}
    by the nonexpansiveness of $\Pi_{\ccalX}$ (the exact posterior mean lies in $\ccalX$), Jensen's inequality, and Assumptions~\ref{as:f_lipschitz} and~\ref{as:rich}, which completes the proof. 
\end{proof}

\subsection{Dual Convergence}
We analyze how the bias in the PDI dual update affects convergence. Lemma~\ref{lemma:dual_recursion} establishes a one-step recursion for the distance to the optimal multiplier in terms of the gradient bias \(\bbb_t(\bblambda_t)\), which we then bound to obtain the last-iterate guarantee in Theorem~\ref{theorem:dual-convergence}. We additionally derive a time-average dual bound, which is used in the proof of Theorem~\ref{theorem:primal-convergece}.

\begin{lemma}\label{lemma:dual_recursion}
Under Assumptions~\ref{as:f_lipschitz} and~\ref{as:dual_strong_concave}, and with $\eta_t=\eta$, the distance between $\bblambda_t$ and the optimum satisfies
    \begin{align}
        \|\bblambda_{t+1} - \bblambda^\star\|^2
        + \eta \big[D^\star - & g(\bblambda_t)  \big]
        + 2\eta\beta  \KL(\muDaggert \| \mu^\star) \nonumber
        \\
        &~\leq~
                 \|\bblambda_{t} - \bblambda^\star\|^2 
               % - 2\eta \beta \KL(\muDagger \| \mu^\star)
               + \frac{2 \eta}{\vartheta}  \| \bbb_t(\bblambda_t) \|^2
               + \eta^2 R^2,
    \end{align}
    or equivalently, by
    \begin{align}
         \|\bblambda_{t+1} - \bblambda^\star\|^2 
               & + 2\eta\beta \KL(\muDaggert \| \mu^\star) \nonumber\\
                   &  ~\leq~
                   \left( 1-\frac{\eta \vartheta}{2} \right) \|\bblambda_{t} - \bblambda^\star\|^2 
                   +\frac{2\eta}{\vartheta}  \|\bbb_t(\bblambda_t)\|^2  
                   +\eta^2 R^2.
    \end{align}
\end{lemma}

\begin{proof}
    We analyze the update as implemented, with the multipliers projected onto the box $\bbLambda\ni\bblambda^\star$ of Assumption~\ref{as:dual_strong_concave}, i.e., $\bblambda_{t+1}=\Pi_{\bbLambda}[\bblambda_t+\eta\widehat\nabla g(\bblambda_t)]$. This is the projection $[\cdot]_+$ followed by clipping at $\lambda_{\max}$. By the non-expansiveness of $\Pi_{\bbLambda}$, we have $\|\Pi_{\bbLambda}[\bblambda] - \bblambda^\star\| \leq \|\bblambda - \bblambda^\star\|$, and therefore,
    \begin{align}
         \|\bblambda_{t+1} - \bblambda^\star\|^2 \,  &~\leq~
                            \|\bblambda_{t} + \eta \widehat\nabla g(\bblambda_t) - \bblambda^\star\|^2,
    \end{align}
    where $\widehat\nabla g(\bblambda_t):= \mbE_{\bbx_{t} \sim p_{t}} \Big[ \bbf \Big( \Pi_{\ccalX}\big[\xpredtheta\big] \Big) \Big]$ is the biased gradient in~\eqref{eq:dual_update}.
    
    Define the true gradient of the dual function as $$\nabla g(\bblambda_t) = \mbE_{\muDaggert}[\bbf(\bby_0)],$$ which is the constraint violations evaluated on clean samples under the current Gibbs distribution $\muDaggert$. PDI replaces this true gradient with the biased gradient $\widehat\nabla g(\bblambda_t)$, Thus, we can write 
    \begin{align}\label{eq:main_triangle_inequality}
         \|\bblambda_{t+1} - \bblambda^\star\|^2 \,  &~\leq~
                            \|\bblambda_{t} + \eta \widehat\nabla g(\bblambda_t) - \bblambda^\star\|^2 \\
                           &~\leq~
                           \|\bblambda_{t} - \bblambda^\star\|^2 
                           +2 \eta \langle
                           \bblambda_{t} - \bblambda^\star, \widehat\nabla g(\bblambda_t)
                           \rangle
                           +\eta^2 \|\widehat\nabla g(\bblambda_t) \|^2 \\
                           &~=~
                           \|\bblambda_{t} - \bblambda^\star\|^2 
                           +2 \eta \langle
                           \bblambda_{t} - \bblambda^\star, \nabla g(\bblambda_t)
                           \rangle 
                           +2 \eta  \langle
                           \bblambda_{t} - \bblambda^\star, \bbb_t(\bblambda_t)
                           \rangle   
                           +\eta^2 \|\widehat\nabla g(\bblambda_t) \|^2,
    \end{align}
    where $\widehat\nabla g(\bblambda_t) = \nabla g(\bblambda_t) + \bbb_t(\bblambda_t)$ and the bias $\bbb_t(\bblambda_t)$ is split into three terms, as characterized in \eqref{eq:bias}.
    % where $\bbb_t(\bblambda_t) = \sum_{i=1}^3\bbb_t^{(i)}(\bblambda_t)$ accounts for three bias sources (see Section~\ref{app:bias}).
    
    By Lemma~\ref{lem:Lagrangian}, the Lagrangian function can be written as 
    $\ccalL(\mu, \bblambda) ~=~ \beta \KL(\mu \| \muDagger) + g(\bblambda),$ 
    which leads to 
    \begin{align}\label{eq:inner_product_grad}
        \langle \bblambda_{t} - \bblambda^\star, \nabla g(\bblambda_t)
                \rangle 
                &~=~
                \ccalL(\muDaggert, \bblambda_t) - \ccalL(\muDaggert, \bblambda^\star) \nonumber\\
                &~=~ \underbrace{g(\bblambda_t) - g(\bblambda^\star)}_{\text{duality gap}} 
                -\beta \KL(\muDaggert \| \mu^\star).
    \end{align}
    The first equality follows from the definition in \eqref{eq:Lagrangian} and the second follows from the definition in the lemma.
    Using Young's inequality, for any $c>0$, we can bound the inner product between $\bblambda_t - \bblambda^\star$ and the bias by
    \begin{align}\label{eq:inner_product_bias}
         2\langle \bblambda_{t} - \bblambda^\star,  \bbb_t
            \rangle
            ~\leq~
            c \|\bblambda_t -\bblambda^\star\|^2 + \frac{1}{c} \|\bbb_t \|^2.
    \end{align}
    Choose $c=\vartheta/2$. By substituting~\eqref{eq:inner_product_grad} and~\eqref{eq:inner_product_bias} in~\eqref{eq:main_triangle_inequality}, we get
    \begin{align}
         \|\bblambda_{t+1} - \bblambda^\star\|^2 
            +2\eta \Big( D^\star &-  g(\bblambda_t) \Big)
                + 2\eta\beta \KL(\muDaggert \| \mu^\star) \nonumber\\
                    &~\leq~
                   \left(1+\frac{\eta\vartheta}{2}\right)\|\bblambda_{t} - \bblambda^\star\|^2 
                   +\frac{2\eta}{\vartheta} \|\bbb_t(\bblambda_t)\|^2  
                   +\eta^2 \|\widehat\nabla g(\bblambda_t) \|^2,
    \end{align}
    with $D^\star := g(\bblambda^\star)$. The segment between $\bblambda_t$ and $\bblambda^\star$ lies in the convex set $\bbLambda$, where $g$ is $\vartheta$-strongly concave (Assumption~\ref{as:dual_strong_concave}), and $\bblambda^\star$ maximizes $g$ over $\reals^d_+\supseteq\bbLambda$. Hence, $D^\star - g(\bblambda_t) \geq \frac{\vartheta}{2} \| \bblambda_t -\bblambda^\star \|^2$.

    \underline{(i) The time-average bound:}
    
    We split the dual gap in the LHS into two halves and lower bound one of them with $\frac{\eta \vartheta}{2} \| \bblambda_t -\bblambda^\star \|^2$. Thus, we have
    \begin{align}\label{eq:one_step_average}
         \|\bblambda_{t+1} - \bblambda^\star\|^2 
            +\eta \big( D^\star &-  g(\bblambda_t) \big)
                + 2\eta\beta \KL(\muDaggert \| \mu^\star) \nonumber\\
                    &~\leq~
                   \|\bblambda_{t} - \bblambda^\star\|^2 
                   +\frac{2\eta}{\vartheta}  \|\bbb_t(\bblambda_t)\|^2  
                   +\eta^2 \|\widehat\nabla g(\bblambda_t) \|^2.
    \end{align}

    \underline{(ii) The final-iterate bound:}
    
    For the final-iterate result, we upper-bound the duality gap term with 
    \begin{align}\label{eq:eq:one_step_final}
         \|\bblambda_{t+1} - \bblambda^\star\|^2 
               & + 2\eta\beta \KL(\muDaggert \| \mu^\star) \nonumber\\
                   &  ~\leq~
                   \left( 1+\frac{\eta \vartheta}{2}-\eta \vartheta \right) \|\bblambda_{t} - \bblambda^\star\|^2 
                   +\frac{2\eta}{\vartheta}  \|\bbb_t(\bblambda_t)\|^2  
                   +\eta^2 \|\widehat\nabla g(\bblambda_t) \|^2.
    \end{align}
    Since the projected Tweedie estimates lie in $\ccalX$, Assumption~\ref{as:f_lipschitz} gives $\|\widehat\nabla g(\bblambda_t)\|\leq R$, which completes the proof.
\end{proof}

\textbf{Ensemble update.} The recursion is deterministic given the update direction. For the implemented ensemble update, it therefore holds pathwise, with $\widehat\nabla g(\bblambda_t)$ replaced by the ensemble average of the projected Tweedie estimates and $\bbb_t$ by its realized error. Taking expectations yields the bounds below with $\|\bbb_t\|^2$ replaced by $\mbE\|\bbb_t\|^2$. The lemmas of Appendix~\ref{app:bias} bound the population part of this error. The realized error additionally contains ensemble fluctuations.

\subsubsection{Convergence of Last-iterate Multiplier: Proof of Theorem \ref{theorem:dual-convergence}}\label{prrof:dual-convergence}    

    First, we bound the bias in the gradients. The bias comprises four components as described in \eqref{eq:bias}. The Tweedie posterior mean error is bounded by Lemma~\ref{lemma:bias_in_posterior}:
    \begin{align}
        \|b_1(\bblambda_t)\|^2  ~\leq~ L_f^2 \,\chi^2_{t}.
    \end{align}
    The marginal mismatch error is bounded by Lemma~\ref{lemma:gap_in_marginals}:
    \begin{align}
         \|b_2(\bblambda_t)\|^2 
         ~\leq~ 2L_h^2 C_{\mathrm{hist}}^2\eta^2R^2
        +
       \ccalO\left( \frac{\epsilon^2}{T^2} \right).
    \end{align}
    Finally, the parameterization error is bounded by Lemma~\ref{lem:param}:
    \begin{align}
         \|b_3(\bblambda_t)\|^2  ~\leq~ L_f^2 \, \frac{\sigma_{T-t}^4}{\alpha_{T-t}^2} \varepsilon^2
         ~\leq~ 
         \ccalO\left(\varepsilon^2\frac{(T-t)^2}{T^2}\right).
    \end{align}
    The reverse-sampling error is bounded by Lemma~\ref{lemma:bias_disc}:
    \begin{align}
         \|b_4(\bblambda_t)\|^2  ~\leq~ 4R^2\,\delta_t^2.
    \end{align}
    The total bias can then be bounded as 
    \begin{align}
         \|\bbb_{t}(\bblambda_t)\|^2 
        &~\leq~
         4   \|b_1(\bblambda_t)\|^2  
         + 4   \|b_2(\bblambda_t)\|^2 
         + 4   \|b_3(\bblambda_t)\|^2
         + 4   \|b_4(\bblambda_t)\|^2. 
    \end{align}
    We then utilize this bound to prove the two statements of the theorem.

    Choose $\eta \in (0, \frac{2}{\vartheta})$ and let $r := 1 - \frac{\eta \vartheta}{2}$. Thus, we have $0 < r < 1$.
    Starting from the dual recursion in~\eqref{eq:eq:one_step_final}, which is 
    \begin{align}
         \|\bblambda_{t+1} - \bblambda^\star\|^2 
               & + 2\eta\beta \KL(\muDaggert \| \mu^\star) \nonumber\\
                   &  ~\leq~
                   \left( 1-\frac{\eta \vartheta}{2} \right) \|\bblambda_{t} - \bblambda^\star\|^2 
                   +\frac{2\eta}{\vartheta}  \|\bbb_t(\bblambda_t)\|^2  
                   +\eta^2 R^2.
    \end{align}
    We then drop the KL term because it is non-negative, and apply recursion to get
    \begin{align}
         \|\bblambda_{t} - \bblambda^\star\|^2 
                   &  ~\leq~
                   r^t \|\bblambda_{0} - \bblambda^\star\|^2 
                   + \sum_{s<t} r^{t-1-s} \Big(
                   \frac{2\eta}{\vartheta}  \|\bbb_s(\bblambda_s)\|^2  
                   +\eta^2 R^2 \Big).
    \end{align}
    Additionally, we have 
    \begin{align}
        \sum_{s<t} r^{t-1-s} \eta^2 R^2  \leq \frac{\eta^2 R^2}{1-r} \leq \frac{2}{\vartheta} \eta R^2,
    \end{align}
    and therefore 
    \begin{align}
         \|\bblambda_{t} - \bblambda^\star\|^2 
                   &  ~\leq~
                   r^t \|\bblambda_{0} - \bblambda^\star\|^2 
                   + \frac{2\eta}{\vartheta} R^2
                   + \frac{2\eta}{\vartheta}  \sum_{s<t} r^{t-1-s} 
                    \|\bbb_s(\bblambda_s)\|^2.
    \end{align}
    This bound holds at every step $t$, and letting $t=T$ completes the proof.

    \subsubsection{Convergence of time-average multipliers}\label{app:average_dual_gap}
    The proof of Theorem~\ref{theorem:primal-convergece} relies on the average dual gap, which we bound here. For this bound, we assume that $\eta_t = \eta = \kappa/\sqrt{T}$ for all $t$.
    
    From Lemma \ref{lemma:dual_recursion}, the dual recursion is 
    \begin{align}
        \|\bblambda_{t+1} - \bblambda^\star\|^2
        + \eta \big[D^\star - & g(\bblambda_t) \big]
        + 2\eta\beta \KL(\muDaggert \| \mu^\star) \nonumber
        \\
        &~\leq~
                            \|\bblambda_{t} - \bblambda^\star\|^2
                           % - 2\eta \beta \KL(\muDagger \| \mu^\star)
                           + \frac{2 \eta}{\vartheta} \| \bbb_{t}(\bblambda_t) \|^2
                           + \eta^2 R^2.
    \end{align} 
    Using the telescope sum, we get
    \begin{align}\label{eq:uniform-telescope}
        \|\bblambda_{T} - \bblambda^\star\|^2
        + \eta \sum_{t=0}^{T-1}\big[D^\star - & g(\bblambda_t)  \big]
        + 2\eta\beta \sum_{t=0}^{T-1} \KL(\muDaggert \| \mu^\star) \nonumber
        \\
        & ~\leq~        
                            \|\bblambda_{0} - \bblambda^\star\|^2
                           + \frac{2 \eta}{\vartheta}  \sum_{t=0}^{T-1} \| \bbb_{t}(\bblambda_t) \|^2
                           + T\eta^2 R^2.
    \end{align}

    Choose $\eta =\kappa /\sqrt{T}$. The bias can be bounded by
        \begin{align}
         \|\bbb_{t}(\bblambda_t)\|^2 
        &~\leq~
        O\!\left(
         \chi_{t}^2
        +
        \delta_t^2
        +
        \frac1T
        +
        \varepsilon^2
        \frac{(T-t)^2}{T^2}
        +
        \frac{\varepsilon^2}{T^2}
        \right).
    \end{align}
    Since the left-side terms in \eqref{eq:uniform-telescope} are non-negative, each one of them is bounded above by the right-hand side of~\eqref{eq:uniform-telescope}. Thus, we get a uniform average dual gap of 
    \begin{align}
          \frac{1}{T} \sum_{t=0}^{T-1} & \big[ D^\star -  g(\bblambda_t)  \big] ~\leq~ 
                \underbrace{\frac{ \|\bblambda_{0} - \bblambda^\star\|^2}{\eta T}}_{\ccalO\big(\frac{1}{\sqrt{T}}\big)}
               + 
               \underbrace{\ccalO\left(\frac{1}{T} \sum_{t=0}^{T-1} \big(\chi_{t}^2+\delta_t^2\big) + \varepsilon^2 + \frac{1}{T}\right)}_{\text{floor error}}
               + \underbrace{\eta R^2}_{\ccalO\big(\frac{1}{\sqrt{T}}\big)}
                \label{eq:duality_gap_uniform_bound}.
    \end{align}
    The rate of convergence of both quantities is $1/\sqrt{T}$ and they converge to a neighborhood determined by the bias.
    
    Define the time-average dual variable $\bar\bblambda = \tfrac{1}{T}\sum_t \bblambda_t$. By concavity of $g$, we have
    \begin{align}
         D^\star - g\big(\bar\bblambda\big) ~\leq~ 
         D^\star - \frac{1}{T} \sum_{t=0}^{T-1} g(\bblambda_t) ~=~ \frac{1}{T} \sum_{t=0}^{T-1} \big[D^\star - g(\bblambda_t)\big],
    \end{align}
    which is bounded by \eqref{eq:duality_gap_uniform_bound}. Furthermore, by the strong concavity of $g$, we can also write
    \begin{align}
         \|\bar\bblambda - \bblambda^\star\|^2 ~\leq~ \frac{2}{\vartheta} \big[ D^\star - g(\bar\bblambda) \big].
    \end{align}

\input{sections/proofs/primal_recursion}

\subsection{Additional Lemmas}
\begin{lemma}[One-step contraction modulus]\label{lem:contraction}
    Under Assumption~\ref{as:score-lipschitz}, for any $\bbx$, $\bby$ and under the synchronous coupling (common noise $\bbepsilon_t$), the reverse-step outputs satisfy $\|\bbX^+ - \bbY^+\|^2 \leq \rho_t^2 \|\bbx - \bby\|^2$, almost surely, with 
    $$\rho_t(\bblambda^\star) = \frac{1}{\sqrt{a_{T-t}}} \, \sup_\bbx \,  \left\| \bbI + b_{T-t} \nabla_\bbx^2 \log  q_{T-t}(\bbx|\bblambda^\star) \right\|_{\emph{op}}.$$
\end{lemma}
\begin{proof}
    Under synchronous coupling, we have
    \begin{align}\label{eq:diff}
        \bbX^+ - \bbY^+ = \frac{1}{\sqrt{a_{T-t}}} \Big( \bbx - \bby + b_{T-t} \big(\nabla \log  q_{T-t}(\bbx|\bblambda^\star) - \nabla \log  q_{T-t}(\bby|\bblambda^\star) \big)\Big).
    \end{align}
    The noise cancels exactly since the two processes share the same noise. Thus, this is deterministic given $\bbx$ and $\bby$. Then by the mean value theorem along the segment from $\bby$ to $\bbx$, we get
    \begin{align}
        \nabla \log  q_{T-t}(\bbx|\bblambda^\star) - \nabla \log  q_{T-t}(\bby|\bblambda^\star) ~=~ 
            \Big(\int_0^1 \nabla^2 \log  q_{T-t}(\bby + s (\bbx-\bby)|\bblambda^\star) ds \Big)
            \cdot (\bbx-\bby).
    \end{align}
    Substitute back into \eqref{eq:diff}, we get
    \begin{align}
        \bbX^+ - \bbY^+ = \frac{1}{\sqrt{a_{T-t}}} (\bbI + b_{T-t} \widehat\bbH ) \cdot (\bbx-\bby),
    \end{align}
    where $\widehat\bbH$ is the averaged Hessian along the segment. Taking norms,
    \begin{align}
        \|\bbX^+ - \bbY^+\| &\leq \frac{1}{\sqrt{a_{T-t}}} \left\| \bbI + b_{T-t} \widehat\bbH \right\|_{\text{op}} \cdot \|\bbx-\bby\| \\ 
        &\leq \frac{1}{\sqrt{a_{T-t}}} \sup_{\bbx} \,  \left\| \bbI + b_{T-t} \nabla_\bbx^2 \log  q_{T-t}(\bbx|\bblambda^\star) \right\|_{\text{op}} \|\bbx-\bby\| \eqqcolon \rho_t \|\bbx - \bby\|. \label{eq:contraction}
    \end{align}
    The last inequality upper bounded the averaged Hessian with the supremum over the compact domain $\ccalX$.
    This completes the proof.
\end{proof}

%%%%%%%%%%%%%%%%%%%%%%%%%%%%%%%%%%%%%%%%

\begin{lemma}\label{lemma:score_lipschitz}
    Under Assumption~\ref{as:f_lipschitz}, the score function $\bbs_t(\cdot \, | \, \bblambda)$ is Lipschitz in $\bblambda$, and
    \begin{align}
        \big\| \bbs_t(\bbx | \bblambda_t) -
                            \bbs_t(\bbx | \bblambda^\star) \big\|
                            ~\leq~ \gamma_t \left\| \bblambda_t - \bblambda^\star \right\|_2,
    \end{align}
    and $\gamma_t ~=~ \frac{R \alpha_{T-t}}{\beta \sigma^2_{T-t}} \sqrt{\sup_{\bblambda, \bbx}\|\textbf{Cov}_{\pi_t} (\bby_0|\bbx, \bblambda) \|_{\text{op}}} \geq 0$.
\end{lemma}

\begin{proof}
    Using the mean value theorem, we can then bound the difference by
    \begin{align}\label{eq:score_diff_lambda}
        \big\| \bbs_t(\bbx | \bblambda_t) -
                            \bbs_t(\bbx | \bblambda^\star) \big\|
                            ~\leq~ \Big(\sup_{\bblambda}   
                                 \big\| 
                                \nabla_{\bblambda} \bbs_t(\bbx|\bblambda)
                                \big\|_{\text{op}}  \Big)
                                 \left\| \bblambda_t - \bblambda^\star \right\|_2,
    \end{align}
    where $\|\cdot\|_{\text{op}}$ is the operator norm. 
    The score function is defined through the Tweedie as
    \begin{align}
        \bbs_t(\bbx|\bblambda) = \frac{1}{\sigma^2_{T-t}} \Big( \alpha_{T-t} 
        \mbE_{\pi_t}[\bby_0 |\bbx, \bblambda] -\bbx
        \Big),
    \end{align}
    where $\pi_t$ is the posterior distribution:
    \begin{align}
        \pi_{t}(\bby_0 | \bbx, \bblambda) \propto \frac{1}{Z(\bblambda)}
        \exp \Big( - E(\bby_0, \bblambda) - \frac{1}{2\sigma^2_{T-t}}  \|\bbx - \alpha_{T-t} \bby_0 \|^2
        \Big).
    \end{align}
    The posterior depends on $\bblambda$ only through the prior. Thus, we have
    \begin{align}
        \nabla_{\bblambda} \mbE_{\pi_t}\big[\bby_0 |\bbx, \bblambda\big] ~=~
        \frac{-1}{\beta} \, \textbf{Cov}_{\pi_t}\Big(\bby_0, \bbf(\bby_0) \big| \bbx, \bblambda\Big).
    \end{align}
    Taking the operator norm, we get
    \begin{align}\label{eq:score_lipshitz_result}
        \frac{\beta \sigma^2_{T-t}}{\alpha_{T-t}} 
        \big\| \nabla_{\bblambda} \bbs_t(\bbx|\bblambda)
                \big\|_{\text{op}} 
        &~\leq~ 
            \Big( 
            \|\textbf{Cov}_{\pi_t} (\bby_0|\bbx, \bblambda) \|_{\text{op}}
            \Big)^{1/2}
            \cdot
            \Big( 
            \|\textbf{Cov}_{\pi_t} (\bbf(\bby_0)|\bbx, \bblambda) \|_{\text{op}}
            \Big)^{1/2}\\
        &~\leq~
        R \sqrt{\|\textbf{Cov}_{\pi_t} (\bby_0|\bbx, \bblambda) \|_{\text{op}}},
    \end{align}
    using Assumption \ref{as:f_lipschitz}. Let $\gamma_t = \frac{R \alpha_{T-t}}{\beta \sigma^2_{T-t}} \sqrt{ \sup_{\bblambda, \bbx} \|\textbf{Cov}_{\pi_t} (\bby_0|\bbx, \bblambda) \|_{\text{op}}}$ to complete the proof.    
\end{proof}

\begin{lemma} \label{lem:Lagrangian}
    The Lagrangian function in \eqref{eq:Lagrangian} is also equivalent to
    $$\ccalL(\mu, \bblambda) ~=~ \beta \KL(\mu \| \muDagger) + g(\bblambda).$$
\end{lemma}
\begin{proof}
    Start with the definition of KL divergence:
    \begin{align}
        \KL(\mu \|\muDaggert) &~=~ \mbE_{\mu} \Big[ \log \mu(\bbx) - \log \muDaggert(\bbx)\Big]\\
                & ~=~ \mbE_\mu [\log \mu(\bbx)] + \frac{1}{\beta} \, \mbE_\mu \Big[ f_0(\bbx) + \bblambda^\top \bbf(\bbx)
                \Big] 
                +\log Z(\bblambda) \\
                & ~=~ \frac{1}{\beta} \, \ccalL(\mu, \bblambda) + \log Z(\bblambda).
    \end{align}
    The second equality follows from the definition of $\muDaggert$ while the last one uses the definition of the Lagrangian. With the fact that $g(\bblambda) = -\beta \log Z(\bblambda)$, we complete the proof.
\end{proof}

%% file: sections/proofs/primal_recursion.tex
\subsection{Primal Convergence: Proof of Theorem \ref{theorem:primal-convergece}}\label{proof:primal-convergence}

We bound the discrepancy between the terminal distribution of PDI and that of an oracle sampler that knows $\bblambda^\star$. The oracle runs the reverse process~\eqref{eq:backward_process_family} with the exact scores and the multiplier fixed at $\bblambda^\star$, started from the same initialization as PDI. Per the notation introduced after~\eqref{eq:bias_Tweedie}, its marginals are $\widetilde{p}_{t}^{\,\bblambda}$, so that $\widetilde{p}_0^{\, \bblambda^*} = p_0 = \mathcal{N}(\bb0, \bbI)$ and 
\begin{align}
    \widetilde{p}_{t+1}^{\, \bblambda^\star} ~=~ \widetilde{p}_{t}^{\, \bblambda^\star} K_t^{\bblambda^\star}.
\end{align}
This identity holds by construction, for every step size and noise schedule, because $\widetilde{p}_{t}^{\, \bblambda^\star}$ is defined as the law of the discrete chain itself rather than as a forward marginal. Let
\begin{align}
    D_t
    :=
    D_{\mathrm{KL}}
    \left(
        p_t
        \,\middle\|\,
        \widetilde{p}_{t}^{\, \bblambda^\star}
    \right),
\end{align}
so that $D_0=0$. Under the population update \eqref{eq:dual_update}, $\bblambda_t$ is fixed at step $t$. Hence,
\begin{align}
    p_{t+1}
    ~=~
    p_tK_{\bbtheta,t}^{\bblambda_t}.
\end{align}
Define the joint measures
\begin{align}
    P_t(d\bbx,d\bby)
    &=
    p_t(d\bbx)
    K_{\bbtheta,t}^{\bblambda_t}
    (d\bby\mid\bbx),
    \\
    Q_t(d\bbx,d\bby)
    &=
    \widetilde{p}_{t}^{\, \bblambda^\star}(d\bbx)
    K_t^{\bblambda^\star}
    (d\bby\mid\bbx).
\end{align}
Their $\bby$-marginals are $p_{t+1}$ and $\widetilde{p}_{t+1}^{\, \bblambda^\star}$. By data processing and the chain rule for relative entropy,
\begin{align}
    D_{t+1}
    \le
    D_t
    +
    \mbE_{\bbx_t\sim p_t}
    \left[
        D_{\mathrm{KL}}
        \left(
            K_{\bbtheta,t}^{\bblambda_t}
            (\cdot\mid\bbx_t)
            \,\middle\|\,
            K_t^{\bblambda^\star}
            (\cdot\mid\bbx_t)
        \right)
    \right].
    \label{eq:primal_recursion}
\end{align}
The two kernels are Gaussian with covariance
$b_{T-t}\bbI$. Their mean difference is
\begin{align}
    \frac{b_{T-t}}{\sqrt{a_{T-t}}}
    \left[
        \bbs_{\bbtheta}
        (\bbx_t,t,\bblambda_t)
        -
        \bbs_t
        (\bbx_t\mid\bblambda^\star)
    \right].
\end{align}
Hence,
\begin{align}
    &D_{\mathrm{KL}}
    \left(
        K_{\bbtheta,t}^{\bblambda_t}
        (\cdot\mid\bbx_t)
        \,\middle\|\,
        K_t^{\bblambda^\star}
        (\cdot\mid\bbx_t)
    \right)
    ~=~
    \frac{b_{T-t}}{2a_{T-t}}
    \left\|
        \bbs_{\bbtheta}
        (\bbx_t,t,\bblambda_t)
        -
        \bbs_t
        (\bbx_t\mid\bblambda^\star)
    \right\|^2 .
    \label{eq:primal_kernel_kl}
\end{align}
We decompose
\begin{align}
    &\bbs_{\bbtheta}
    (\bbx_t,t,\bblambda_t)
    -
    \bbs_t
    (\bbx_t\mid\bblambda^\star)
     =
    \left[
        \bbs_{\bbtheta}
        (\bbx_t,t,\bblambda_t)
        -
        \bbs_{\bbtheta}
        (\bbx_t,t, \bblambda^\star)
    \right]
    +
    \left[
        \bbs_{\bbtheta}
        (\bbx_t,t, \bblambda^\star)
        -
        \bbs_t
        (\bbx_t\mid\bblambda^\star)
    \right].
    \label{eq:primal_score_split}
\end{align}
Using
$\|\bbu+\bbv\|^2
\le
2\|\bbu\|^2+2\|\bbv\|^2$,
Assumptions~\ref{as:rich} and~\ref{as:lambda-lipschitz}, we obtain
\begin{align}
    D_{t+1}
    \le
    D_t
    +
    \frac{b_{T-t}}{a_{T-t}}
    \left[
        \varepsilon^2
        +
        L^2
        \|\bblambda_t-\bblambda^\star\|^2
    \right].
    \label{eq:primal_one_step}
\end{align}
Summing \eqref{eq:primal_one_step} from $D_0=0$ gives
\begin{align}
    D_{\mathrm{KL}}(p_T\|\widetilde{p}_{T}^{\, \bblambda^\star})
    \le
    \sum_{t=0}^{T-1}
    \frac{b_{T-t}}{a_{T-t}}
    \left[
        \varepsilon^2
        +
        L^2
        \|\bblambda_t-\bblambda^\star\|^2
    \right].
    \label{eq:primal_terminal}
\end{align}
Under Assumption~\ref{as:VP},
\begin{align}
    \frac{b_{T-t}}{a_{T-t}}
    \le
    \frac{c_b}{T-c_b},
\end{align}
and therefore
\begin{align}
    D_{\mathrm{KL}}(p_T\|\widetilde{p}_{T}^{\, \bblambda^\star})
    \le{}&
    \frac{c_bT}{T-c_b}\varepsilon^2 +
    \frac{c_bL^2}{T-c_b}
    \sum_{t=0}^{T-1}
    \|\bblambda_t-\bblambda^\star\|^2 .
    \label{eq:primal_vp_bound}
\end{align}
Since the segment between each iterate and $\bblambda^\star$ lies in $\bbLambda$, where $g$ is $\vartheta$-strongly concave (Assumption~\ref{as:dual_strong_concave}), and $\bblambda^\star$ maximizes $g$ over $\reals^d_+$, we have $\|\bblambda_t-\bblambda^\star\|^2\leq\frac{2}{\vartheta}\big[D^\star-g(\bblambda_t)\big]$. The average dual-gap bound in~\eqref{eq:duality_gap_uniform_bound} then gives
\begin{align}
    \frac{1}{T}\sum_{t=0}^{T-1}
    \|\bblambda_t-\bblambda^\star\|_2^2
    \le
    O\!\left(\frac{1}{\sqrt{T}}\right)
    +
    O\!\left(
    \frac{1}{T}\sum_{t=0}^{T-1}\big(\chi_{t}^2+\delta_t^2\big)
    + \varepsilon^2
    + \frac{1}{T}
    \right).
\end{align}
Substituting the above bound into~\eqref{eq:primal_vp_bound} gives
\begin{align}
    D_{\mathrm{KL}}(p_T\|\widetilde{p}_{T}^{\, \bblambda^\star})
    \le
    O\!\left(\frac{1}{\sqrt{T}}\right)
    +
    O\!\left(
    \frac{1}{T}\sum_{t=0}^{T-1}\big(\chi_{t}^2+\delta_t^2\big)
    + \varepsilon^2
    \right),
\end{align}
which completes the proof.

\textbf{Remarks.} The comparison with the oracle avoids two issues that a comparison with $\mu^\star$ would face. A discrete Gaussian reverse step does not map $q_{\tau}(\cdot|\bblambda^\star)$ exactly onto $q_{\tau-1}(\cdot|\bblambda^\star)$, and $\KL(p_T\|\mu^\star)$ is infinite because $p_T$ has full support while $\mu^\star$ is supported on $\ccalX$. Both effects are part of the oracle's own error below. Under Assumption~\ref{as:VP}, $\sum_\tau b_\tau\leq c_b$, so $\alpha_T^2\geq e^{-2c_b}$ for $T\geq2c_b$. The initialization part of this error therefore does not vanish as $T$ grows with $c_b$ fixed, and it is controlled by the noising horizon $c_b$.

\subsection{Feasibility and Optimality Guarantees}
Let $\varepsilon_\star\coloneqq\mathrm{TV}(\Pi_{\ccalX\#}\,\widetilde{p}_{T}^{\, \bblambda^\star},\mu^*)$ be the total-variation (TV) distance error of the oracle after projecting its outputs onto $\ccalX$. It stems only from discretization and initialization, is common to every diffusion sampler for $\mu^*$ with exact scores, and is not bounded here~\citep{chen2023sampling, benton2024nearly}.
\begin{corollary}[Expected cost and approximate feasibility]\label{cor:primal-feasibility}
Let $\widehat p_T\coloneqq\Pi_{\ccalX\#}\,p_T$ be the law of the PDI outputs after projection onto $\ccalX$, and let $\delta_T\coloneqq\varepsilon_\star+\sqrt{\KL(p_T\|p_T^\star)/2}$. Then $\mathrm{TV}(\widehat p_T,\mu^*)\leq\delta_T$, $\mbE_{\widehat p_T}[f_j(\bbx)]\leq 2R\,\delta_T$ for every constraint $j$, and $|\mbE_{\widehat p_T}[f_0(\bbx)]-\mbE_{\mu^*}[f_0(\bbx)]|\leq 2\|f_0\|_\infty\,\delta_T$.
\end{corollary}
\begin{proof}[Proof of Corollary~\ref{cor:primal-feasibility}]
Since $\mu^*$ is supported on $\ccalX$ and $\Pi_{\ccalX}$ is the identity on $\ccalX$, $\Pi_{\ccalX\#}\,\mu^*=\mu^*$. The triangle inequality, the data-processing inequality for the total variation distance, and Pinsker's inequality give
\begin{align}
    \mathrm{TV}(\widehat p_T,\mu^*)
    ~\leq~ \mathrm{TV}\big(\Pi_{\ccalX\#}\,p_T,\Pi_{\ccalX\#}\,\widetilde{p}_{T}^{\, \bblambda^\star}\big)+\mathrm{TV}\big(\Pi_{\ccalX\#}\,\widetilde{p}_{T}^{\, \bblambda^\star},\mu^*\big)
    ~\leq~ \sqrt{\KL(p_T\|\widetilde{p}_{T}^{\, \bblambda^\star})/2}+\varepsilon_\star ~=~ \delta_T .
\end{align}
For any bounded measurable $h$ on $\ccalX$, $|\mbE_{\widehat p_T}[h]-\mbE_{\mu^*}[h]|\leq2\|h\|_\infty\,\mathrm{TV}(\widehat p_T,\mu^*)$. Taking $h=f_j$, with $\|f_j\|_\infty\leq R$ by Assumption~\ref{as:f_lipschitz} and $\mbE_{\mu^*}[f_j]\leq0$ by feasibility of $\mu^*$, gives the constraint bound. Taking $h=f_0$ gives the cost bound.
\end{proof}
Note that the corollary controls the expected cost and the expected constraint values, but not the entropy term of the regularized objective, which total variation does not control.

%% file: sections/appendix_dual_training.tex
\section{Dual Training}\label{app:dt}
PDI updates the dual variables during inference, allowing the sampler to react instantaneously to constraint violations. Another alternative is to estimate the optimal dual variables during training while training a score model to sample directly from the optimal distribution. We refer to this approach as dual training (DT). The idea behind DT is related to prior work on constrained diffusion models \citep{khalafi2024constrained,khalafi2025composition}. However, this prior work typically considers a single optimization problem, such as fine-tuning a pre-trained model to satisfy a fixed set of constraints, where there is only one optimal dual variable to estimate. In contrast, our formulation considers a family of problem instances, each with its own optimal distribution and corresponding optimal dual variable. In this section, we extend DT to this setting by estimating the dual variables separately for each problem instance during training. The resulting method then serves as an ablation baseline to test the effect of inference-time dual updates.

The dual problem in \eqref{eq:dual_problem}, for a given problem instance $\ccalG$, can be re-written as a bi-level problem,
\begin{align}
    D^*(\ccalG) ~=~ \max_{\bblambda \succeq \mathbf{0}}  & \ \
                        \ccalL(\muDagger, \bblambda; \ccalG)
    \nonumber\\
    \text{with } & \, \, \muDagger = \argmin_{\mu \in \ccalP_2} \
                                    \ccalL(\mu, \bblambda; \ccalG).
\end{align}
The inner problem is equivalent to an unconstrained sampling problem and the diffusion process defined in \eqref{eq:forward_process_family}--\eqref{eq:backward_process_family} provides a mechanism to sample from the target distribution $\muDagger$. Therefore, we replace the inner problem with the problem of training the score model of the reverse process \eqref{eq:backward_process_family}. Let $\bbs_{\bbphi}(\bbx, t, \ccalG)$ be the score model, parameterized by $\bbphi$, and $p_T^{\bbphi}$ be the terminal distribution of the reverse process under a given dual variable $\bblambda$. The parameterized bi-level problem can then be cast as
\begin{align}
    \quad D_{\bbphi}^*(\ccalG) = \max_{\bblambda \succeq \mathbf{0}}  & \ \
                        \ccalL(p_T^{\bbphi_{\bblambda}}, \bblambda; \ccalG)
    \nonumber\\
    \text{ with} & \ \ \bbphi_{\bblambda}(\ccalG) = \argmin_{\bbphi} \
                                    \mbE_{t,\bbx_t} \Big[ \omega(t)\big\|
                                    \bbs_{\bbphi}(\bbx_t, t, \ccalG) - \nabla \log q_{T-t}(\bbx_t|\bblambda; \ccalG)
                                    \big\|_2^2\Big], \ \forall \bblambda.
\end{align}
The expectation is with respect to the noisy samples and the diffusion time. The inner problem in its current form assigns a different parametrization $\bbphi$ for each dual variable $\bblambda$ and problem instance $\ccalG$. This is not a training objective and is generally infeasible. To overcome this challenge, we alternate between the two problems so that, for an outer iteration $k$, the score model is trained only against the most recent dual variable, which is then updated based on the recent constraint violations. More concretely, the training algorithm follows the following dual-ascent iterations:
\begin{align}
    \bbphi_{k} & ~=~  \argmin_{\bbphi} \
                                    \mbE_{\ccalG, t,\bbx_t} \Big[ \omega(t) \big\|
                                    \bbs_{\bbphi}(\bbx_t, t, \ccalG) - \nabla \log q_{T-t}\big(\bbx_t|\bblambda_k(\ccalG); \ccalG\big)
                                    \big\|_2^2 \Big], \label{eq:DT_primal} \\
    \bblambda_{k+1}(\ccalG) & ~=~
                \Bigl[\bblambda_k(\ccalG)
                  + \eta_{\text{{DT}}}\,
                  \mbE_{p_T^{\bbphi_k}} \big[\bbf(\bbx_T; \ccalG)\big]
                \Bigr]_+,
                \ \forall\, \ccalG, \label{eq:DT_dual}
\end{align}
where $\eta_{\text{{DT}}}$ is the dual step size. The expectation in the training objective becomes jointly over diffusion time, noisy samples, and the problem distribution so that a single $\ccalG$-conditional parameterization learns the score field for a family of problems. The dual variables, however, remain instance-specific. Each problem instance $\ccalG$ maintains its own multiplier $\bblambda_k(\ccalG)$, updated only from the constraint violations observed for that instance. 

Updating the dual variables requires estimating the constraint violations under the current sampler, which in turn requires running the full reverse process to obtain clean samples. Implementing \eqref{eq:DT_primal}--\eqref{eq:DT_dual} to completion would therefore be computationally prohibitive. In practice, we use an alternating implementation. We perform a few gradient steps on the score-matching objective in \eqref{eq:DT_primal}, then update the dual variables for a subset of problem instances. Specifically, at each outer iteration, Algorithm~\ref{alg:dt} samples a batch of instances $\ccalB_{\bblambda}$ and updates only their corresponding multipliers, while the remaining dual variables are kept unchanged. As in Algorithm~\ref{alg:training}, the score target is computed using the MC estimator in \eqref{eq:MC_approximation} and the noisy samples $\bbx_t$ are obtained from rollouts of the inference dynamics at the beginning of each outer iteration.

At inference time, DT starts from Gaussian noise and runs the reverse diffusion process using the trained score model $\bbs_{\bbphi^*}(\bbx_t,t,\ccalG)$. For a test instance $\ccalG$, the reverse dynamics are
\begin{align}
\bbx_{t+1} & ~=~
\frac{1}{\sqrt{a_{T-t}}} \Big(
 \bbx_{t} + b_{T-t} \bbs_{\bbphi^*}\big(\bbx_t, t, \ccalG\big)
\Big)
+ \sqrt{b_{T-t}} \bbepsilon_t .
\end{align}
No dual variable is initialized, provided to the score model, or updated during this process. The constraint information enters only through the learned parameters $\bbphi^*$, which were trained using the instance-dependent dual estimates in \eqref{eq:DT_primal}--\eqref{eq:DT_dual}. Thus, DT produces samples in a single reverse pass and does not adapt the trajectory to constraint violations observed during inference.

\begin{algorithm}[t]
\caption{Dual Training}
\label{alg:dt}
\begin{algorithmic}[1]
\State Initialize $\bblambda(\ccalG) \leftarrow \bblambda_0$
  for all $\ccalG$
\State Initialize score network $\bbs_{\bbphi}$
\For{$d = 1, \dots, D$}
    \Comment{Outer: dual iterations}
    \State Initialize replay buffer $\mathcal{B} \leftarrow \varnothing$
    \For{each problem $\ccalG$ in a batch $\ccalB_{\text{train}}$}
            \State \textbf{Rollout}: sample $\mathbf{x}_0^{(i)} \sim \mathcal{N}(\mathbf{0}, \mathbf{I})$;
              run inference with $\bbs_{\bbphi}$
              to obtain
              $\{(\mathbf{x}_t^{(i)}, t)\}_{t=0}^{T-1}, \ \forall i \in [I]$
            \State Push $\{(\mathbf{x}_t^{(i)}, t, \ccalG)\}_{i,t}$ to $\mathcal{B}$
    \EndFor
    \For{$e = 1, \dots, E$}
        \Comment{Inner: train at fixed $\bblambda$}
        \State Sample a minibatch
          $\{(\mathbf{x}_t^{(j)}, t^{(j)}, \ccalG^{(j)})\}_{j=1}^B$
          from $\mathcal{B}$
          \For{$j = 1, \ldots, B$}
        \If{$\mathrm{Bernoulli}(\rho_{\text{pert}}) = 1$} \Comment{Perturb the sample}
                \State $\bbx_t^{(j)} \leftarrow \bbx_t^{(j)} + \epsilon_{\bbx} \cdot \boldsymbol{z}_x$, \quad $\boldsymbol{z}_x \sim \mathcal{N}(\mathbf{0}, \mathbf{I})$
            \EndIf
        \EndFor
        \State $\bbphi \leftarrow \bbphi
          - \alpha\,\nabla
          \frac{1}{B}\sum_{j}
          w(t^{(j)})\,
          \bigl\lVert
            \bbs_{\bbphi}(\mathbf{x}_t^{(j)}, t^{(j)}, \ccalG^{(j)})
            - \nabla \log q_{T-t^{(j)}} (\bbx_t^{(j)}\mid \bblambda(\ccalG^{(j)}); \ccalG^{(j)})
            \bigr\rVert_2^2$
    \EndFor
    \For{each problem $\ccalG$ in a batch $\ccalB_{\bblambda}$}
        \Comment{Dual ascent}
        \State Generate samples
          $\{\bbx_T^{(i)}\sim p_T^{\bbphi}\}_{i=1}^I$
          via full reverse diffusion using the most recent $\bbphi$
        \State $\bblambda(\ccalG)
                        \leftarrow
                        \Bigl[\bblambda(\ccalG)
                          + \eta_{\text{{DT}}}\,
                          \sum_i \bbf(\bbx_T^{(i)}; \ccalG) / I
                        \Bigr]_+$
    \EndFor
\EndFor
\State \Return $\bbphi$
\end{algorithmic}
\end{algorithm}

%% file: sections/appendix_synthetic_data.tex
\section{Extended Numerical Results: Mixture of Gaussians} \label{app:MoG}

We first validate PDI on the problem of sampling from a weighted mixture of Gaussians in $\reals^{d}$,
\begin{align}
    f_0(\bbx) ~=~ - \log \sum_{k=1}^K w_k \, \ccalN\big( \bbx; \bbmu_k, \bbSigma_k \big),
\end{align}
truncated to a polytope $\{\bbx: \bbA\bbx\le\bbb\}$.  The optimization problem is
\begin{align}
  \min_{\mu(\bbx)}\;
    \mathbb{E}_{\bbx\sim \mu} \bigl[f_0(\bbx)\bigr]
  \qquad\text{s.t.}\qquad
    \mathbb{E}_{\bbx\sim \mu} \bigl[\bbA^\top\bbx - \bbb\bigr]
      \;\preceq\; \mathbf{0},
\end{align}
i.e., the sampler must produce a distribution that concentrates on
high-density regions of the mixture while satisfying $M$ linear
inequality constraints in expectation. This is not an everywhere constrained sampling problem, where each sample $\bbx \in \reals^{d}$ is required to satisfy the constraints. In \eqref{eq:synthetic}, the constraints are imposed on average over the target distribution. This allows some samples to violate the constraints while others should be strictly feasible to compensate.

\subsection{Baseline configuration}
Each baseline shares the evaluation seed, batch size, and diffusion discretisation $T=500$ where applicable.

\paragraph{Unconstrained DM.} We run our trained model with $\bblambda_t = \mathbf{0}$ for all time steps. This eliminates the constraint term from the energy function, and lets the model sample from the unconstrained Boltzmann density $\exp(-f_0(\bbx)/\beta)$.

%% ====================================================================
\paragraph{PDM.}
Similarly, we run a reverse process with $\bblambda_t=\mathbf{0},\forall t$.  At each step, the Tweedie prediction~$\hat{\bbx}_0$ is projected onto the feasible set before being used in the DDPM posterior,
\begin{align}\label{eq:pdm}
  \tilde{\bbx}_0
  \;=\;
  \Pi_{\mathcal{F}}\!\bigl(\hat{\bbx}_0\bigr),
  \qquad
  \mathcal{F} = \bigl\{\bbx: \bbf(\bbx) \preceq \mathbf{0},\;
    j=1,\dots,M\bigr\}.
\end{align}
PDM then forces feasibility by construction and is therefore a hard-constraint method, unlike PDI.

%% ====================================================================
\paragraph{PDL.} Direct Langevin dynamics in the $\bbx$-space with
per-sample dual variables:
\begin{align}\label{eq:pdl}
  \bbx_{t+1} &= \bbx_{t}
    - \eta_\rmp\;\nabla_{\bbx_t}\Big(f_0(\bbx_t) + \bblambda_t^\top \bbf(\bbx_t)\Big)
    + \sqrt{2\eta_\rmP}\,\bbepsilon_t, \\
      \bblambda_{t+1} &= \bigl[\bblambda_{t}
    + \eta_\rmd\,\bbf(\bbx_{t+1})\bigr]_+,
\end{align}
where $\bbepsilon_t \sim \ccalN(0, \bbI)$.
The main difference between PDL and PDI is that the former operates in the clean sample space. It therefore uses the gradient of the energy function to update $\bbx$. In PDI, the dynamics evolve under a decreasing noise level, and at each time step, we follow the score of the corresponding denoised distribution.

\begin{figure}
    \centering
    \includegraphics[width=0.9\linewidth]{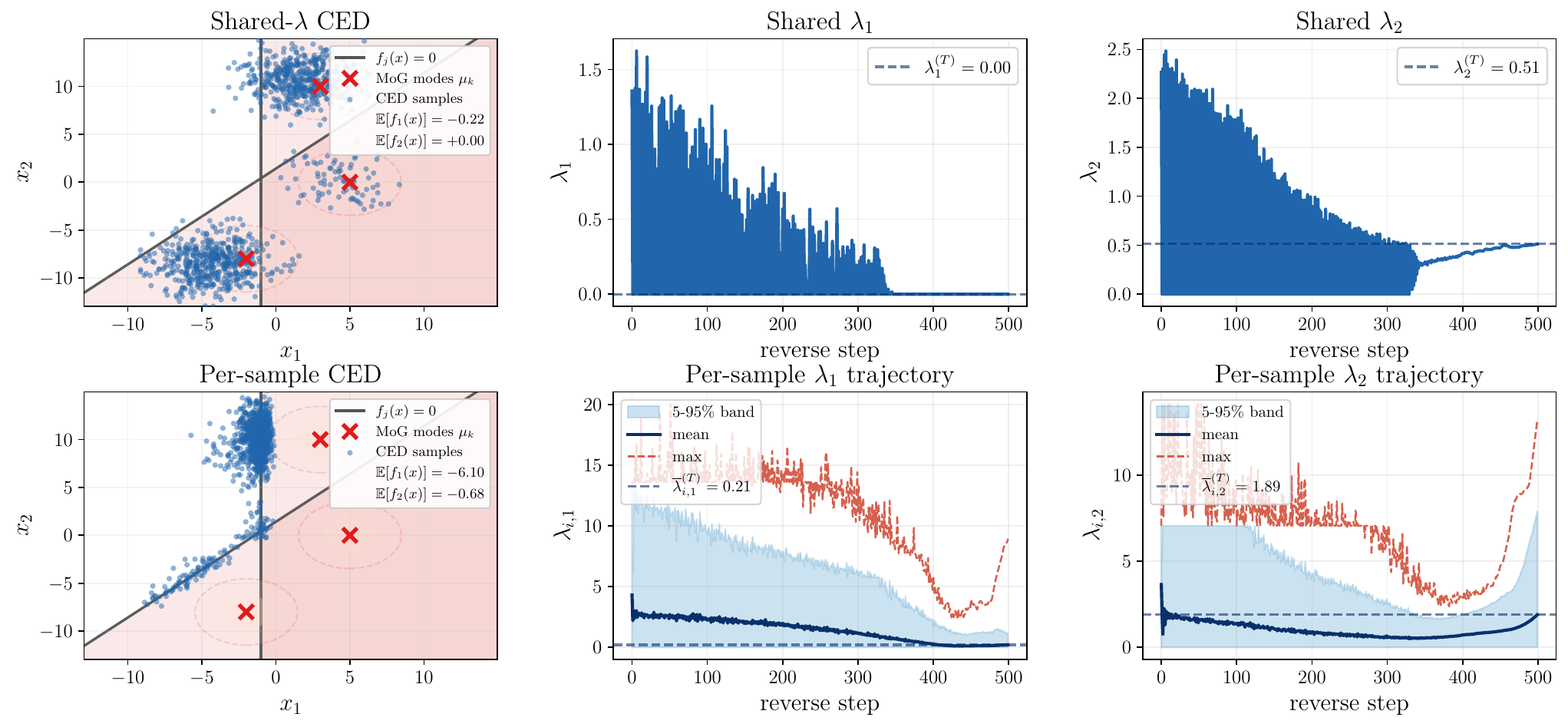}
    \vspace{-10pt}
    \caption{Comparison of \textbf{C}onstrained \textbf{E}nergy-based \textbf{D}iffusion-sampler (CED) in enforcing the constraints in expectation (top) vs per sample (bottom). Per-sample enforcement produces conservative samples that satisfy the constraints individually but fall short in providing a competitive objective.}
    \label{fig:per-sample-lambda}
\end{figure}

A distinction arises between these baselines and our method PDI. In all the aforementioned updates, the correction is made to each sample individually, forcing all samples to abide by the constraints. Their solutions are therefore feasible in our problem formulation, but suboptimal because they are more conservative. Figure \ref{fig:per-sample-lambda} shows a toy example of the problem with $d=2, K= 3$, and $M=2$. Per-sample enforcement sacrifices the opportunity to sample from the high-density regions of the MoG while still satisfying the constraints on average. Instead, it concentrates samples near the boundary of the feasible set, shown as the white region.

In the following table, we provide all the hyperparameters of the baselines:
\begin{center}
\begin{tabular}{l c c c c c c}
\toprule
Method & $\eta_\rmd$ & $\eta_\rmp$  & $T$ & Diffusion Noise Schedule \\
\midrule
PDL         & 10   & $10^{-3}$  & 500 & -- \\
PDM          & --   & --  & 500 & cosine \\
% DPS         & --   & -- & 500 & cosine \\
Unconstrained  & -- & -- & 500 & cosine \\
\bottomrule
\end{tabular}
\end{center}

\subsection{Data and Architecture}
The problem consists of a $K$-component Gaussian mixture, truncated by $M$ halfplane constraints. Mode centers are placed uniformly at random in a ball of radius proportional to a spread parameter, with isotropic covariances drawn from a specified range. The constraint normals are chosen so that some halfplanes separate pairs of modes while others cut through mode basins, ensuring that constraints are active near high-density regions. Constraint levels $\bbb$ are calibrated so that a target fraction of modes have infeasible centers and the remaining feasible modes sit close to constraint boundaries.
Since this example is used as a proof of concept, the dataset has only one problem instance.
% The Constraint levels $b_j$ are calibrated so that a target fraction of modes have infeasible centres, and feasible centres sit within a controlled margin of the nearest active constraint.

The score network is an MLP with FiLM conditioning.  The input $(\bbx_t, \bblambda)$ is projected into a hidden dimension, added to the sinusoidal timestep embedding, and passed through a stack of residual blocks.  Each block applies a linear layer, layer normalization with FiLM modulation from the timestep embedding (scale and shift), and a SiLU activation. The output is projected back to $\mathbb{R}^d$ to predict the noise $\bbepsilon_t$.

\subsection{Hyperparameter Choices}
We ran a sweep over the hyperparameters: $\beta, T, \eta, \rho, K_{\text{MC}}, \nu_{\text{max}}, \epsilon_{\bbx}$, and $\epsilon_{\bblambda}$, and Table \ref{tab:hyperparams} shows the values we use in our experiments.
All experiments were run on a single NVIDIA GeForce RTX 3090 card.

\begin{table}[!t]
\centering
\caption{PDI training hyperparameters for the MoG experiment.}
\label{tab:hyperparams}
\small
\setlength{\tabcolsep}{4pt}
\begin{tabular}{@{}lc@{\hspace{1.5em}}lc@{}}
\toprule
\textbf{Parameter} & \textbf{Value}
&
\textbf{Parameter} & \textbf{Value} \\
\midrule

\multicolumn{2}{@{}l}{\textit{Problem}}
&
\multicolumn{2}{l@{}}{\textit{Training}} \\
Dimension $d$ & 30
&
Optimizer & AdamW \\
Mixture components $K$ & 12
&
Learning rate & $10^{-3}$ \\
Constraints $M$ & 10
&
SGD steps & 8000 $(800 \times 10)$ \\
&
&
Batch size $B$ & 256 \\
&
&
MC samples $K_{\mathrm{MC}}$ & 256 \\
&
&
Rollout batch $|\ccalB_{\text{train}}|$ & 4 graphs \\
&
&
Exploitation fraction $\rho_{\text{exp}}$ & 0 $\to$ 0.7 \\

\midrule
\multicolumn{2}{@{}l}{\textit{Diffusion}}
&
\multicolumn{2}{l@{}}{\textit{Training -- dual ascent rollouts}} \\
Noise schedule & cosine
&
Training dual step size $\eta_{\mathrm{train}}$ & 0.01 \\
Timesteps $T$ & 500
&
Dual variable cap $\bblambda_{\max}$ & 50 \\
Inverse temperature $\beta^{-1}$ & 50
&
& \\

\midrule
\multicolumn{2}{@{}l}{\textit{Score network}}
&
\multicolumn{2}{l@{}}{\textit{Training -- $\bblambda$ prior}} \\
Architecture &  MLP
&
$\bblambda$-prior family &  $\exp(\nu)$ \\
Hidden dimension & 512
&
$\nu_{\min}$ & 0.1 \\
Number of layers & 8
&
$\nu_{\max}$ & 20.0 \\
Parameters & 8.73M
&
& \\
Conditioning inputs & $(\bbx_t,t,\bblambda)$
&
& \\

\midrule
\multicolumn{2}{@{}l}{\textit{Training -- perturbation}}
&
\multicolumn{2}{l@{}}{\textit{Inference}} \\
Perturbation fraction $\rho_{\text{pert}}$ & 1.0
&
Inference dual step size $\eta$ & 1.0 \\
$\bbx_t$ perturbation std $\epsilon_{\bbx}$ & 2.0
&
Initial $\bblambda_0$ & $\mathbf{0}$ \\
$\bblambda$ perturbation std $\epsilon_{\bblambda}$ & 5.0
&
Feasibility tolerance & 0.02 \\
&
&
Inverse temperature $\beta^{-1}$ & 50 \\

\bottomrule
\end{tabular}
\end{table}

\subsection{Additional Results}
\begin{figure}[t]
    \centering
    \includegraphics[width=0.8\linewidth]{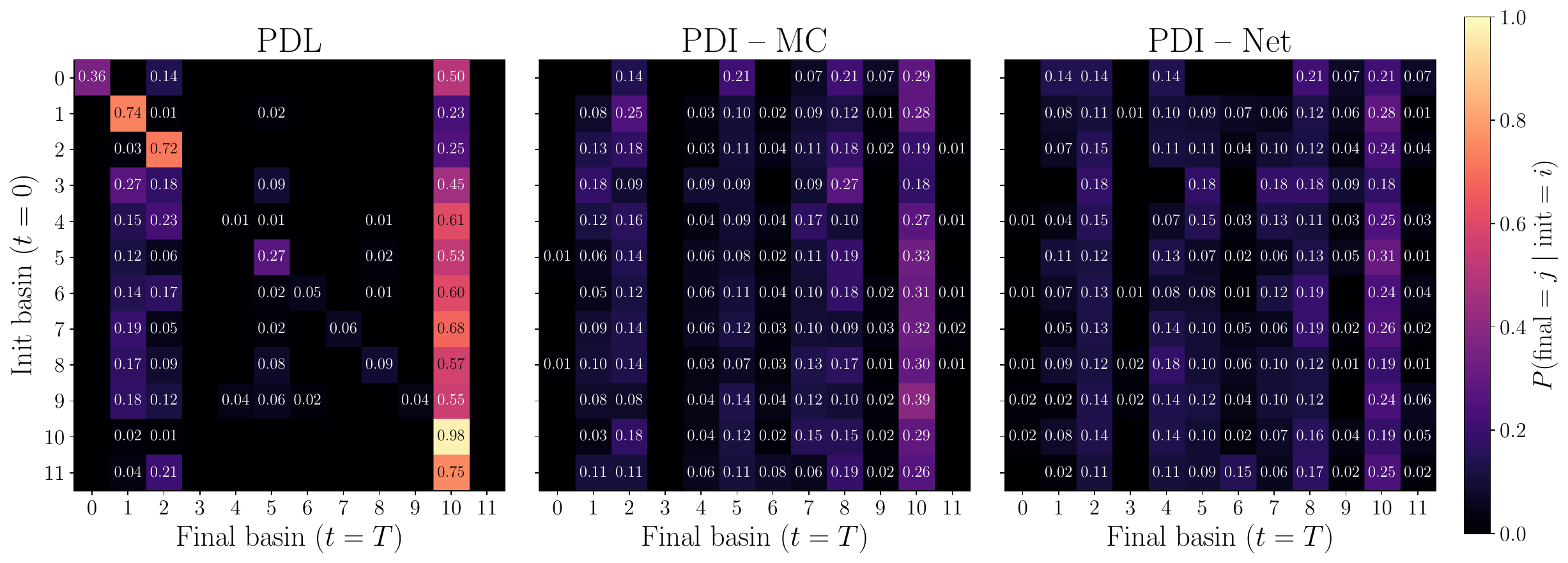}
    \caption{\textbf{Initial to final transition matrix.} The matrix is almost diagonal in the PD Langevin case, reflecting the fact that samples converge to their nearest modes.} 
    % $(\beta^{-1}=10, \eta_t=0.1)$.}
    \label{fig:syn_transition_matrix}
\end{figure}

We also record the input-to-output transition matrix and report it in Figure~\ref{fig:syn_transition_matrix}. The figure shows that PDL divides the mass between one mode ($\# 10$) and the diagonal of the transition matrix. The mass on the diagonal indicates that each input sample typically converges to its nearest mode. In contrast, PDI does not exhibit this behavior. This is partly due to the noise injected along the reverse trajectory, which acts as an annealing mechanism and allows samples to move across modes. It is also due to the entropy regularization in the PDI formulation, controlled by the temperature parameter, which encourages a distributional solution rather than deterministic convergence to the nearest mode. This temperature-driven randomization is absent in PDL.

%% file: sections/appendix_wra.tex
\section{Extended Numerical Results: Wireless Resource Allocation} \label{app:wra}

\def \numNodes {200}
\def \bandwidthInMHz {40}
\def \noisePSDIndBmPerHz {-174}
\def \PmaxInMW {10}

\def \deploymentRhigh {5800}

\pgfmathsetmacro{\deploymentAreahigh} {(\deploymentRhigh / 1000) * (\deploymentRhigh / 1000)}

\pgfmathsetmacro{\densityhigh} { \numNodes * (1 / \deploymentAreahigh)}

We study stochastic constrained optimization for optimal wireless resource allocation
% We investigated this setup with slight variations in 
\citep{uslu2025generative, uslu2026graphsignaldiffusionmodels, uslu2025faststateaugmentedlearningwireless}.  

Consider an ad-hoc wireless network comprised of $N = \numNodes$ transmitter-receiver (tx-rx) pairs deployed uniformly over a square area of side length $R = \deploymentRhigh$ meters, yielding an average density of $\nu = \pgfmathprintnumber[fixed, precision=1]{\densityhigh}$ tx-rx pairs/\SI{}{\kilo\meter\squared}. Each transmitter communicates with a single designated receiver, and each receiver treats incoming signals from all other transmitters as interference. The resource allocation (optimization) variable is the transmit power vector $\mathbf{x} \in [0, P_{\max}]^N$, where $P_{\max} = \PmaxInMW$~mW.

\paragraph{Channel model.}
The channel model includes both large-scale and small-scale fading effects. We denote by $h_{ij}$ the large-scale channel gain from transmitter~$i$ to receiver~$j$, governed by a dual-slope path-loss model with log-normal shadowing of standard deviation $\sigma = 7$~dB. A \emph{network configuration} $\mathbf{H} \in \mathbb{R}^{N \times N}_+$ is the matrix of all large-scale gains $[\mathbf{H}]_{ij} = h_{ij}$, and it is jointly determined by the random geometric deployment of the tx-rx pairs, which fixes the pairwise distances and hence the path-loss components, and the realization of the log-normal shadowing. Each network configuration~$\mathbf{H}$ remains fixed over the operating horizon, while the small-scale fading, which follows a Rayleigh distribution with a pedestrian velocity of $1$~m/s, produces time-varying instantaneous channel gains $h_{ij,t}$ at each time slot~$t$ of duration $50$~milliseconds. We set the channel bandwidth to $W = \bandwidthInMHz$~MHz and the noise power spectral density to $N_0 = \noisePSDIndBmPerHz$~dBm/Hz.

Given an allocation $\bbx_t$ and fading realization $\mathbf{H}_t$ at time slot~$t$, the instantaneous rate of receiver~$j$ is
\begin{equation}
\label{eq:sinr_rate}
\tilde{r}_j(\bbx_t, \mathbf{H}_t) = \log_2\!\left(1 + \frac{[\bbx_t]_j \cdot |h_{jj,t}|^2}{W N_0 + \sum_{i \neq j} [\bbx_t]_i \cdot |h_{ij,t}|^2}\right).
\end{equation}
The ergodic rate of receiver~$j$ is obtained by averaging the instantaneous rates over $T$ time slots, i.e., $r_j \approx (1/T) \sum_{t=1}^{T} \tilde{r}_j(\bbx_t, \mathbf{H}_t)$.

\paragraph{Optimization objective.}
For a given network state~$\mathbf{H}$, the goal is to maximize the ergodic sum-rate subject to per-user minimum ergodic rate constraints. That is,
\begin{equation}
\label{eq:power_control_app}
P^*_{\text{pc}}~=~\max_{\mu(\bbx)} \; \mathbf{1}^\top \mathbf{r}\!\left(\mu, \mathbf{H}\right) + \beta \ccalH(\mu), \quad \text{s.t.} \quad \mathbf{r}\!\left(\mu, \mathbf{H}\right) \geq \mathbf{1} \cdot r_{\min},
\end{equation}
where $\mu$ denotes a power allocation policy, and $r_{\min}$ is the minimum ergodic rate requirement. The solution to this problem is rarely a degenerate policy. Because users compete for the same communication resources and interfere with one another, it is generally impossible for all users to transmit at high power simultaneously while satisfying their rate requirements. The optimal policy is therefore often a switching policy, where different users are activated at different times so that each receives enough transmission opportunities to meet its rate constraint on average. This naturally leads to a multimodal optimal distribution, with different modes corresponding to different active-user patterns.

\subsection{Data and Architecture}
\paragraph{Dataset Generation.}
We generate a dataset of $256$ network configurations in total, viewed as independent samples from the underlying distribution of network configurations induced by the random geometry of tx-rx deployments and random shadowing for the given value of $R$. A 5:1:2 train-validation-test split results in a training dataset of $|\ccalD| = 160$ networks, $|\ccalV_{\text{al}}| = 32$ validation networks, and $|\ccalT| = 64$ test networks.

\paragraph{Graph representation.}
We represent each network configuration $\mathbf{H}$ as a directed graph $\mathcal{G}_{\mathbf{H}} = (\mathcal{V}, \mathcal{E}, \bbw)$, where the node set $\mathcal{V} = \{1, \dots, N\}$ corresponds to the tx-rx pairs. We assign each directed edge $(i, j) \in \mathcal{E}$ a weight proportional to the log-normalized channel gain given by
\begin{equation}
\label{eq:edge_weight}
e_{ij} \propto \log_2\!\left(1 + \frac{P_{\max} |h_{ij}|^2}{W N_0}\right),
\end{equation}
and sparsify the graph by thresholding the weak edges so that we retain only the top-$10$ strongest incoming edges per receiver node. 

\paragraph{Graph Neural Networks (GNNs).} The power allocations~$\mathbf{x}$ and ergodic rate vectors~$\bbr$ are naturally interpreted as node-level signals supported on~$\mathcal{G}_{\mathbf{H}}$. We therefore use GNNs as the base parameterization for the score model, which takes the problem representation $\ccalG_{\bbH}$ as input. We implement graph convolutions using Torch Geometric's TAGConv (with $K$=2). The model consists of $L{=}8$ TAGConv  residual blocks with hidden dimension $d_h{=}256$.

The timestep  information is injected through  Feature-wise Linear Modulation (FiLM).
% a sinusoidal embedding produces per-layer scale and shift parameters applied after LayerNorm.  
The dual variable
$\bblambda\in\reals^N$ is concatenated to the input features $\bbx$.

\subsection{Baselines}
\paragraph{PDM.} 
We run our trained score model with $\bblambda_t=0$ for all the denoising steps to eliminate the dual ascent.
After each step, we perform a projection into the feasible set by solving the augmented-Lagrangian problem:
\begin{equation}\label{eq:pdm_project}
  \bbx_t \leftarrow \arg\min_{\bbw} \;\|\bbw - \bbx_t\|^2
  + \tfrac{\zeta}{2}\|[\bbf(\bbw)]_+\|^2,
\end{equation}
where $[\bbx]_+ = \max\{\mathbf{0}, \bbx \}$. We solve the problem via a few steps of gradient descent.

\paragraph{DPS.}
We run a reverse process with our trained score model while forcing $\bblambda_t=\mathbf{0}, \forall t,$ to eliminate the dual ascent.
At each step in the inference process, the prediction $\bbx_t$ is corrected by a gradient step that penalizes constraint violation before it is fed to the next iteration,
\begin{align}\label{eq:dps}
  {\bbx}_t
  \leftarrow
  {\bbx}_t
  \;-\;
  s_t\;\nabla_{{\bbx}_t}
      \bigl\| \bbf(\bbx_t) \bigr\|^2.
\end{align}
The gradient is applied directly to the denoised signal at the end of each step, and weighted by a scalar step size $s_t$. The step size is set to the inverse of the noise variance at each denoising step. 

\paragraph{ST \& PD Expert Policy.}
Following the methodology in \citep{uslu2026graphsignaldiffusionmodels}, for each training network $\{ \bbH^{(b)} \}_{b \in \ccalD}$, we solve the power control problem using an expert algorithm, specifically a dual gradient descent algorithm. The expert algorithm generates trajectories of primal iterates that are near-optimal and feasible in the ergodic sense. We view these iterates as draws from a stochastic, optimal policy, collecting $M = 200$ samples from the convergence regime of the algorithm to obtain an expert dataset of resource allocation vectors $\{ \bbx^{(m)} \left( \bbH^{(b)} \right) \}_{m = 1}^{M}$.

We train a diffusion model to imitate the expert data generated by the PD Expert. 
% A conventional denoising diffusion model (DDIM~\citep{song2020ddim})
% trained on the data distribution of feasible allocations.
The neural network $\boldsymbol{\epsilon}_\theta(\bbx_t, t)$
is trained via the standard denoising score matching objective
and sampled using the deterministic DDIM update:
\begin{equation}\label{eq:ddim}
  \mathbf{x}_{t-1} = \sqrt{\bar{\alpha}_{t-1}}\,\hat{\mathbf{x}}_0
  + \sqrt{1 - \bar{\alpha}_{t-1}}\,\boldsymbol{\epsilon}_\theta(\bbx_t, t).
\end{equation}
This method has no explicit constraint enforcement mechanism;
feasibility depends entirely on the training data distribution and the expressivity of the parameterization.

In order to replicate the performance of~\citep{uslu2026graphsignaldiffusionmodels}, we use the U-Graph Neural Network (U-GNN) parametrization proposed therein, which was designed for graph signal diffusion. This ensures that the DDIM imitation baseline follows the original architectural setup. We do not use U-GNN as the base parametrization for our method because it is computationally heavy. Instead, we use a vanilla GNN, since PDI enforces the constraints through dual ascent during sampling and therefore does not need to rely solely on a highly expressive score parametrization for feasibility.

\paragraph{DT.} The training procedure is described in Appendix~\ref{app:dt}. The score model shares the same architecture as PDI-Net, except that it uses three conditioning channels instead of four.

\begin{wrapfigure}{r}{0.44\textwidth}
    \centering
    \vspace{-10pt}
    \includegraphics[width=\linewidth]{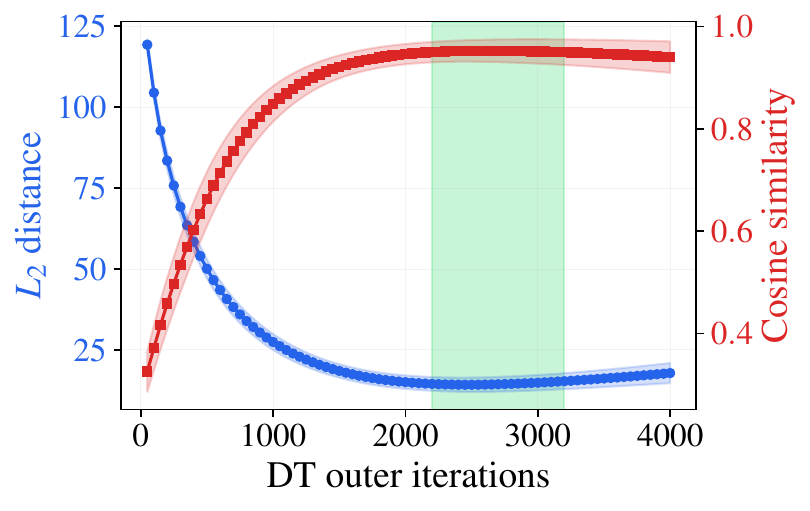}
    \vspace{-10pt}
    \caption{\small \textbf{Dual Training.} Cosine similarity and $L_2$ distance between the dual variables across the DT training iterations and the last-iterate PDI dual variables. The shaded region indicates the training iterations where the cosine similarity peaks above $0.95$. The metrics then drift slightly afterwards. 
    % (Right) Complementary slackness of PDI and DT (at peak cosine similarity) on a training example.
    }
    \vspace{-40pt}
    \label{fig:dt_wra}
\end{wrapfigure}

All hyperparameters of these baselines are given in Table~\ref{tab:wra_baseline_hparams}.

\begin{table}[t]
\centering
\caption{Baseline hyperparameters for the wireless allocation experiment. The DT settings differ from PDI in Table~\ref{tab:wra_ced_hparams}; all other problem, diffusion, architecture, and evaluation settings are identical.}
\label{tab:wra_baseline_hparams}
\small
\setlength{\tabcolsep}{4pt}
\begin{tabular}{@{}lc@{\hspace{1.5em}}lc@{}}
\toprule
\textbf{Parameter} & \textbf{Value}
&
\textbf{Parameter} & \textbf{Value} \\
\midrule

\multicolumn{2}{@{}l}{\emph{ST}}
&
\multicolumn{2}{l@{}}{\emph{PDL}} \\
Backbone & UGNN \citep{uslu2026graphsignaldiffusionmodels}
&
Iterations & 500 \\
Base channels / levels & 64 / 4
&
Primal learning rate $\eta_p$ & $10^{-4}$ \\
Graph filter & $K=2$ hops
&
Dual learning rate $\eta_d$ & 1 \\
Layers per level & 2
&
Initial value $\bblambda_0$ & 0 \\
Timesteps $T$ & 500
&
& \\
Noise schedule & Linear
&
& \\
Optimizer & AdamW ($\beta_1\!=\!0.9$, $\beta_2\!=\!0.98$)
&
& \\
Learning rate / weight decay & $10^{-4}$ / $10^{-4}$
&
& \\
LR schedule & Cosine
&
& \\
Batch size & 960
&
& \\
Conditioning & 3 channels $(\bbx, t, \ccalG)$
&
& \\
Epochs & 5\,000
&
& \\

\midrule
\multicolumn{2}{@{}l}{\emph{PDM}}
&
\multicolumn{2}{l@{}}{\emph{DPS}} \\
Base model & PDI-Net with $\bblambda=0$
&
Base model & PDI-Net with $\bblambda=0$ \\
Projection iterations & 100
&
& \\
Projection step size & 0.05
&
& \\
Regularization $\zeta$ & 30 & & \\

\midrule
\multicolumn{4}{@{}l}{\emph{DT}} \\
Score model & Same PDI-Net architecture
&
Outer iterations $D$ & 4\,000 \\
Conditioning channels & 3 $(\bbx, t, \ccalG)$
&
Inner SGD steps $E$ & 10 \\
Dual input & none
&
Minibatch size $|\ccalB_{\text{train}}|$ & 4 problems \\
Samples per network $I$ & 20
&
% MC score chunk & 16 \\
Initial $\bblambda_0$ & 10 \\
$\bblambda$ update batch size $|\ccalB_{\bblambda}|$ & 16 problems
&
Dual learning rate $\eta_{\text{DT}}$ & 0.1 \\
\bottomrule
\end{tabular}
\end{table}

\subsection{Additional Results} \label{app:wra_results}

\textbf{Dual training (continued).} We investigate the difference between PDI and DT more closely by comparing the dual trajectories. Figure~\ref{fig:dt_wra} plots the cosine similarity and the $L_2$ distance between the dual variables generated along the DT trajectory and the last dual iterate obtained by the PDI dynamics. These quantities are averaged over the training examples. We use the training dataset because DT does not infer dual variables during inference.

\begin{figure}
    \centering
    \includegraphics[width=\linewidth]{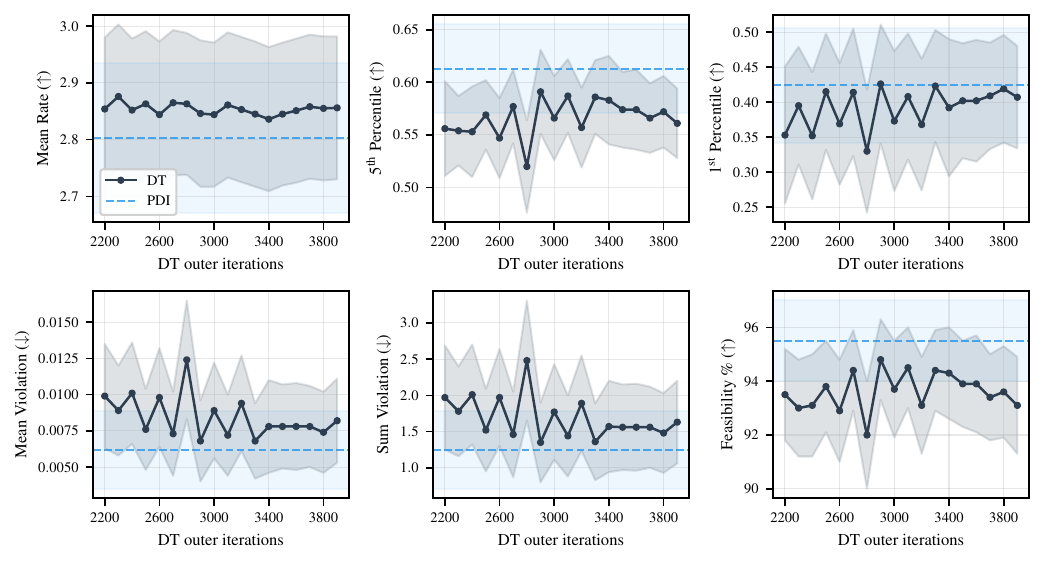}
    \vspace{-10pt}
    \caption{\small \textbf{Performance of DT checkpoints along the training trajectory.} The DT models fail to attain the same $5$th percentile rates, mean violations and feasibility percentage as PDI. All the metrics are evaluated over the test set.}
    \label{fig:dt_metrics}
\end{figure}

The cosine similarity reaches its maximum between the epochs $2200$ and $3200$, after which the DT dual variables begin to drift away from the PDI dual variables.
We therefore save all DT checkpoints after epoch $2200$, evaluate them on the test set and report the resulting metrics in Figure~\ref{fig:dt_metrics}. The figure shows that all DT checkpoints achieve higher mean rates (objective) than PDI, shown by the dashed blue line. This suggests that the models offer higher mean rates by reducing interference through more conservative power allocations. This reduction in interference benefits some users but sacrifices tail performance. This is reflected in the lower fifth and first percentile rates and feasibility percentages across all checkpoints.
% The most recent DT models nevertheless maintain comparable mean and sum violations to those of PDI. This is because the higher $1$st-percentile rates compensate for the lower $5$th-percentile rates when violations are aggregated across users. 

These observations suggest that the dual variables generated by DT do not induce the same primal solutions obtained by PDI dynamics.
This can be attributed to the fact that the PDI score field is conditioned on the current dual variable $\bblambda_t$ while that of DT is an unconditional score field obtained after training over a history of changing dual variables.
 Therefore, even when the DT dual variable is close to the PDI dual variable, the checkpoint at that epoch does not need to approximate the score field associated with that particular dual variable. The reason is that the checkpoint represents the accumulated effect of the preceding training trajectory, not a score model explicitly indexed by the current dual variable.

 To further separate the quality of the learned dual variables from the quality of the corresponding score model, we continue training several DT checkpoints for extra $1000$ outer iterations while freezing their dual variables. We refer to the resulting models as DT+. This experiment asks whether the dual variables produced by DT are useful once the score model is given enough optimization time to adapt to them. Table~\ref{tab:wra_dt} reports the two best DT+ models and shows a substantial improvement over the original DT checkpoints. This indicates that DT can reach informative dual variables, but the score model at the corresponding checkpoint has not necessarily learned the score field induced by that dual value. Rather, because the model is updated only partially while the dual variables evolve, the checkpoint reflects the accumulated effect of the preceding training trajectory. In this sense, practical DT behaves as a score model shaped by a history of dual variables, and not as a fully optimized score model for the current dual variable. The improvement of DT+ also comes with a much larger training cost ($4000$ original outer iterations in addition to extra $1000$ iterations for tuning the model), whereas PDI reaches comparable performance using only $400$ training outer iterations.

\begin{table}[t]
\centering
\small
\caption{\small \textbf{DT after extra training at fixed dual variables (DT+).} We continue training several DT checkpoints with no dual updates and report the best two results (corresponding to iterations 2200 and 2900). PDI achieves comparable performance with much lighter training.}
\label{tab:wra_dt}
\begin{tabular}{lcccccc}
\toprule
  Method  & Mean rates $(\uparrow)$ & 5th\%ile rate $(\uparrow)$ & 1st\%ile rate $(\uparrow)$ & Mean viol. $(\downarrow)$ & Sum viol. $(\downarrow)$   \\
    \midrule
    PDI--Net &
    ${2.803} {\scriptstyle\pm 0.132}$ & ${0.613} {\scriptstyle\pm 0.042}$ & $0.424 {\scriptstyle\pm 0.082}$ & ${0.0062} {\scriptstyle\pm 0.0027}$ & ${1.25} {\scriptstyle\pm 0.54}$    \\
    DT+ (2200) & $2.814 {\scriptstyle\pm 0.128}$ & $0.656 {\scriptstyle\pm 0.051}$ & $0.426 {\scriptstyle\pm 0.112}$ & $0.0060 {\scriptstyle\pm 0.0029}$ & $1.19 {\scriptstyle\pm 0.59}$  \\
    DT+ (2900)  & $2.822 {\scriptstyle\pm 0.128}$ & $0.615 {\scriptstyle\pm 0.045}$ & $0.422 {\scriptstyle\pm 0.090}$ & $0.0063 {\scriptstyle\pm 0.0028}$ & $1.25 {\scriptstyle\pm 0.57}$ \\
\bottomrule
\end{tabular}
\vspace{-5pt}
\end{table}

The improvement of DT+ shows that the dual variables produced by DT are informative. However, it does not remove the limitation of using an unconditional score model. Since DT does not take the dual variable as input, the correction induced by the dual trajectory is absorbed into the score-model parameters rather than represented explicitly. This can work well for problem instances whose required correction is close to the one encoded by the trained model. At inference, however, new problem instances may induce different optimal multipliers or different constraint residuals along the denoising trajectory. The DT score field cannot adapt to these instance-dependent corrections during sampling, because the dual variable is not provided to the model.

PDI reduces this limitation by learning a $\bblambda$-conditioned family of score fields. During inference, the current dual iterate is passed directly to the score model, so each reverse step uses a score field matched to the dual correction used at that step. This does not mean that PDI samples from the Gibbs distribution of one fixed dual variable, since the final sample is produced by a sequence of kernels conditioned on different dual iterates. Rather, the advantage is that the constraint correction and denoising dynamics remain coupled along the trajectory, allowing constraint violations to influence subsequent denoising steps more directly.

\begin{figure}
    \centering
    \includegraphics[width=\linewidth]{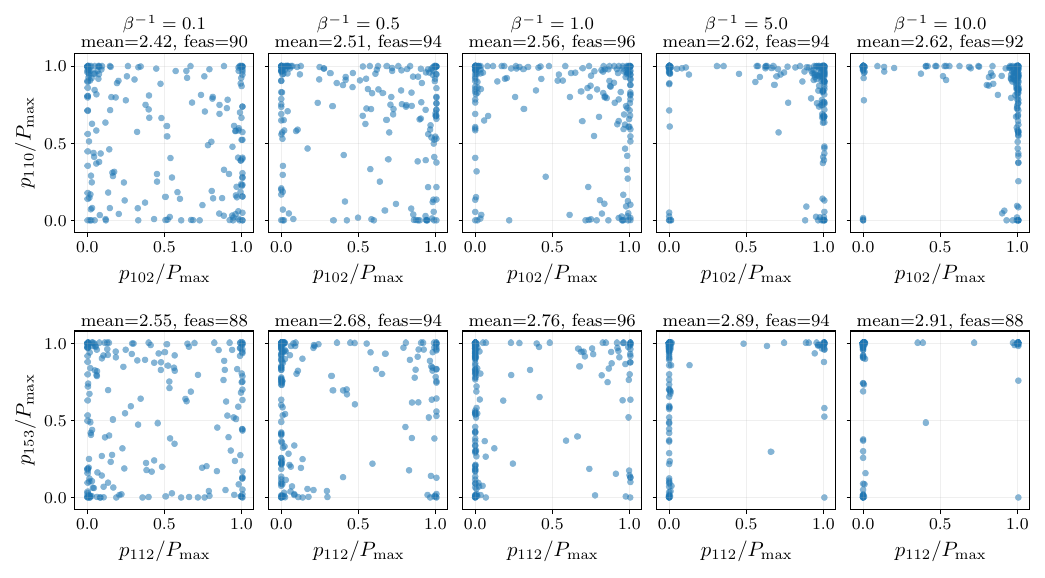}
    \hspace{-10pt}
    \caption{\small \textbf{Temperature effect.} Two-dimensional scatter plots of generated samples of two test networks. The title shows the mean rates and feasibility of the network. Higher temperature (left) promotes more diverse allocations while lower temperature (right) generates more degenerate samples. The value $\beta=1$ provides a good balance between mean rates and feasibility.}
    \label{fig:temp_wra}
\end{figure}

\textbf{Temperature effect.} In Figure~\ref{fig:temp_wra}, we analyze the effect of the temperature parameter $\beta$. Large $\beta$, equivalently small $\beta^{-1}$, generate more uniform samples, which cause severe interference and in turn the lowest mean rates and feasibility. Lower temperatures produce more degenerate distributions that concentrate the samples at the corners. This prevents severe interference and provides better mean rate, but does not fully benefit from time sharing (i.e., low feasibility). We find that $\beta =1$ in this experiment balances the mean rates with feasibility.

\begin{figure}
    \centering
    \includegraphics[width=\linewidth]{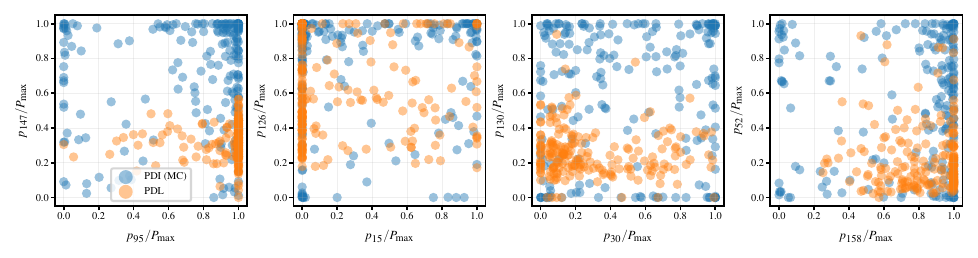}
    \vspace{-20pt}
    \caption{\small \textbf{Annealing effect of PDI.} A scatter plot of the predicted power allocation of four pairs of agents. PDI discovers more transmission modes, allowing more agents to transmit with higher power to achieve higher rates on average. PDL concentrates the samples in the same modes. }
    \label{fig:annealing}
\end{figure}

\textbf{Noise annealing effect.} PDI differs from PDL because it follows the score fields of a sequence of progressively denoised distributions, rather than the score field of the clean target distribution. This mechanism is reminiscent of simulated annealing. Early noisy diffusion steps smooth the target landscape and promote exploration across modes, while later low-noise steps gradually sharpen the distribution and refine the samples toward the constrained target. Unlike classical simulated annealing, however, the annealing in PDI is induced by the diffusion noise schedule rather than by the temperature of the energy function. 
Figure~\ref{fig:annealing} shows this effect, represented in the diverse samples generated by PDI compared with the more concentrated samples produced by PDL.

\subsection{Hyperparameters Choices}
Table \ref{tab:wra_ced_hparams} collects the hyperparameters of the problem and PDI training.
All experiments were run on a single NVIDIA GeForce RTX 3090 card.

\begin{table}[t]
\centering
\caption{PDI hyperparameters for the wireless power allocation experiment.}
\label{tab:wra_ced_hparams}
\small
\setlength{\tabcolsep}{4pt}
\begin{tabular}{@{}lc@{\hspace{1.5em}}lc@{}}
\toprule
\textbf{Parameter} & \textbf{Value}
&
\textbf{Parameter} & \textbf{Value} \\
\midrule

\multicolumn{2}{@{}l}{\emph{Problem setup}}
&
\multicolumn{2}{l@{}}{\emph{Training}} \\
Number of users $N$ & 200
&
Optimizer & AdamW \\
Max transmit power $P_{\max}$ & 10\,dBm
&
Learning rate & $10^{-3}$ $(\to 10^{-5})$ \\
Bandwidth & 40\,MHz
&
Weight decay & $10^{-4}$ \\
Minimum rate $r_{\min}$ & 0.6\,bps/Hz
&
Outer iterations $N_{\text{outer}}$ & 400 \\
Number of networks & 256
&
& \\

\midrule
\multicolumn{2}{@{}l}{\emph{Score network}}
&
\multicolumn{2}{l@{}}{\emph{Training}} \\
Hidden dimension & 256
&
Inner SGD steps $N_{\text{inner}}$ & $10 \to 30$ \\
GNN layers & 8
&
Batch size $|\ccalB_{\text{train}}|$ & 4 problems \\
Filter order $K$ & 2
&
Replay buffer capacity $|\ccalB|$ & 4\,096 \\
Time embedding dimension & 128
&
Minibatch size $B$ & 128 \\
Conditioning channels & 4 $(\bbx,t,\bblambda,\ccalG)$
&
& \\
Activation / normalization & SiLU / LayerNorm
&
& \\

\midrule
\multicolumn{2}{@{}l}{\emph{Diffusion process}}
&
\multicolumn{2}{l@{}}{\emph{Training -- perturbation and prior}} \\
Timesteps $T$ & 500
&
Exploitation  fraction $\rho_{\text{exp}}$ & 0$\to$0.7 \\
Noise schedule & Cosine
&
$\bblambda$ prior range $[\nu_{\min}, \nu_{\max}]$ & $[2,10]$ \\
MC candidates $K_{\text{MC}}$ & $8$
&
Perturbation std $(\epsilon_{\bbx}/\epsilon_{\bblambda})$ & $0.1/0.5$ \\
&
&
Perturbation fraction $\rho_{\text{pert}}$ & 0.5 \\
\midrule
\multicolumn{2}{@{}l}{\emph{Dual variable (inference)}}
&
\multicolumn{2}{l@{}}{\emph{Evaluation}} \\
Inverse temperature $\beta^{-1}$ & 1.0
&
Policies per network $K$ & 200 \\
Step size $\eta_t$ & 0.05
&
Time-sharing timeslots $R_{\mathrm{eval}}$ & 500 \\
Initialization $\bblambda_0$ & 10.0
&
Test networks & 64 \\
Clamp $\bblambda_{\max}$ & 50
&
Evolution trials & 50 \\
Sub-batch groups $C$ & 10
&
& \\

\bottomrule
\end{tabular}
\end{table}

%% file: sections/appendix_portfolio.tex
\section{Extended Numerical Results: Portfolio Management} \label{app:pf}

We consider constrained portfolio optimization over $N=500$ assets organized into 10 sectors, with returns generated from a latent factor model. The goal is to learn a distribution $\mu(\bbx)$ over portfolio weights $\bbx\in\Delta^{N-1}$ that maximizes expected return while satisfying per-asset risk, sector-exposure, and diversification constraints on average:
\begin{align}\label{eq:portfolio}
    \max_{\mu(\bbx)}
    \quad &
    \mbE_{\bbx\sim\mu}
    \left[
        \frac{1}{R}\sum_{k=1}^{R}\bbr^{(k)\top}\bbx
    \right]
    + \beta H(\mu)
    \nonumber\\
    \text{s.t.}\quad
    &
    \mbE_{\bbx\sim\mu}
    \left[x_j(\bbSigma\bbx)_j\right]
    \le b_j,
    && j=1,\ldots,N,
    \nonumber\\
    &
    \underline e_s
    \le
    \mbE_{\bbx\sim\mu}
    \left[\sum_{j\in\mathcal S_s}x_j\right]
    \le
    \overline e_s,
    && s=1,\ldots,S,
    \nonumber\\
    &
    \mbE_{\bbx\sim\mu}
    \left[
        N_{\mathrm{eff}}(\bbx)
    \right]
    \ge N_{\mathrm{eff}}^{\min},
\end{align}
where
\begin{align}
    N_{\mathrm{eff}}(\bbx)
    =
    \exp\!\left(
        -\sum_{j=1}^{N}x_j\log x_j
    \right).
\end{align}
The first set of constraints limits the average variance contribution of each asset. The second bounds the average exposure to each sector, while the last promotes diversification by imposing a minimum effective number of assets on average. Altogether, the problem contains 521 inequality constraints.

 In practice, portfolio construction is rarely a single-solution problem. Even under common market assumptions, different clients or rebalancing instances may require different allocations due to heterogeneous preferences, liquidity needs, tax considerations, and model uncertainty. This motivates learning a distribution over portfolios rather than a single deterministic allocation. 
 The average risk constraints allow the sampler to trade risk across different portfolio samples, admitting occasional high-risk, high-return allocations while ensuring that the overall exposure of each asset remains within its prescribed budget in expectation.

 We note that $\ccalX$ is the probability simplex in $\reals^N$, and volume, uniform distribution, and entropy are taken with respect to its $(N{-}1)$-dimensional volume. 
 % Differential entropy in $\reals^N$ equals $-\infty$ for every distribution supported on the simplex.

 \subsection{Data \& Architecture}

 \paragraph{Data.} We generate synthetic portfolio instances using a latent factor model with sector structure. Each instance consists of $N=500$ assets organized into $S=10$ sectors. Each asset's log-returns are drawn from a multivariate normal with a sector-structured covariance matrix $\bbSigma$, where assets within the same sector are more correlated with each other than with assets in other sectors. We draw $1000$ return scenarios and convert to simple returns via exponentiation.

 \paragraph{Graph construction.} To construct the graph, we compute the absolute correlation matrix from $\bbSigma$, remove self-loops, and for each asset retain edges to its $k$-nearest neighbors (we use $k{=}20$) by absolute correlation magnitude. Edge weights are normalized by the largest eigenvalue of the resulting adjacency matrix.

\paragraph{Conditioning and architecture.}
We use the same GNN architecture used in the wireless allocation problem. The timestep information is injected through  Feature-wise Linear Modulation (FiLM).
% a sinusoidal embedding produces per-layer scale and shift parameters applied after LayerNorm.  
The shared dual variable
$\bblambda\in\reals^N$ is concatenated to the input features.
The network comprises $L{=}6$ residual blocks with hidden dimension $d_h{=}128$.

\subsection{Baselines}

The Hyperparameters of the baselines are summarized in Table \ref{tab:pf_baseline_hparams}.

\begin{table}[t]
\centering
\caption{Baseline hyperparameters in the portfolio management problem.}
\label{tab:pf_baseline_hparams}
\small
\setlength{\tabcolsep}{4pt}
\begin{tabular}{@{}lc@{\hspace{1.5em}}lc@{}}
\toprule
\textbf{Parameter} & \textbf{Value}
&
\textbf{Parameter} & \textbf{Value} \\
\midrule

\multicolumn{4}{@{}l}{\emph{PD-Langevin}} \\
Primal step size $\eta_\rmp$ & $10^{-2}$
&
Iterations & 500 \\
Dual step size $\eta_\rmd$ & 100
&
& \\

\midrule
\multicolumn{4}{@{}l}{\emph{PDM}} \\
Projection iterations & 50
&
$\bblambda$ (net input) & 0 \\

\midrule
\multicolumn{4}{@{}l}{\emph{PDI--MC}} \\
Inverse temperature $\beta^{-1}$ & 2\,000
&
dual step size $\eta_t$ & 300 \\
MC candidates $K_{\text{MC}}$ & 512
&
$\bblambda_{\text{decay}}$ & 0.001 \\

\midrule
\multicolumn{4}{@{}l}{\emph{Evaluation}} \\
Generated samples & 1024
&
Test instances & 40 \\

\bottomrule
\end{tabular}
\end{table}

\subsection{Additional Results}
We ran additional experiments to evaluate different aspects of our algorithm, and the results match our previous findings.

\begin{figure}
     \centering
     \includegraphics[width=0.48\linewidth]{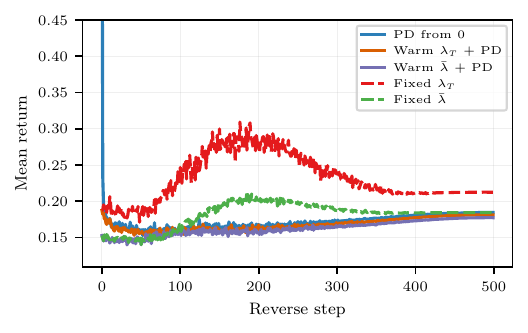}
     \includegraphics[width=0.48\linewidth]{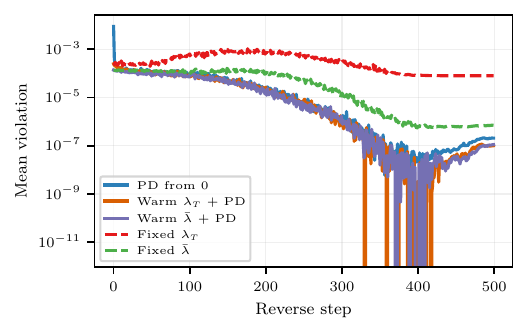}
     \caption{\textbf{Performance of PDI while fixing $\bblambda$ along the diffusion steps.} The time-average dual variable gives better objective and constraint satisfaction compared to the last iterate. However, running the dual updates beats both of them and does not depend on the initial value.}
     \label{fig:pf-fixed-lam}
 \end{figure}

First, we compare the full PDI algorithm with variants that freeze the dual variable after an initial primal-dual run and report the results in Figure \ref{fig:pf-fixed-lam}. In this experiment, we consider the last dual iterate and the time-averaged dual variable. The results show that the time-averaged multiplier performs better than the last iterate. 
% This is consistent with our convergence result, which guarantees convergence for the time-average of the dual variables rather than for the final iterate itself. 
However, both fixed-dual variants remain below the full PDI performance. We also notice that the initial $\bblambda_0$ does not affect the PDI trajectory. This suggests that PDI is not simply estimating one good multiplier and then sampling from it. Its advantage comes from updating the multiplier throughout the denoising trajectory.

\begin{figure}
     \centering
     \includegraphics[width=0.48\linewidth]{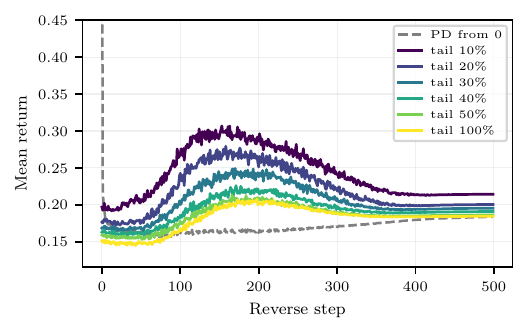}
     \includegraphics[width=0.48\linewidth]{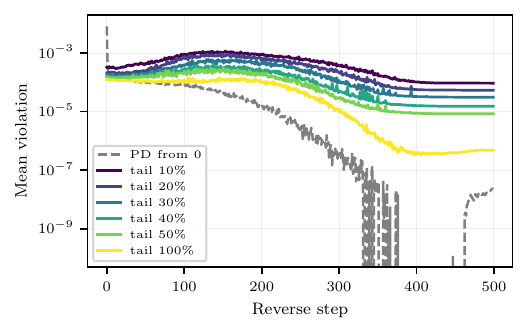}
     \caption{\textbf{Frozen $\bar\bblambda_{\text{tail}}$ during sampling.} The effect of running the sampler with a fixed dual variable equal to the time-average of the tail dual iterates.}
     \label{fig:pf-tail}
 \end{figure}

Second, we test in Figure \ref{fig:pf-tail} whether using only a late-stage average of the dual trajectory is enough. The answer is negative. The performance degrades when the average is computed only over a suffix of the trajectory. This suggests that the early dual iterates are not merely transient noise; they contribute to the effective multiplier that governs the sampled distribution. 
% In other words, the full trajectory matters, but its aggregate effect can be captured well by the full time-average.

\begin{figure}
     \centering
     \includegraphics[width=0.48\linewidth]{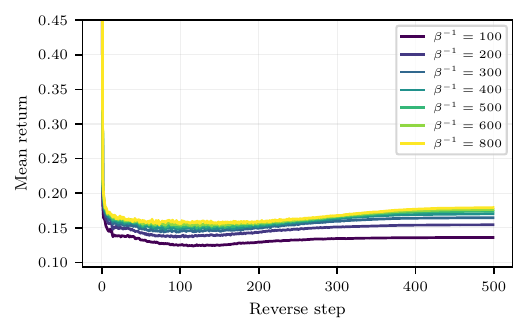}
     \includegraphics[width=0.48\linewidth]{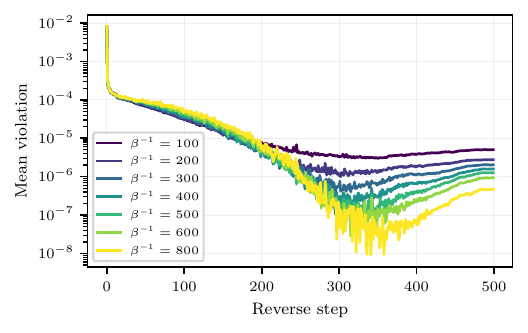}
     \caption{\textbf{Temperature effect.} Lower temperature $\beta$, equivalently larger $\beta^{-1}$ (yellow), helps improve returns and constraint satisfaction.}
     \label{fig:pf-temp}
 \end{figure}

Our third experiment in Figure \ref{fig:pf-temp} investigates the role of the entropy regularization on the final solutions and trajectories. The figure shows that when the temperature $\beta$ is high, the sampler produces more diverse samples, but the objective can be weaker. When the temperature is lower, the distribution becomes more concentrated around high-quality solutions. This improves the objective, but it can also make the dynamics sharper and more sensitive. 
% Therefore, the temperature controls the tradeoff between diversity, objective value, and stability.

\begin{figure}
     \centering
     \includegraphics[width=0.48\linewidth]{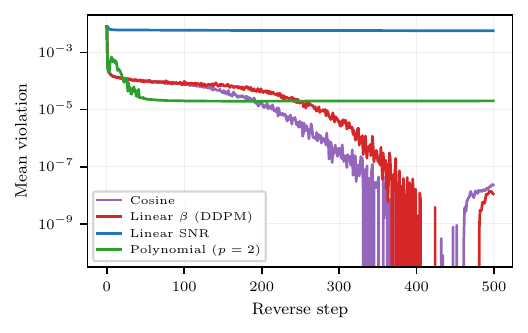}
     \caption{\textbf{Effect of noise schedules.} Standard DDPM cosine and linear schedules give better feasibility. Schedules that reconstruct the clean sample too early make the sampler sensitive to perturbations in the dual variable before the dual variable has stabilized.}
     \label{fig:pf-noise}
 \end{figure}

Finally, the noise-schedule comparisons of Figure \ref{fig:pf-noise} show that the schedule is important for convergence. Schedules that keep the process noisy for longer, such as the standard cosine and DDPM linear schedules, give more stable behavior and better feasibility. In contrast, schedules that reconstruct the clean sample too early make the sampler sensitive before the dual variable has stabilized. This leads to larger violations and weaker convergence.

\subsection{Hyperparameter Choices}
We ran a sweep over $\beta$ and $\eta_t$. Table \ref{tab:portfolio_hyperparams} displays all hyperparameters used in experiments.
All experiments were run on a single NVIDIA GeForce RTX 3090 card.

\begin{table}[t]
\caption{PDI hyperparameters in the portfolio management problem (training and inference).}
\label{tab:portfolio_hyperparams}
\small
\centering
\begin{tabular}{@{}llll@{}}
\toprule
\textbf{Parameter} & \textbf{Value} & \textbf{Parameter} & \textbf{Value} \\
\midrule
\textit{Diffusion process} & & \textit{Training} & \\
Timesteps $T$ & 500 & Optimizer & AdamW \\
Noise schedule & Cosine & Learning rate & $3\times10^{-4}$ \\
Inverse temperature $\beta^{-1}$ & 1 & $\lambda$ prior $[\nu_{\min}, \nu_{\max}]$ & $[0.001, 60]$ \\
MC candidates $K_{\mathrm{MC}}$ & 512 & $\lambda$ perturbation (mult.) $\epsilon_\lambda$ & 0.5 \\
 & & Exploitation ratio $\rho_{\exp}$ & $0 \to 0.7$ \\
 & & Minibatch size $B$ & 128 \\
 & & Training instances & 200 \\
\midrule
\textit{Dual variable (inference)} & & \textit{Training} & \\
Step size $\eta_t$ & 0.01 & Validation instances & 5 \\
Initialization $\boldsymbol{\lambda}_0$ & 7 & Inner SGD steps $N_{\mathrm{inner}}$ & $10 \to 80$ \\
Clamp $\lambda_{\max}$ & 500 & Outer iterations $N_{\mathrm{outer}}$ & 800 \\
Weight decay & 0.0 & Replay buffer capacity & 8\,192 \\
Constraint normalization & None & Rollout minibatch $|\mathcal{B}_{\mathrm{train}}|$ & 4 problems \\
 & & Rollout batch size & 1\,024 \\
\midrule
\textit{Score network} & & & \\
Backbone & GNN & & \\
Hidden dimension $d_h$ & 128 & & \\
Residual blocks $L$ & 6 & & \\
Graph filter order $K$ & 1 & & \\
Graph sparsity (top-$k$) & 20 & & \\
Conditioning & FiLM + $\boldsymbol{\lambda}$ concat & & \\
\bottomrule
\end{tabular}
\end{table}